\documentclass[pdflatex,sn-basic,iicol]{sn-jnl}

\usepackage{graphicx}%
\usepackage{multirow}%
\usepackage{amsmath,amssymb,amsfonts}%
\usepackage{amsthm}%
\usepackage{mathrsfs}%
\usepackage[title]{appendix}%
\usepackage{xcolor}%
\usepackage{textcomp}%
\usepackage{manyfoot}%
\usepackage{booktabs}%
\usepackage{algorithm}%
\usepackage{algorithmicx}%
\usepackage{algpseudocode}%
\usepackage{listings}%
\usepackage{subcaption}
\usepackage{caption}
\usepackage{newfloat}
\usepackage{xspace}
\usepackage{graphbox}
\usepackage{wrapfig}
\usepackage{tikz}
\usepackage{adjustbox}
\usepackage{stackengine}
\usepackage{tablefootnote}
\usepackage{ragged2e}
\usepackage{tabularx}
\usepackage{bbm}
\usepackage{dsfont}
\usepackage{colortbl}
\usepackage[inline, shortlabels]{enumitem}
\usepackage{ifthen}
\usepackage[normalem]{ulem}
\newcommand{\hfilll}{\hspace{0pt plus 1filll}}

\makeatletter
\newcommand*{\rom}[1]{\expandafter\@slowromancap\romannumeral #1@}
\makeatother

\DeclareMathOperator*{\argmax}{arg\,max}

\makeatletter
\DeclareRobustCommand\onedot{\futurelet\@let@token\@onedot}
\def\@onedot{\ifx\@let@token.\else.\null\fi\xspace}

\def\eg{\emph{e.g}\onedot} 
\def\ie{\emph{i.e}\onedot} 
 
 \def\vs{\emph{vs}\onedot}
 
\def\etal{\emph{et al}\onedot}
\makeatother

\makeatletter
\def\OurMethod{HUWSOD\xspace}
\def\OurMethodx{HUWSOD$^*$\xspace}

\makeatother

\usepackage{natbib}
\usepackage{hyperref}

\theoremstyle{thmstyleone}%
\theoremstyle{thmstyletwo}%

\theoremstyle{thmstylethree}%

\begin{document}

\title[Article Title]{\OurMethod: Holistic Self-training for Unified Weakly Supervised Object Detection}


\author[1]{\fnm{Liujuan} \sur{Cao}}\email{caoliujuan@xmu.edu.cn}

\author[1]{\fnm{Jianghang} \sur{Lin}}\email{hunterjlin007@stu.xmu.edu.cn}

\author[1]{\fnm{Zebo} \sur{Hong}}\email{debology@stu.xmu.edu.cn}

\author*[2]{\fnm{Yunhang} \sur{Shen}}\email{shenyunhang01@gmail.com}

\author[3]{\fnm{Shaohui} \sur{Lin}}\email{shaohuilin007@gmail.com}

\author[1]{\fnm{Chao} \sur{Chen}}\email{chenchao.tencent@gmail.com}

\author[1]{\fnm{Rongrong} \sur{Ji}}\email{rrji@xmu.edu.cn}

% \author*[1,2]{\fnm{First} \sur{Author}}\email{iauthor@gmail.com}



% \affil[1]{\orgdiv{Key Laboratory of Multimedia Trusted Perception and Efficient Computing, Ministry of Education of China}, \orgname{School of Informatics, Xiamen University}, \orgaddress{\street{No. 4221, Xiang'an South Road, Xiang'an District}, \city{Xiamen}, \postcode{361102}, \state{Fujian}, \country{China}}}
\affil[1]{\orgdiv{Media Analytics and Computing Lab, School of Informatics}, \orgname{Xiamen University}, \orgaddress{ \postcode{361102}, \country{China}}}

\affil[2]{\orgdiv{Youtu Lab}, \orgname{Tencent}, \orgaddress{\postcode{200233}, \country{China}}}

\affil[3]{\orgdiv{School of Computer Science and Technology}, \orgname{East China Normal University}, \orgaddress{\postcode{200062}, \country{China}}}

\abstract{
% Informal abstract. I like you to write an initial abstract (could be rewritten several time latter on) to clarify what you are doing, what is the basic idea, and what would be the expected results.
As an emerging problem in computer vision, weakly supervised object detection~(WSOD) aims to use only image-level annotations to train object detectors.
Most WSOD methods rely on traditional object proposals to generate candidate regions and are confronted with unstable training, which easily gets stuck in a poor local optimum.
In this paper, we propose a unified and high-capacity WSOD network with holistic self-training framework, termed \OurMethod, which is self-contained, and requires no external modules or additional supervision.
To this end, the proposed framework innovates in two perspectives, \ie, unified network structure and self-training scheme, both of which are rarely touched in WSOD before.
First, we design a self-supervised proposal generator and an autoencoder proposal generator with multi-rate resampling pyramid to hypothesize object locations, which replace the traditional object proposals and enable the strictly end-to-end training and inference for WSOD.
Second, we introduce holistic self-training scheme that consists of step-wise entropy minimization and consistency-constraint regularization, which refine both detection scores and coordinates progressively, as well as enforcing consistency to match predictions produced from stochastic augmentations of the same image.
Extensive experiments on PASCAL VOC and MS COCO show that the proposed \OurMethod performs competitively to the state-of-the-art WSOD methods, while getting rid of offline proposals and additional data.
Moreover, the upper-bound performance of \OurMethod with class-agnostic ground-truth bounding boxes approaches Faster R-CNN, which demonstrates that \OurMethod is capable to achieve fully-supervised accuracy.
Our work also brings one important finding that random initialized boxes, although drastically different from the offline well-designed object proposals, are also effective object candidates for WSOD training.
The code is available at: \url{https://github.com/shenyunhang/HUWSOD}.
}


\maketitle

\section{Introduction}
\label{sec_Introduction}
%You need to introduce, paragraph-by-paragraph, the following points:
%(1)
%This problem is important and emerging. The basic idea of solution in this problem
%(2)
%How the existing methods work. What is the key drawback(s) here. And why the existing methods cannot tackle the drawback(s)
%(3)
%We propose to address this drawback from a novel aspect of... The key idea is...
%(4)
%More specially, we present a new method (framework\algorithm), termed, ***. This methods works by the following steps:....
%(5)
%In sum, the main contribution we made are: (1),...(2),..(3),...
%(6)
%We have quantitatively evaluation on **** datasets, with comparisons to *** cutting-edge methods, we report *** performance gains over the state-of-the-arts.
%(7)
%The rest of this paper is organized as:...
%%%%%%%%%%%%%%%%%%%%%%%%%%%%%%%%%%%%%%%%%%%%%%%%%%%
Last decade has witnessed rapid advances in object detection~\cite{FASTRCNN,Cai2019,MASKRCNN,FASTRCNN}, which are accommodated by open-source object datasets such as PASCAL VOC~\cite{PASCALVOC} and MS COCO~\cite{MSCOCO}.
Despite this great success, the expensive object annotations have long plagued the scalable and real-world application of existing fully supervised object detection~(FSOD).
To reduce heavily instance-level annotations, \ie, bounding-box and dense-pixel labels, one promising potential is weakly supervised learning that trains object detectors from only image-level supervisions that indicating the presence or absence of objects, termed weakly supervised object detection~(WSOD).
\begin{figure*}[t]
    \begin{Center}
        \begin{subfigure}[t]{0.45\linewidth}
            \centering
            \includegraphics[width=1.0\textwidth]{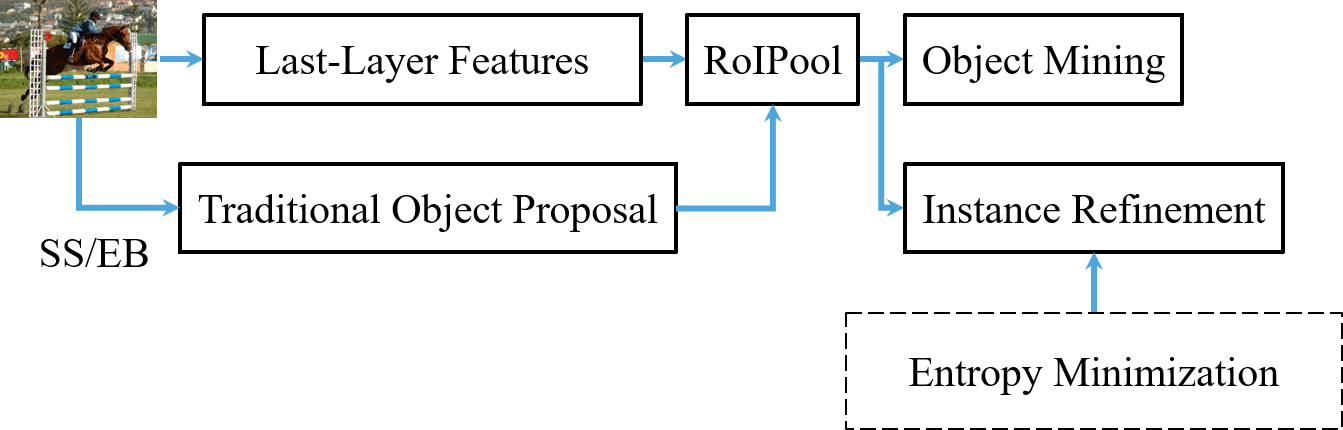}
            \caption{Common WSOD framework.}
            \label{fig_head_wsod}
        \end{subfigure}
        \hspace{10pt}
        \begin{subfigure}[t]{0.48\linewidth}
            \centering
            \includegraphics[width=1.0\textwidth]{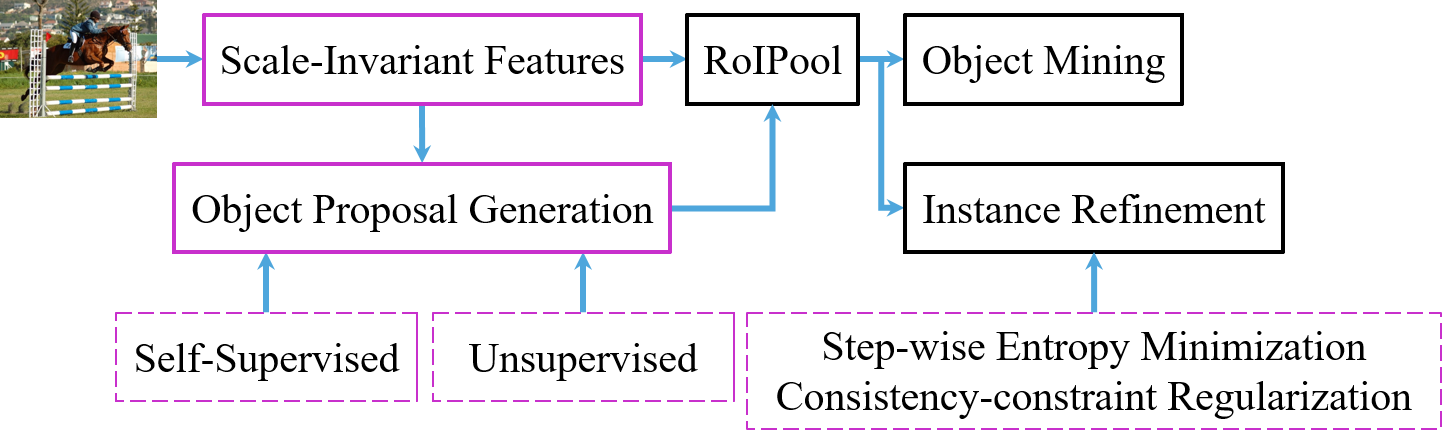}
            \caption{The proposed HUWSOD framework.}
            \label{fig_head_our}
        \end{subfigure}
    \end{Center}
    \caption{
        An illustration of common WSOD (a) and the proposed HUWSOD framework (b). Our key innovations compared to existing methods include end-to-end object proposal generators and holistic self-training scheme.
    }
    \label{fig_head}
    \vspace{-20pt}
\end{figure*}

Most cutting-edge WSOD paradigms follow a two-phase training procedure, which consists of \textit{object mining} and \textit{instance refinement} phases, as illustrated in Fig.~\ref{fig_head_wsod}.
The object mining phase is commonly formulated as an image classification task via multiple-instance learning~(MIL)~\cite{Amores2013}, which treats input images as bags and candidate regions of the images as instances~\cite{Cinbis2015,Wang2015}, \ie, a positive bag contains at least one positive instance for the target category and all instances in negative bags are negative samples.
Typically, those candidate regions are estimated offline from images via traditional proposal methods, such as selective search~(SS)~\cite{Uijlings2013} and edge boxes~(EB)~\cite{Zitnick2014}.
Then, instance-level MIL classifiers~\cite{Amores2013} are trained to distinguish candidate regions among objects and background~\cite{Bilen2016,Kantorov2016,Wei2018TS2C}.
After obtaining potential object instances, instance refinement phase is kicked in to build multiple parallel branches to further minimize classification entropy, \ie, encouraging the model to produce higher-confidence results, and adjust object localization by using the preceding predictions as pseudo targets~\cite{Tang2017,Yanga}.
Although the above paradigm has achieved promising results, there still exist two major challenges in WSOD.
First, object proposal modules work in an external manner, which is independent of the object detection learning and less efficient.
Recent endeavors in~\cite{Tang,Singh,Cheng2020} proposed to improve object proposals especially for WSOD.
However, those methods are still not in an end-to-end fashion, which simply ensemble traditional object proposals~\cite{Tang} and motion segmentation~\cite{Singh} with additional video dataset~\cite{Cheng2020}, respectively.
Besides, WSOD is sensitive to initialization, thus how to produce high-quality proposals in the early training stage to alleviate the instability remains unexplored.
On the other hand, the existing WSOD methods use multi-scale image pyramids to remedy the scale-variation problem during training and inference, which however neglect how to design the internal network architecture to strengthen single-image proposals.
Second, most instance refinement methods are based on entropy minimization principle with pseudo labels~\cite{Diba2017,Tang2017}, \ie, using high-confident prediction as pseudo-ground-truth bounding boxes, which however may not well cover objects and are unstable~\cite{Liu}, thus over-fitting to incorrect pseudo-labels.
One way to reduce mislocalizations is bounding-box regression~\cite{Gao2018,Zeng2019,Fang2020,Ren2020}.
However, those methods either require super-pixel evidence and additional supervision to fine-tune bounding boxes or fail to achieve the trade-off between precision and recall in different branches of instance refinement, which also easily gets stuck in a poor local minimum.
Meanwhile they only consider a one-off unidirectional connection between instance refinement branches, \ie, taking predictions from the preceding branch to supervise the succeeding one, which neglects the mutual benefits between MIL classifiers and instance refinement.
In this paper, we propose a holistic self-training based, unified weakly-supervised object detection framework, termed \OurMethod, which innovates in two essential prospects, \ie, {unified network structure} and {holistic training scheme}, both of which are rarely touched in WSOD before.
The proposed \OurMethod framework is illustrated in Fig.~\ref{fig_head_our}.
To design a unified network structure, we focus on learnable object proposals and make use of multi-scale information in WSOD setting.
First, we propose two object proposal generators, namely self-supervised object proposal generator~(SSOPG) and autoencoder object proposal generator~(AEOPG), both of which are end-to-end training to hypothesize object locations without additional supervision.
Specifically, SSOPG leverages final results predicted by \OurMethod as objectness classification and regression targets in a self-supervised fashion, while AEOPG is designed to capture salient objects by learning low-rank decomposition of feature maps via an unsupervised autoencoder.
Second, to further increase proposal diversities and enrich object features, we construct a multi-rate resampling pyramid~(MRRP) to aggregate multi-scale contextual information with various dilation rates in backbone networks, which is the first network-embedded feature hierarchy to handle scale variation in WSOD.
Different from common FSOD that attaches new parameters to build feature pyramids~\cite{Lin2017a}, MRRP does not need to learn new parameters and thus avoids over-fitting by sharing the same parameters of pre-trained backbones.
We further introduce a holistic self-training scheme including step-wise entropy minimization~(SEM) as well as consistency-constraint regularization~(CCR), which seamlessly reduces entropy as the dominant WSOD paradigms while integrating consistency constraint.
Our motivation of CCR lies in the smoothness assumption~\cite{VanEngelen2020}:
Two object instances that lie close together in the input space should have the same label and minor variations in the input space should only cause minor variations in the output space.
This assumption is commonly used in semi-supervised learning, such as pseudo-ensemble agreement regularization~\cite{Bachman2014}, but has little explore for WSOD.
In this paper, we extend the smoothness assumption transitively to the learning of weakly-labeled data.
In details, SEM progressively selects high-confidence object proposals as positive samples to reduce entropy and bootstraps the quality of predicted bounding boxes.
And CCR enforces consistency to match predicted boxes produced from stochastic augmentations, \ie, views, of the same image.
We also explore the exponential moving average scheme to propagates partial learned knowledge among object mining and instance refinement to improve the performance of all branches.
We conduct extensive experiments on PASCAL VOC~\cite{PASCALVOC} and MS COCO~\cite{MSCOCO} to elaborately confirm the effectiveness of our method.
In particular, the proposed \OurMethod achieves competitive results with the state-of-the-art WSOD methods while not requiring external modules or additional supervision.
Our ablation study also includes basic fully-supervised learning experimental setting that are often ignored or not reported when new WSOD methods are proposed.
The results shows that the upper-bound performance of \OurMethod with class-agnostic ground-truth bounding boxes approaches Faster R-CNN~\cite{FASTERRCNN}, which demonstrates that \OurMethod is capable to achieve fully-supervised accuracy.
We also quantitatively hint that using traditional proposal modules in WSOD is a historical workaround so far when the community mainly focused on object mining and instance refinement phases.
In addition, traditional proposals have been largely thought of as a ``free'' resource, attributed to a series of open-source codes and benchmarks.
However, along with ever-increasing data and computation resources, \eg, large-scale web images and high-capacity models, learning end-to-end object proposal generators becomes more feasible, especially when training data contains information like image-level labels, that can not be well digested by traditional object proposal approaches.
\OurMethod is able to learn effective object candidates from random initialized boxes with only image-level labels, which enables strictly end-to-end training and inference at scale.
The contributions of this work are concluded as follows:
\begin{itemize}
    \item We propose a unified and high-capacity weakly supervised object detection~(WSOD) framework, which is self-contained and requires \textbf{no} external modules in online inference nor additional supervision in offline training, to develop a general WSOD model.
    \item We design two object proposal generators with self-supervised and autoencoder-style unsupervised learning, respectively, which is cooperated with a multi-rate resampling pyramid to utilize contextual information.
    \item We introduce a holistic self-training scheme to seamlessly reduces classification entropy while integrating consistency constraint, which make full use of weakly-labeled data.
\end{itemize}
Compared to our conference version in~\cite{Shen2020UWSOD}, we first extend the original self-supervised object proposal generator as well as a complementary object proposal generator using an unsupervised autoencoder.
We further exploit consistency-based self-training with exponential moving average, which is widely overlooked in the existing literatures.
Equipped with autoencoder object proposal generator and consistency-constraint regularization, \OurMethod significantly outperforms the original method~\cite{Shen2020UWSOD}.
Detailed analysis is further given with an in-depth comparison to the most recent related works.
Finally, we provide more experiment results to shown the effectiveness of our method.
The rest of our paper is organized as follows.
In Sec.~\ref{sec_Related_Work}, related work is introduced.
In Sec.~\ref{sec_The_Proposed_Method}, the details of our method are described.
Elaborate experiments and analyses are conducted in Sec.~\ref{sec_Quantitative_Evaluation}.
Finally, conclusions and future directions are discussed in Sec.~\ref{sec_Conclusion}.
%%%%%%%%%%%%%%%%%%%%%%%%%%%%%%%%%%%%%%%%%%%%%%%%%%%%%%%%%%%%%%%%%%%%%%%%%
\section{Related Work}
\label{sec_Related_Work}
%You should categorize the related works into several groups, like related works in this problem, related works in the method, etc. Each group is a sub-section.
%
\subsection{Weakly Supervised Object Detection}
Most cutting-edge WSOD procedures consist of two phases: \textit{object mining} and \textit{instance refinement}.
The object mining phase is formulated using multiple-instance learning~(MIL) to implicitly model latent object locations with image-level labels, which alternates between localizing object instances and training an appearance model.
Given a target class, positive bags are assumed to contain at least one instance of the object, while negative bags only have background samples or other objects.
The goal is to train instance-level classifiers that distinguish object presence or absence.
Early efforts, such as multi-fold training~\cite{Cinbis2015}, objective function smoothing~\cite{song2014learning}, and MIL constraint relaxation~\cite{Wang2015}, focused on directly solving the classical MIL problem.
Bilen~\etal~\cite{Bilen2016} selected proposals by parallel detection and classification networks in the deep learning era.
Contextual information~\cite{Kantorov2016}, gradient map~\cite{Shen2019CSC}, and semantic segmentation~\cite{Wei2018TS2C} are leveraged to learn outstanding object proposals.
Since the objective function is not convex, the optimization is easily trapped in a local optimum.
In this paper, we explore exponential moving average scheme that enables MIL classifiers to aggregate information from the subsequent instance refinement phrase, which alleviates local minimum and avoids over-fitting.
The instance refinement phase is typically posed as a self-training problem by implicitly constructing pseudo labels from high-confidence predictions on weakly-labeled data.
Thus, learning instance refinement can be viewed as a form of entropy minimization, which reduces the density of data points at the decision boundaries and encourages the network to make low-entropy predictions.
Tang~\etal~\cite{Tang2017} and Jie~\etal~\cite{Jie2017} took confident candidate proposals from the object mining phase to learn instance-level instance for acquiring tight positive objects.
Different strategies~\cite{Tang2018b,Kosugi2019,Zhang2018b,Yanga,Zhang,Chen2020,Ren2020} are also proposed to estimate pseudo targets and assign labels to proposals.
Some methods simultaneously learn above two-phase modules with randomness minimization~\cite{Li2016j,Wan}, proposal adjacent relationship~\cite{Wan2019,Zhang2020l}, utilizing uncertainty~\cite{Arun2018a,Liu}, generative adversarial learning~\cite{Shen2018GAL} and knowledge distillation~\cite{Zeng2019}.
Collaboration mechanisms are also introduced to take advantage of the complementary interpretations of different weakly supervised tasks~\cite{Shen2019WSJDS,Shen2021JTSM,Shen2021PDSL} and models~\cite{CMIDN}.
Our work proposes to combine consistency-based constraint with entropy minimization as a holistic self-training framework for WSOD, which makes full use of weakly-labeled data available to enhance robustness.
With the output of the above two-phase paradigm, a separated detector can also be trained to further improve performance.
Many efforts~\cite{Zhang,Ge2018} mine pseudo-ground-truth bounding boxes from pre-trained WSOD models for FSOD.
Recent, some works~\cite{Huang2022,Sui} also manage to select high-quality predictions to learn semi-supervised object detectors.
The proposed \OurMethod is a generic WSOD framework and is compatible with those orthogonal improvements by taking our well-trained \OurMethod models as the WSOD part in their framework.
Another line of research exploits knowledge transfer for cross-domain adaptation~\cite{Cao2021,Zhong2020,Hou2021,Xuc}, \ie, adapting the fully supervised detectors to a novel target domain with only image-level annotations.
Methods in~\cite{Fang2020,Xu,Yang2019,Shen2020fWebSOD,Chen2021b,Vo2022} trained object detectors from different supervisions.
Chen~\etal~\cite{Chen2021b} learned from small fully-labeled images and large point-labeled images.
In contrast to the above work, the proposed \OurMethod does not use any additional annotation or external proposal models.
%%%%%%%%%%%%%%%%%%%%%%%%%%%%%%%%%%%%%%%%%%%%%%%%%%%%%%%%%%%%%%%
\subsection{Object proposal generation}
Object proposal methods aim to generate candidate regions for object detection and image segmentation models.
Traditional methods are either based on grouping super-pixels, \eg, selective search~\cite{Uijlings2013} and multiscale combinatorial grouping~\cite{APBMM2014}, or based on sliding windows, \eg, edge boxes~\cite{Zitnick2014}.
In most existing WSOD works, object proposal methods were typically integrated as external modules independent of training the detection networks.
Few literatures exploit trainable object proposal generation under weakly supervised settings.
Cheng~\etal~\cite{Cheng2020} combined selective search~\cite{Uijlings2013} and gradient-weighted class activation map to generate proposals.
Tang~\etal~\cite{Tang} used a lightweight WSOD network~\cite{Tang2017} to refine the coarse proposals generated by edge boxes~\cite{Zitnick2014} on edge-like response maps.
Singh~\etal~\cite{Singh} learned object proposals from motion information in weakly-labeled videos.
Dong~\etal~\cite{Dong2021a} leveraged the bounding box regression knowledge from auxiliary dataset with instance-level annotations.
However, these methods are not in an end-to-end manner and still need traditional object proposals~\cite{Tang}, motion segmentation~\cite{Singh}, and additional video dataset~\cite{Cheng2020}.
Instead in this work, the proposed self-supervised generator and autoencoder proposal generator are end-to-end trainable, requiring no external modules nor additional information.
%%%%%%%%%%%%%%%%%%%%%%%%%%%%%%%%%%%%%%%%%%%%%%%%%%%%%%%
\subsection{Handling scale variation}
Representing features at multiple scales is crucial for object detection models to handle scale variation in challenging conditions.
Most WSOD methods adopt image pyramids to detect objects across scales during training and inference to remedy scale variation.
However, the image pyramid method increases the inference time and does not consider building internal network structures.
SNIP~\cite{Singh2018a} selectively trained the objects of appropriate sizes in each image scale.
However, SNIP is not adaptable to WSOD due to the lack of instance-level annotations.
Another stream of utilizing multi-scale information is to consider both low and high-level information, \eg, U-Net~\cite{Ronneberger2015} and FPN~\cite{Lin2017a}
However, it requires attaching new layers to build feature pyramids and may converge to an undesirable local minimum in WSOD.
In this paper, we introduce a simple yet efficient multi-rate resampling pyramid architecture to aggregate multi-scale contextual information, in which each level shares the same parameters and predicts objects independently before aggregation.
To the best of our knowledge, this is the first attempt to learn scale-invariant network-embedded feature hierarchy in WSOD.
%%%%%%%%%%%%%%%%%%%%%%%%%%%%%%%%%%%%%%%%%%%%%%%%%
\subsection{Consistency Regularization}
Consistency regularization minimizes variation in the outputs of models when they are subject to noise on their inputs or internal state~\cite{Bachman2014}.
Consistency regularization has been successfully applied to semi-supervised image classification~\cite{Sohn2020a,Xie2020c,Zhao2022b}, object detection~\cite{Liu2021a,Sohn2020,Chen2022d,Guo2022a} and semantic segmentation~\cite{Liu2022a,Fan2022} to encourage task prediction consistency, where unlabeled examples are used as regularizers to help smooth data manifold.
Inspired by STAC~\cite{Sohn2020}, we combine both consistency regularization and entropy minimization in a holistic self-training framework for WSOD, which serves as a new way of leveraging the weakly labeled data to find a smooth manifold on which the dataset lies.
Different from the student-teacher based siamese architecture in STAC~\cite{Sohn2020}, we explore consistency-constraint regularization among different branches within a single WSOD model, \ie, each branch is the student of the preceding branch and the teacher of the subsequent branch, simultaneously.
Our work is also related to CASD~\cite{Huang2020}, which proposed self-distillation to learn consistent attention on low-to-high feature layers.
Different from attention-based feature learning in CASD~\cite{Huang2020}, we conduct consistent regularization on detection predictions, which explicitly aggregates supervision within WSOD networks, such as the different pseudo labels produced under various image transformations and detection branches.
%%%%%%%%%%%%%%%%%%%%%%%%%%%%%%%%%%%%%%%%%%%%%%%%%
%%
%%
\begin{figure*}[t]
    \begin{center}
        \begin{subfigure}[t]{1.0\textwidth}
            \begin{center}
                \includegraphics[width=1.0\textwidth]{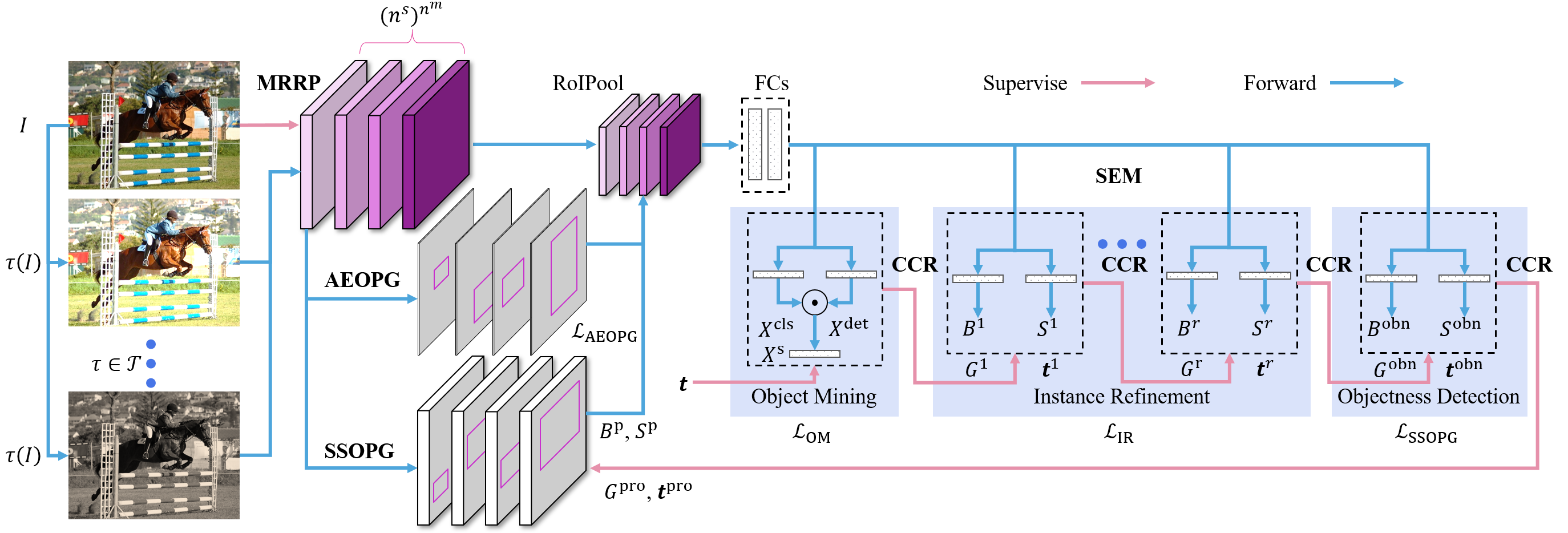}
            \end{center}
        \end{subfigure}
    \end{center}
    \caption{The overall framework of the proposed \OurMethod.
        We replace the traditional object proposals with a self-supervised object proposal generator~(SSOPG) and an autoencoder object proposal generator~(AEOPG), which hypothesize object locations in an end-to-end manner.
        We further construct a multi-rate resampling pyramid~(MRRP) to connect backbone and WSOD head, which strengthen the progress of object proposals.
        The optimization algorithm is based on a holistic self-training scheme that consists of step-wise entropy minimization~(SEM) and consistency-constraint regularization~(CCR).
    }
    \label{fig_framework}
\end{figure*}
%%%%%%%%%%%%%%%%%%%%%%%%%%%%%%%%%%%%%%%%%%%
%%
%%
\section{The Proposed Method}
\label{sec_The_Proposed_Method}
%Framework introduction
%Mathematical formulation
%Detailed algorithm
%Complexity analysis, etc.
%
In this section, the proposed \OurMethod framework is introduced.
In subsections~\ref{sec_The_Basic_Framework} and~\ref{sec_The_Overall_Framework}, we first introduce the preliminary of the basic WSOD and the proposed \OurMethod framework.
In subsection~\ref{sec_OPG}, self-supervised proposal generator~(SSOPG) and autoencoder object proposal generator~(AEOPG) are introduced to hypothesize candidate object locations.
A multi-rate resampling pyramid~(MRRP) is also constructed to aggregate multi-scale contextual information.
In subsection~\ref{sec_HST}, step-wise entropy minimization~(SEM) and consistency-constraint regularization~(CCR) are introduced to refine both detection scores and coordinates progressively and stably.
%%%%%%%%%%%%%%%%%%%%%%%%%%%%%%%%%%%%%%%%%%%%%
\subsection{The Preliminary}
\label{sec_The_Basic_Framework}
We first introduce the basic WSOD method which the proposed \OurMethod framework is built upon.
Given an input image, most methods firstly use traditional proposal algorithms, \eg, selective search~\cite{Uijlings2013} and edge boxes~\cite{Zitnick2014}, to extract candidate bounding boxes.
Then the corresponding proposal features are computed on the last image feature maps $F$ from backbones, \eg, VGG~\cite{VGGNet} and DRN~\cite{Shen2020DRN}, via RoIPool~\cite{FASTERRCNN} layer.
Lastly, those proposal features are fed into the WSOD head, which consists of object mining and instance refinement.
Object mining phase~\cite{Bilen2016,Kantorov2016} forks the proposal features into classification and detection streams, producing two score matrices $X^\mathrm{cls}, X^\mathrm{det} \in \Re^{n^\mathrm{pro} \times n^\mathrm{cat}}$ by two fully-connected layers, respectively.
$n^\mathrm{pro}$ and $n^\mathrm{cat}$ denote the number of object proposals and categories, respectively.
Both score matrices are normalized by the softmax $\sigma(\cdot)$ over categories and proposals, respectively: $\sigma(X^\mathrm{cls})_{p, c} = {e^{X^\mathrm{cls}_{p, c}}} / {\sum^{n^\mathrm{cat}}_{i=1}e^{X^\mathrm{cls}_{p, i}}}$ and $ \sigma(X^\mathrm{det})_{p, c} = {e^{X^\mathrm{det}_{p, c}}} / {\sum^{n^\mathrm{pro}}_{i=1}e^{X^\mathrm{det}_{i, c}}} $.
In this way, $\sigma(X^\mathrm{cls})_{p,c}$ estimates the probability of the $p^\mathrm{th}$ proposal belonging to the $c^\mathrm{th}$ category, and $\sigma(X^\mathrm{det})_{p,c}$ indicates the contribution of the $p^\mathrm{th}$ proposal to image being classified to the $c^\mathrm{th}$ category.
Thus, we interpret $\sigma(X^\mathrm{cls})$ as a term that predicts categories, whereas $\sigma(X^\mathrm{det})$ selects regions.
Then the element-wise product of the output of these two streams is again a score matrix: $X^\mathrm{s} = \sigma(X^\mathrm{cls}) \odot \sigma(X^\mathrm{det})$.
This equation can be considered as re-weighting classification results with localization scores.
To acquire image-level classification scores, we apply a sum pooling: $\mathbf{y}_c = \sum^{n^\mathrm{pro}}_{p=1}X^\mathrm{s}_{p,c}$.
Note that $\mathbf{y}_c$ is a weighted sum of the element-wise product of softmax normalized scores over all regions and falls in the range of $(0, 1)$.
We build binary cross-entropy objective function ${\mathcal{L}}_\mathrm{OM}$ as follows:
\begin{equation}
    \label{equ:loss_OM}
    %\begin{aligned}
    {\mathcal{L}}_\mathrm{OM} = \sum_{c=1}^{n^\mathrm{cat}} \Big\{ \mathbf{t}_c \log \mathbf{y}_c + (1-\mathbf{t}_c) \log (1-\mathbf{y}_c) \Big\},
    %\end{aligned}
\end{equation}
where $\textbf{t} \in \{0, 1\}^{n^\mathrm{cat}}$ is the image-level labels, and $\mathbf{t}_i$ is the ground-truth labels of whether an object of the $c^\mathrm{th}$ category is presented in input image.
The object mining phase forms the WSOD problem as multiple-instance learning and implicitly classifies the region proposals with only image-level labels.
In this way, the model can correctly classify an image even only ``see'' a part of the object, and as a result, the proposal scores may not correctly indicate probabilities that proposals properly cover the entire objects of target categories.
To further reduce the mislocalization in instance refinement, we tweak a similar idea from instance refinement~\cite{Tang2017} and bounding-box regression~\cite{Yanga,Zeng2019} to reduce classification entropy and adjust object location.
To this end, instance refinement contains multiple branches, and each branch has proposal classification and bounding-box regression networks, which enables the refinement of both bounding-box scores and coordinates.
In details, it produces new classification scores ${S}^{r} \in \Re^{ n^\mathrm{pro} \times ( n^\mathrm{cat} + 1 ) }$ and bounding boxes ${B}^{r} \in \Re^{ n^\mathrm{pro} \times n^\mathrm{cat} \times 4 }$ for the $r^\mathrm{th}$ refinement head, where $n^\mathrm{cat} + 1$ denotes $n^\mathrm{cat}$ object categories and $1$ background.
During training, we utilize the detection results from the $(r-1)^\mathrm{th}$ branch to generate classification labels $\mathbf{t}^{r} \in \Re^{ n^\mathrm{pro}}$ and regression targets ${G}^{r} \in \Re^{ n^\mathrm{pro} \times 4 }$ for object proposals in the $r^\mathrm{th}$ branch.
Noted that we make use of the score matrix $X^\mathrm{s}$ from the object mining phase to supervise the first branch, \ie, $r=1$.
In detail, for the $c^\mathrm{th}$ category that $ \mathbf{t}_c = 1 $, we select all boxes whose class confidences from the previous predictions are greater than the pre-defined threshold, \ie, $0.5$, as pseudo ground truths.
Especially, if no box is selected, we seek the box with the highest score.
With the above pseudo-ground-truth boxes, we assign positive/negative labels for each proposal and form the training targets $\mathbf{t}^{r}$ and regression targets ${G}^{r}$ following Faster R-CNN~\cite{FASTERRCNN}.
Thus, the corresponding objective function is:
\begin{equation}
    \begin{aligned}
        {\mathcal{L}}_\mathrm{IR} = &
        \sum_{r=1}^{n^\mathrm{irf}} \sum_{p=1}^{n^\mathrm{pro}} \mathbf{y}_{\mathbf{t}^{r}_{p}} \mathcal{L}_\mathrm{CE} ({S}^{r}_p, \mathbf{t}^{r}_p) + \\
        & \mathds{1}[\mathbf{t}^{r}_p > 0] \mathbf{y}_{\mathbf{t}^{r}_{p}} \mathcal{L}_\mathrm{SL1} ({B}^{r}_{p,\mathbf{t}^{r}_{p}} , {G}^{r}_{p} )
    \end{aligned}
    ,
    \label{equ_IR_baseline}
\end{equation}
where $ n^\mathrm{irf} $ is the branch number in instance refinement
$L_\mathrm{CE}$ is the softmax cross-entropy loss, and $\mathcal{L}_\mathrm{SL1}$ is the smooth L1 loss~\cite{FASTRCNN}.
The indicator bracket indicator function $\mathds{1}[\mathbf{t}^{r}_p > 0]$ evaluates to $1$ if $\mathbf{t}^{r}_p > 0$ and $0$ otherwise.
Noted that $\mathbf{t}^{r}_p = 0$ denotes background category.
From another perspective of semi-supervised learning, the instance refinement exploits the noisy predictions from the teacher, \ie, object mining phase, and learn student detectors.
Thus, it works by selecting pseudo ground-truths from current predictions and learning a new detection branch with both classification and regression iteratively.
Proposal scores inferred from weak supervision are propagated to their spatially overlapped proposals to calibrate the classifiers.
In testing, the average output of all branches is used.
%%%%%%%%%%%%%%%%%%%%%%%%%%%%%%%%%%%%%%%%%%%%%%%%%%%%
\subsection{The \OurMethod Framework}
\label{sec_The_Overall_Framework}
Based on the above basic framework, we remove the external module of traditional object proposal and build learnable object proposal generators in a fully end-to-end manner.
Then we construct a network-embedded feature hierarchy between the backbone and WSOD head to handle large scale variations.
Finally, we propose a holistic self-training scheme to replace vanilla pseudo-labeling training.
The overall framework of the proposed \OurMethod is shown in Fig.~\ref{fig_framework}.
First, given an input image, conventional and strong augmentations $\tau$ are applied to it separately, and scale-invariant full-image feature maps are extracted from the backbone with a multi-rate resampling pyramid~(MRRP).
MRRP aggregates multi-scale contextual information at the top of the backbone, which leverages a set of different receptive fields to remedy scale-variation problem.
Second, self-supervised object proposal generator~(SSOPG) and autoencoder object proposal generator~(AEOPG) simultaneously predict a set of high-quality and high-confidence object proposals, which is followed by the RoIPool layer to generate proposal features.
SSOPG learns object candidate regions from the final predictions of \OurMethod, and AEOPG captures the salient objects in image by modeling low-rank approximation that retains important values and correlative relationships in full-image feature maps.
Third, the object mining phase outputs initial detection scores, and the multi-branch instance refinement is involved to refine both the scores and coordinates of proposals to bootstrap the quality of predicted results.
Thus, our model is a \textit{unified} WSOD network with strictly end-to-end training and inference, and is capable to achieve promising results even under fully/semi-supervised settings.
The overall loss function is:
\begin{equation}
    \mathcal{L} = \mathcal{L}_\mathrm{SSPOG} + \mathcal{L}_\mathrm{AEOPG} + \mathcal{L}_\mathrm{OM} + \mathcal{L}_\mathrm{IR}
    ,
    \label{eq_overall_loss}
\end{equation}
where $\mathcal{L}_\mathrm{SSOPG}$ and $\mathcal{L}_\mathrm{AEOPG}$ are the loss functions of the proposed SSOPG and AEOPG, which will be detailed in the following subsection.
SEM self-training provides a stepwise procedure to minimize the prediction entropy of instance refinement for pseudo-learning, which aims to consider the trade-off between precision and recall in different refinement branches.
And CCR self-training imposes consistency-constraint regularization to the predictions produced from stochastic augmentations of input images.
Thus, our scheme is a \textit{holistic} self-training framework for WSOD which incorporates ideas from the dominant paradigms of weakly supervised learning~\cite{Zhou2018c}, \ie, entropy minimization~\cite{Miyato2017,Berthelot2019}, consistency regularization~\cite{Xie2020c,Huang2020} and exponential moving average~\cite{Tarvainen2017,Laine2017}.
During inference, the proposed AEOPG, SEM, and CCR self-training are safely removed, and only MRRP and SSOPG remain with the basic WSOD network.
%%%%%%%%%%%%%%%%%%%%%%%%%%%%%%%%%%%%%%%%%%%%%
\subsection{Unified Weakly Supervised Detection Network}
\label{sec_OPG}
\subsubsection{Self-supervised object proposal generator~(SSOPG)}
\label{sec_SSOPG}
We propose an anchor-based self-supervised object proposal generator, which takes full-image feature maps as input and outputs a set of rectangular object proposals, each with an objectness score.
Generating anchors with the sliding window manner in feature maps has been widely adopted by anchor-based various detectors~\cite{FASTERRCNN,Cai2019}.
Anchors are regression references and classification candidates to predict object proposals.
SSOPG uses a small fully convolutional network to map each sliding window anchor to a low-dimensional feature, as in~\cite{FASTERRCNN}.
To this end, SSOPG has a $3\times3$ convolutional layer with $256$ channels followed by two siblings $1\times1$ convolutional layers for objectness classification and regression, respectively.
Formally, we denote $n^\mathrm{anc}$ as the number of anchors in each location, and $h$ and $w$ as the height and width of feature maps, respectively.
Thus, the regression layer has $4n^\mathrm{anc}$ outputs encoding the coordinates of $n^\mathrm{anc}$ boxes, and the objectness layer outputs $n^\mathrm{anc}$ scores that estimate the probability of the object for each proposal.
Given a feature maps with spatial size $( h \times w )$, SSOPG outputs object proposals $ B^\mathrm{pro} \in \Re^{ n^\mathrm{anc}hw \times 4 }$ and objectness scores $S^\mathrm{pro} \in \Re^{ n^\mathrm{anc}hw \times 1 }$.
We apply non-maximum suppression~(NMS) on $ B^\mathrm{pro} $ and keep the top-$ n^\mathrm{pro} $ object proposals.
We leverage self-supervised learning to train SSOPG with supervision created by \OurMethod without additional human effort or external modules.
Our intuition is that as WSOD can discover category-specific objects, it also has the potential to learn objectness instances.
To this end, we attach a new objectness detection branch to the proposal features with two sibling fully-connected layers to predict objectness classification $S^\mathrm{obn} \in \Re^{ n^\mathrm{pro}_\mathrm{} }$ and regression $ B^\mathrm{obn} \in \Re^{ n^\mathrm{pro}_\mathrm{} \times 4 }$, respectively.
Given the output from instance refinement phase, \ie, scores $S^{r} \in \Re^{ n^\mathrm{pro}_\mathrm{} \times n^\mathrm{cat} }$ and bounding boxes $ B^{r} \in \Re^{ n^\mathrm{pro}_\mathrm{} \times n^\mathrm{cat} \times 4 }$, we generate objectness pseudo-ground-truth boxes as $ \{ B^{r}_{p,c} | p = \argmax_{p}{S^{r}_{p,c}}, c =  \{ i | \mathbf{t}_i = 1 \} \}$, where $B^{r}_{p,c}$ and $S^{r}_{p,c}$ denote the $p^\mathrm{th}$ predicted bounding box and score for the $c^\mathrm{th}$ category.
For each object proposal $B^p$, we assign classification labels $\mathbf{t}^\mathrm{obn}_p$ and regression targets ${G}^\mathrm{obn}_p$ with an intersection-over-union~(IoU) ratio of $\lambda^\mathrm{obn}$.
Finally, we further select the top-$n^\mathrm{obn}_\mathrm{}$ predicted boxes from objectness detection branch as pseudo ground-truths to supervise proposal generator, where $n^\mathrm{obn}_\mathrm{} = | \{\mathbf{t}_i | \mathbf{t}_i=1 \} |$.
We label object proposals $B^p$ by an IoU ratio of $\lambda^\mathrm{pro}$ with ${G}^\mathrm{pro}$ to compute the classification labels $\mathbf{t}^\mathrm{pro}_p$ and regression targets ${G}^\mathrm{pro}_p$ for each object proposal $B^p$.
The overall loss function of SSOPG is:
\begin{equation}
    \begin{split}
        L_\mathrm{SSOPG} =
        & \sum^{n^\mathrm{obn}_\mathrm{}}_p L_\mathrm{CE} (S^\mathrm{obn}_p, \mathbf{t}^\mathrm{obn}_p) +  L_\mathrm{SL1} (B^\mathrm{obn}_p, {G}^\mathrm{obn}_p ) \\ +
        & \sum^{n^\mathrm{pro}_\mathrm{}}_p L_\mathrm{BCE} (S^\mathrm{pro}_p, \mathbf{t}^\mathrm{pro}_p) +  L_\mathrm{SL1} (B^\mathrm{pro}_p, {G}^\mathrm{pro}_p )
    \end{split}
    ,
\end{equation}
where $L_\mathrm{BCE}$ is the binary sigmoid cross-entropy loss.
%%%%%%%%%%%%%%%%%%%%%%%%%%%%%%%%%%%%%%%%%%%%%%%%%
%%
%%
\begin{figure}[t]
    \centering
    \begin{center}
        \begin{subfigure}[t]{0.48\textwidth}
            \begin{center}
                \includegraphics[width=1.0\textwidth]{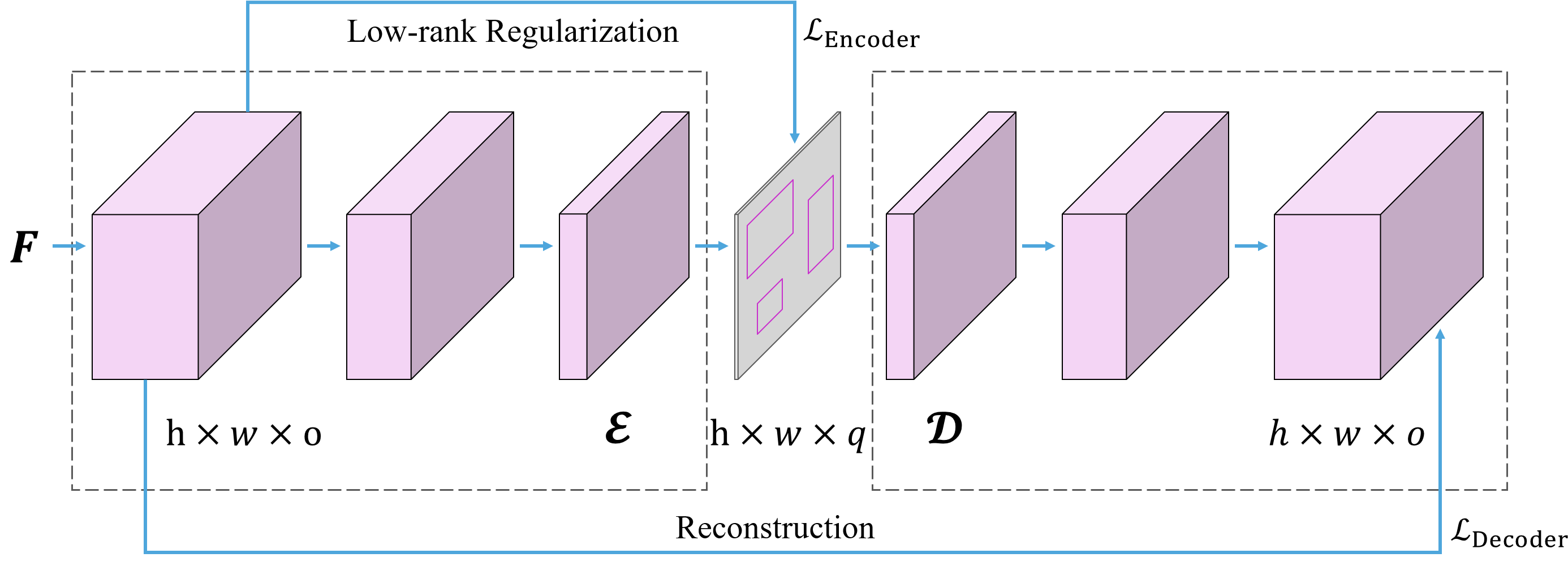}
            \end{center}
        \end{subfigure}
    \end{center}
    \caption{Autoencoder architecture in AEOPG.}
    \label{fig_aeopg}
\end{figure}
%%%%%%%%%%%%%%%%%%%%%%%%%%%%%%%%%%%%%%%%%%%%%%%%%
%%
%%
\subsubsection{Autoencoder object proposal generator~(AEOPG)}
\label{sec_AEOPG}
Without ground-truth-bounding-boxes information, SSOPG may not generate high-quality object proposals in the early training stage and reduce the learning efficiency of the subsequent object mining and instance refinement, which in turn provides low-quality supervision for SSOPG.
We observe that the output feature maps from self-supervised pre-trained backbones have high activations around the spatial locations of salient objects in images.
And we also assume that feature vectors at different spatial locations should have high correlations if they belong to the same category.
Thus, reducing the dimension of feature maps from the backbone by low-rank approximation can find a representation of the data that retains the important values and correlative relationship.
To this end, we propose an autoencoder architecture for object proposals, which takes full image feature maps as input and outputs high-quality proposals, as shown in Fig.~\ref{fig_aeopg}.
Like all autoencoders, AEOPG has an encoder $\mathcal{E}$ that project image feature maps $F \in \Re^{ h w \times o }$ from the backbone to a $q$-channel compressed representation, \ie, $\mathcal{E}(F) \in \Re^{ h w \times q }$, and a decoder $\mathcal{D}$ that reconstructs the original signal from the compressed representation.
We set $o \gg q$, and $h$ and $w$ as the height and width of feature maps, respectively.
To prevent the autoencoder from learning the identity function, we leverage matrix decomposition as low-rank regularization on latent representation to improve its ability to capture important salient objects in images.
The overall loss function $\mathcal{L}_\mathrm{AEOPG}$ is formulated as:
\begin{equation}
    \mathcal{L}_\mathrm{AEOPG} = \mathcal{L}_\mathrm{Encoder} + \mathcal{L}_\mathrm{Decoder},
    \label{eq_AEOPG}
\end{equation}
where $\mathcal{L}_\mathrm{Encoder}$ is the low-rank regularization loss and $\mathcal{L}_\mathrm{Decoder}$ is the reconstruction loss.
Concretely, we project image feature maps $F$ to the first $q$ principal components of low rank matrix $F^q \in \Re^{ h w \times q }$ via linear matrix decomposition algorithm, \eg, SVD, which is then binarized at a threshold of $0$.
Thus, the loss function $\mathcal{L}_\mathrm{Encoder}$ is formulated as:
\begin{equation}
    %\mathcal{L}_\mathrm{Encoder} = \frac{1}{HW}\sum^{HW}_{i} \mathcal{L}_\mathrm{BCE}(E(F)_i, L_{q}(F)_i)
    \mathcal{L}_\mathrm{Encoder} = \sum^{hw}_{i} \mathcal{L}_\mathrm{BCE}(\mathcal{E}(F)_i, F^q_i)
    .
    \label{eq_Encoder}
\end{equation}
The encoder $\mathcal{E}$ can be viewed as a form of non-linear dimensional reduction to find a low-dimensional embedding of full-image feature maps that preserves semantic correspondences across them, which is guided by linear low-rank decomposition.
Thus, the larger the absolute value in $\mathcal{E}(F)$ is, the higher the positive correlation will be.
Specifically, in the weakly-supervised learning scenario, the positive correlation indicates the salient characteristic in images.
Therefore, the value zero is used as a natural threshold for dividing $\mathcal{E}(F)$ and $F^q$ into binary segmentation maps.
To capture salient objects, we binarize the low-rank regularized feature maps $\mathcal{E}(F)$ at a threshold of $0$ and compute all connected regions.
To achieve this, two pixels are connected when they are neighbors and have the same value.
Then the tightest bounding boxes of connected regions are considered as object proposals for salient objects.
Different from SSOPG which relies on WSOD models itself in a self-supervision manner, AEOPG captures the salient objects in low-rank feature maps decomposed by an unsupervised encoder-decoder network.
Additionally, to constrain the quality of the compressed feature maps, we input the compressed feature maps into a decoder to reconstruct the original feature maps.
Thus, $\mathcal{L}_\mathrm{Decoder}$ loss computes the squared difference between the input and reconstructed feature maps.:
\begin{equation}
    %\mathcal{L}_\mathrm{Decoder} = \frac{1}{HW}\sum^{HW}_{i}\left\| F - D(E(F)) \right\|_2
    \mathcal{L}_\mathrm{Decoder} = \sum^{hw}_{i}\left\| F_i - \mathcal{D}(\mathcal{E}(F))_i \right\|_2
    .
    \label{eq_Decoder}
\end{equation}
We concatenate all object proposals from SSOPG and AEOPG to extract proposal features during training, which is the input of the subsequent object mining and instance refinement phases.
Note that we only use AEOPG to generate high-quality object proposals in the training stage, so this module does not affect the inference speed.
%%%%%%%%%%%%%%%%%%%%%%%%%%%%%%%%%%%%%%%%%%%%%%%%%%%%%%%%%%%%%%%
\subsubsection{Multi-rate resampling pyramid~(MRRP)}
\label{sec_MRRP}
\begin{figure}[t]
    \begin{center}
        \begin{subfigure}[t]{0.45\textwidth}
            \begin{center}
                \includegraphics[width=1.0\textwidth]{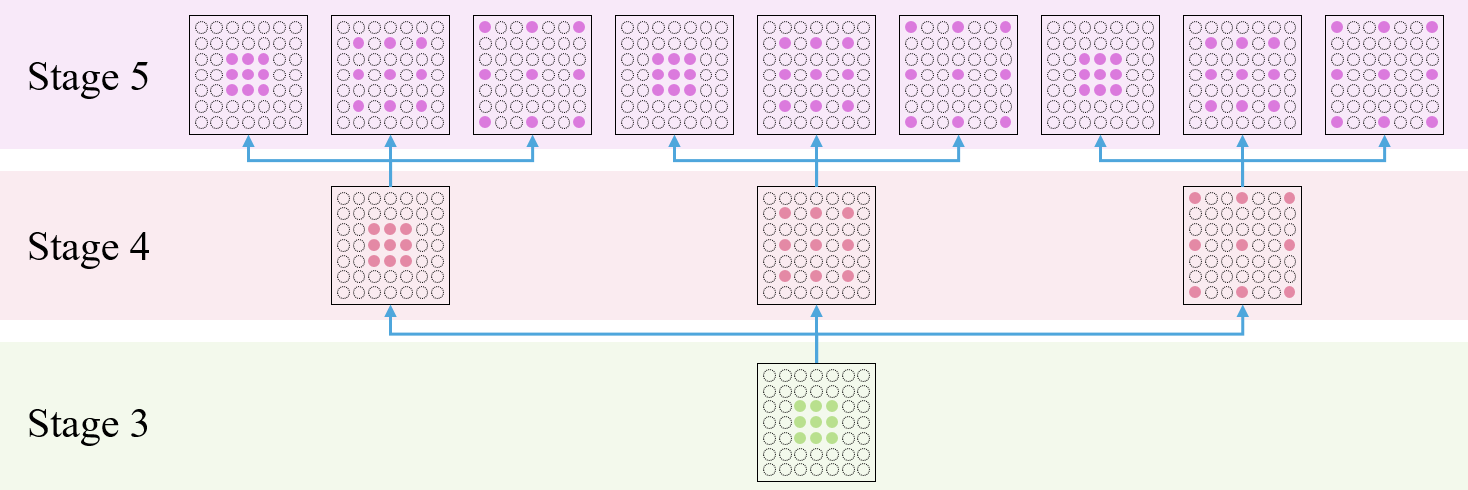}
            \end{center}
        \end{subfigure}
    \end{center}
    \caption{Illustration of MRRP on the $\protect 4^\mathrm{th}$ and $\protect 5^\mathrm{th}$ stages of backbone. We set $\protect n^{\mathrm{stg}} = 2$, $\protect n^{\mathrm{pls}} = 3$ and $\protect \alpha^{\mathrm{vdl}} = \{1, 2, 3\}$.}
    \label{fig_mrrp}
\end{figure}
%%%%%%%%%%%%%%%%%%%%%%%%%%%%%%%%%%
%%
%%
Aggregating multi-scale information is critical for detectors to exploit context and improve performance in large scale variations.
Existing WSOD methods leverage multi-scale image pyramids to remedy scale-variation problem.
However, it does not design specific network architecture to handle the large variation of scales.
Especially, we observe that directly employing an encode-decoder structure, \eg, U-Net~\cite{Ronneberger2015} and FPN~\cite{Lin2017a}, does not achieve competitive results in WSOD.
As it requires attaching new layers to build feature pyramids and may converge to an undesirable local minimum.
Inspired by spatial pyramid pooling~\cite{He2014} in fully supervised learning, we construct a multi-rate resampling pyramid~(MRRP) to aggregate multi-scale contextual information, in which each level shares the same parameters.
Our intuition is that integrating information from other receptive fields helps widen the scales, thus alleviating ambiguities and reducing information uncertainty in the local area.
Thus, we use a large range of receptive fields to describe objects at different scales.
Note that current backbone models commonly set receptive fields at the same size with a regular sampling grid on a feature map.
Therefore, we generalize trident block~\cite{Li2019g} to iteratively replicate $ n^\mathrm{pls} $ parallel streams for the last $n^\mathrm{stg}$ stages of backbones, which share the same structures and parameters, but with various dilation rates $ {\alpha}^\mathrm{vdr} = \{ {\alpha}^\mathrm{vdr}_{i}, \dots, {\alpha}^\mathrm{vdr}_{n^\mathrm{pls}} \} $.
Taking the $4^\mathrm{th}$ and $5^\mathrm{th}$ stages of backbone as an example in Fig.~\ref{fig_mrrp}, we first replicate the original $4^\mathrm{th}$ stage $ n^\mathrm{pls} $ times with various dilation rates, and for each output feature map we repeat the replicating operation in the $5^\mathrm{th}$ stage.
Finally, MRRP outputs $ { ( n^\mathrm{stg} ) }^{ n^\mathrm{pls} } $ feature maps, and $n^\mathrm{stg}$ is the number of MRRP stages.
% $\phi = \{ \phi_{i}, \dots, \phi_{{ ( n^\mathrm{stg} ) }^{ ( n^\mathrm{pls} ) }} \}$
%%

%%
We apply SSOPG and AEOPG on each feature map separately to generate object proposals.
The top $n^\mathrm{pro}_\mathrm{infer}$ proposals from SSOPG and all proposals from AEOPG are mapped to their feature maps to extract proposal features in the RoIPool layer.
Although it does not directly use the entire resampling pyramid, scale-invariant features are still distilled into the backbone by optimization of the shared parameters.
%%%%%%%%%%%%%%%%%%%%%%%%%%%%%%%%%%%%%%%%%%%
\subsection{Holistic Self-Training with Weak Supervision}
\label{sec_HST}
\subsubsection{Step-wise entropy minimization~(SEM)}
\label{sec_SEM}
Most WSOD methods only apply instance refinement~\cite{Tang2017} to rescore object proposals, which may result in low-quality bounding boxes predicted.
To explain, the bounding box predictions heavily rely on the quality of candidate boxes generated by object proposal algorithms, which limits further performance improvement.
Although recent methods~\cite{Gao2018,Zeng2019,Fang2020,Ren2020} integrate bounding-box regression in WSOD, they either require external modules or additional supervision and fail to consider the trade-off between precision and recall in different refinement branches.
To reduce mislocalizations, we propose a step-wise entropy minimization strategy, which progressively selects high-confidence object proposals as positive samples to refine both detection scores and coordinates.
Our intuition is that the former branches in instance refinement have large ambiguity of selecting positive and negative samples, as their pseudo ground truths are noisy and do not cover objects well.
Thus, we step-wisely learn instance refinement from low-quality to high-quality positive samples.
To this end, we first add bounding-box regression to each branch of vanilla classifier refinement~\cite{Tang2017}, which can fine-tune both scores and coordinates in all refinement branches.
We use a series of IoU thresholds $ {\Lambda} = \{ {\lambda}^{1}, \dots, {\lambda}^{n^\mathrm{irf}} \} $ to label positive and negative proposals and optimize each branch at separated IoU level, where $ n^\mathrm{irf} $ is the number of refinement branches.
Thus, the loss function of Eq.~\ref{equ_IR_baseline} is replaced to:
\begin{equation}
    \begin{aligned}
        {\mathcal{L}}_\mathrm{SEM} =
        & \sum_{r=1}^{n^\mathrm{irf}} \sum_{p=1}^{n^\mathrm{pro}}  \mathbf{y}_{\mathbf{t}^{r}_{p}} \mathcal{L}_\mathrm{CE} ({S}^{r}_p, \mathbf{t}^{r}_p({\lambda}^{r}) ) + \\
        & [\mathbf{t}^{r}_p({\lambda}^{r} ) > 0] \mathbf{y}_{\mathbf{t}^{r}_{p}} \mathcal{L}_\mathrm{SL1} ({B}^{r}_{p,\mathbf{t}^{r}_{p}} , {G}^{r}_{p}({\lambda}^{r}) )
    \end{aligned}
    ,
    \label{equ_IRr_SEM}
\end{equation}
where $\mathbf{t}^{r}_{p}({\lambda}^{r})$ and ${G}^{r}_{p}({\lambda}^{r})$ are the classification and regression targets for the $i^\mathrm{th}$ object proposal in the $r^\mathrm{th}$ branch under IoU threshold of ${\lambda}^{r}$, respectively.
We restrict $ {\Lambda} $ to be in ascending order, \ie, $\{ {\lambda}^{1} \le \dots \le {\lambda}^{n^\mathrm{irf}} \} $, which offers a good trade-off between precision and recall among refinement branches.
As the former branches establish a high-recall set of positive samples, while the successive branches receive high-precision positive samples.
Step-wise fashion guarantees a sequence of effective refinement branches of increasing quality.
As positive samples decrease quickly with $ \Lambda $, we sample each branch to keep a fixed proportion of positive and negative samples.
%%%%%%%%%%%%%%%%%%%%%%%%%%%%%%%%%%%%%%%%%%%%%%%%%%%%%%
\subsubsection{Consistency-constraint regularization~(CCR)}
Inspired by the effectiveness of consistency-based learning~\cite{Bachman2014}, we propose a consistency-constraint regularization scheme for WSOD.
Consistency-constraint regularization aims to enforce consistent predictions produced from stochastic augmentations, \ie, views, of the same input.
Concretely, we construct the same pseudo ground truths for different augmentations of the same images, thus imposing consistency-constraint regularization to the learning process.
In detail, each branch in instance refinement is supervised by the pseudo ground truths generated from the original images and the corresponding branch.
Thus, CCR objective function of the $r^\mathrm{th}$ instance refinement branch is:
\begin{equation}
    \begin{aligned}
        & {\mathcal{L}}_\mathrm{CCR}  = \sum_{\tau \in \mathcal{T}} \sum_{r=1}^{n^\mathrm{irf}}  \sum_{p=1}^{n^\mathrm{pro}} \mathbf{y}_{\mathbf{t}^{r}_{p}}(I) \mathcal{L}_\mathrm{CE} ({S}^{r}_p(\tau(I)), \mathbf{t}^{r}_p(I, {\lambda}^{r}) )  \\
        & + [\mathbf{t}^{r}_p(I, {\lambda}^{r} ) > 0] \mathbf{y}_{\mathbf{t}^{r}_{p}}(I) \mathcal{L}_\mathrm{SL1} ({B}^{r}_{p,\mathbf{t}^{r}_{p}}(\tau(I)) , {G}^{r}_{p}(I, {\lambda}^{r}) )
    \end{aligned}
    ,
    \label{equ_IRr_CCR}
\end{equation}
where $\mathbf{t}^{r}(I, {\lambda}^{r}) \in \Re^{ n^\mathrm{pro}}$ and ${G}^{r}(I, {\lambda}^{r}) \in \Re^{ n^\mathrm{pro} \times 4 }$ are the classification and regression targets of the original image $I$ in the $r^\mathrm{th}$ head under IoU threshold of ${\lambda}^{r}$, respectively.
Specifically, the transformation operation $\tau$ from a candidate set of augmentations $\mathcal{T}$ is applied on image $I$ to generate stochastic augmented images $\tau(I)$.
${S}^{r}_p(\tau(I))$ and ${B}^{r}_{p,\mathbf{t}^{r}_{p}}(\tau(I))$ denote the classification and regression predictions on augmented image $\tau(I)$.
Thus, final objective function ${\mathcal{L}}_\mathrm{IR}$ in Eq.~\ref{eq_overall_loss} combines Eqs.~\ref{equ_IRr_SEM} and~\ref{equ_IRr_CCR}:
\begin{equation}
    \begin{aligned}
        %{\mathcal{L}}_\mathrm{IR} =  \sum_{r=1}^{n^\mathrm{irf}} {{\mathcal{L}}^{r}_\mathrm{SEM} +\mathcal{L}}^{r}_\mathrm{CCR},
        {\mathcal{L}}_\mathrm{IR} =  {{\mathcal{L}}_\mathrm{SEM} +\mathcal{L}}_\mathrm{CCR}.
    \end{aligned}
    \label{equ_IR}
\end{equation}
Different from consistency-based semi-supervised object detection~\cite{Sohn2020} that relies on student-teacher-based siamese architecture, the proposed method imposes consistency-constraint regularization within a single network.
In other words, each branch is the preceding of the previous branch and the teacher of the succeeding branch.
For each branch, besides providing supervision signals to the succeeding one, we also explore propagating information to the preceding branch.
Therefore, we update branch weight $W^r$ from the succeeding branch weight $W^{r+1}$ with exponential moving average~(EMA).
At $i^\mathrm{th}$ iteration, we have
$W^r(i) =  \beta W^r(i-1) + (1 - \beta) W^{r+1}(i)$.
We perform this information propagation from the last branch of instance refinement to the detection branch of object mining.
Thereby, more accurate pseudo targets lead to a positive feedback loop between each consecutive branch, resulting in better accuracy.
Consequently, EMA weights transfer and share knowledge among different branches and lead to better generalization~\cite{Tarvainen2017}.
%%%%%%%%%%%%%%%%%%%%%%%%%%%%%%%%%%%%%%%%%%%%%%%%%%%%%%%%%%%%
\section{Quantitative Evaluations}
\label{sec_Quantitative_Evaluation}
%Organize and schedule your preparation for evaluation in the following aspects (each aspect is a subsection):
\subsection{Datasets}
We evaluate the proposed \OurMethod framework on PASCAL VOC 2007, PASCAL VOC 2012~\cite{PASCALVOC} and MS COCO 2014~\cite{MSCOCO} datasets, which are the widely-used benchmark.
PASCAL VOC 2007 consists of $5,011$ \emph{trainval} images, and $4,092$ \emph{test} images over $20$ categories.
PASCAL VOC 2012 consists of $11,540$ \emph{trainval} images, and $10,991$ \emph{test} images over $20$ categories.
Following the standard settings of WSOD, we use the \emph{trainval} set with only image-level labels for training.
MS COCO 2014 consists of $80$ object categories, which is among the most challenging datasets for instance segmentation and object detection.
Our experiments use $83$k \emph{train} set with image-level labels for training, and $41$k \emph{val} set for testing.
%%%%%%%%%%%%%%%%%%%%%%%%%%%%%%%%%%%%%%%%%%%%%%%%%%%%%%%%%%%%%%%%
\subsection{Evaluation Protocol}
Two evaluation protocols are used for PASCAL VOC: mean Average Precision~(\emph{m}AP) and Correct Localization~(CorLoc).
The \emph{m}AP follows PASCAL VOC protocol to report the mean AP at $50\%$ intersection-over-union~(IoU) of the detected boxes with the ground-truth ones.
CorLoc quantifies the localization performance by the percentage of images that contain at least one object instance with at least $50\%$ overlap to the ground-truth boxes.
The localization performance, \ie, CorLoc, is defined as predicting boxes when categories are known, which is evaluated on the \textit{train+val} set.
And detection performance, \ie, \emph{m}AP, is defined as predicting categories and boxes simultaneously, which is evaluated on the \textit{testing} set.
For MS COCO, we report COCO metrics, including AP at different IoU thresholds.
%%%%%%%%%%%%%%%%%%%%%%%%%%%%%%%%%%%%%%%%%%%%%%%%%%%%%
%%
\begin{table*}[t]
    \centering
    \caption{
        Ablation study of SSOPG, SEM and MRRP on PASCAL VOC 2007 in terms of CorLoc~(\%) and $m$AP~(\%).
    }
%    \footnotesize
    %    \begin{center}
        %        \begin{adjustbox}{max width=1.0\linewidth}
            \begin{tabular}{c|c|c|c|c|c|c|cc}
                \toprule

                \multirow{3}{*}{}&\multicolumn{2}{c}{SSOPG}&\multicolumn{2}{|c|}{SEM}&\multicolumn{2}{c|}{MRRP}&\multirow{3}{*}{CorLoc}&\multirow{3}{*}{$m$AP}\\

                \cmidrule{2-7}

                &$n^\mathrm{pro}_\mathrm{train}$&$n^\mathrm{pro}_\mathrm{infer}$&$n^\mathrm{irf}$&$\Lambda$&$ n^\mathrm{pls} $ & $ {\alpha}^\mathrm{vdr} $\\
                %==============================================================================================================
                \midrule

                %WSDDN~\cite{Bilen2016}&--&\multicolumn{1}{c}{--}&\multicolumn{1}{c}{--}&
                %53.5&34.8\\
                %ContextLocNet~\cite{Kantorov2016}&--&--&--&--&--&--&
                %55.1&36.3\\
                &--&--&--&--&--&--&
                55.1&36.3\\

                %\midrule

                a&--&--&--&--&--&--&
                55.7&36.7\\

                \midrule

                b&$1024$&$1024$&--&--&--&--&
                50.8&32.1\\

                c&$2048$&$2048$&--&--&--&--&
                50.9&32.3\\

                d&$4096$&$4096$&--&--&--&--&
                50.8&32.1\\

                \midrule

                e&$2048$&$2048$&$3$&$\{0.3, 0.4, 0.5\}$&--&--&
                60.5&41.8\\

                f&$2048$&$2048$&$4$&$\{0.35, 0.4, 0.45, 0.5\}$&--&--&
                60.8&42.1\\

                g&$2048$&$2048$&$5$&$\{0.3, 0.35, 0.4, 0.45, 0.5\}$&--&--&
                60.5&42.0\\

                \midrule

                %                h&$2048$&$2048$&$4$&$\{0.35, 0.4, 0.45, 0.5\}$&$2$&$\{ 1, 2 \}$&
                %                62.4&43.3\\
                %
                %                i&$2048$&$2048$&$4$&$\{0.35, 0.4, 0.45, 0.5\}$&$3$&$\{ 1, 2, 4 \}$&
                %                {62.6}&\textbf{44.0}\\
                %
                %                j&$2048$&$2048$&$4$&$\{0.35, 0.4, 0.45, 0.5\}$&$3$&$\{ 1, 2, 4 \}^*$&
                %                \textbf{63.0}&\textbf{44.0}\\
                %
                %                k&$2048$&$2048$&$4$&$\{0.35, 0.4, 0.45, 0.5\}$&$4$&$\{ 1, 2, 4, 8 \}$&
                %                62.8&43.6\\

                h&$2048$&$2048$&$4$&$\{0.35, 0.4, 0.45, 0.5\}$&$2$&$\{ 1, 2 \}$&
                62.3&43.2\\

                j&$2048$&$2048$&$4$&$\{0.35, 0.4, 0.45, 0.5\}$&$3$&$\{ 1, 2, 4 \}$&
                \textbf{63.0}&\textbf{44.0}\\

                k&$2048$&$2048$&$4$&$\{0.35, 0.4, 0.45, 0.5\}$&$4$&$\{ 1, 2, 4, 8 \}$&
                62.7&43.6\\

                %==============================================================================================================
                \bottomrule
            \end{tabular}
            %        \end{adjustbox}
        %    \end{center}
    \label{table_voc2007_abl}
\end{table*}
%%%%%%%%%%%%%%%%%%%%%%%%%%%%%%%%%%%%%%%%%%%%%%%%%%%%%%%%%%
%%
\subsection{Implementation Details}
We use VGG16~\cite{VGGNet} and WS-ResNet~\cite{Shen2020DRN} backbones, which is initialized with the weights pre-trained on ImageNet~\cite{ImageNet}.
We use synchronized SGD training on $8$ Tesla V100 with a total batch size of $8$.
We use a learning rate of $0.001$, a momentum of $0.9$, a dropout rate of $0.5$, a learning rate decay weight of $0.1$, and a step size of $140,000$ iterations.
The total number of training iterations is $200,000$.
We adopt $400,000$ training schedules for MS COCO.
In the first quarter of iterations, we only train the AEOPG and set the weights of other losses zeros.
In the multi-scale setting, we use scales ranging from $480$ to $1,216$ with stride $32$.
We randomly adjust the exposure and saturation of the images by up to a factor of $1.5$ in the HSV space.
A random crop with $0.9$ of the image sizes is applied.
Our proposed modules use the following parameter settings in all experiments unless specified otherwise.
We set $n^\mathrm{pro}$ to $2,048$ in SSOPG and the latent representation channels $q$ to $1$ in AEOPG.
We fix the number of anchors $n^\mathrm{anc}$ to $9$ with $3$ scales and $3$ aspect ratios following~\cite{FASTERRCNN}.
We set labeling threshold $ \lambda^\mathrm{obn} $ and $ \lambda^\mathrm{p} $ to $0.5$ and $0.7$, respectively.
For SEM, we set the number of refinement branches $ n^\mathrm{irf} $ to $4$, and $ {\Lambda} $ to $\{0.35, 0.4, 0.45, 0.5\}$.
We apply MRRP on the last stage of backbone with $ n^\mathrm{pls} = 3 $ and $ \mathbf{\alpha}^\mathrm{vdr} = \{ 1, 2, 4 \} $.
For CCR, we also randomly apply strong transforms from a set of candidate augmentations, \ie, color jittering, gray scale, gaussian blur, and CutOut, which follows the setting in~\cite{Liu2021a}.
We use default $\beta = 0.9996$ in EMA.
%%%%%%%%%%%%%%%%%%%%%%%%%%%%%%%%%%%%%%%%%%%%%%%%%%%%%%
\subsection{Ablation Study}
Without loss of generality, our ablation study is conducted on PASCAL VOC 2007 and ResNet18.
When tuning each group of hyper-parameters, others are kept as default.
%%%%%%%%%%%%%%%%%%%%%%%%%%%%%%%%%%%%%%%%%%%%%%%%%%%%%%%%%%%%
\subsubsection{Self-supervised object proposal generator}
The proposed SSOPG is designed to leverage the prediction of the WSOD model to learn object locations without relying on external proposal algorithms, such as selective search~\cite{Uijlings2013}, edge boxes~\cite{Zitnick2014} and MCG~\cite{APBMM2014}.
Thus we first build a strong baseline ContextLocNet~\cite{Kantorov2016}, which is widely used in recent WSOD methods.
Our implementation of ContextLocNet in the row~(a) of Tab.~\ref{table_voc2007_abl} achieves superior performance, which is due to larger mini-batch size and epochs.
In the rows~(b-d) of Tab.~\ref{table_voc2007_abl}, we vary the number of proposals in training and testing via hyper-parameters $n^\mathrm{pro}_\mathrm{train}$ and $n^\mathrm{pro}_\mathrm{infer}$.
We find that SSOPG is generally robust for a wide range of values between $[1,024, 4,096]$, which shows competitive performance compared to the baseline~(a).
We observe that directly replacing traditional proposals with SSOPG decreases about $5\%$ CorLoc and $4\%$ $m$AP.
However, the compromising performance enables end-to-end training of the entire model.
It also reveals that the existing WSOD models have the inherent characteristic of objectness localization, which has been ignored in the previous methods.
%%%%%%%%%%%%%%%%%%%%%%%%%%%%%%%%%%%%%%%%%%%%%%%%%%%
%%
\begin{table*}[t]
    \centering
    \caption{
        Ablation study of AEOPG on PASCAL VOC 2007 in terms of CorLoc~(\%) and $m$AP~(\%).
    }

    % \footnotesize
    % \begin{center}
    % \begin{adjustbox}{max width=1.0\linewidth}
    \begin{tabular}{c|c|c|c|c|c|cc}
        \toprule

        \multirow{3}{*}{}&\multicolumn{5}{c|}{AEOPG}&\multirow{3}{*}{CorLoc}&\multirow{3}{*}{$m$AP}\\

        \cmidrule{2-6}

        &$\mathcal{L}_\mathrm{Encoder}$&$\mathcal{L}_\mathrm{Decoder}$&$E$&$q$&$D$\\

        %==============================================================================================================
        \midrule

        &--&--&--&--&--&
        63.0&44.0\\

        \midrule

        a&\checkmark&--&$(C, 32, 2)$&$1$&-&
        67.2&48.4\\

        b&&\checkmark&$(C, 32, 2)$&$1$&$(32, C, 2)$&
        65.8&46.5\\

        c&\checkmark&\checkmark&$(C, 32, 2)$&$1$&$(32, C, 2)$&
        68.0&\textbf{50.2}\\

        \midrule

        d&\checkmark&\checkmark&$(C, 32, 2)$&$2$&$(32, C, 2)$&
        51.1&34.1\\

        e&\checkmark&\checkmark&$(C, 32, 2)$&$3$&$(32, C, 2)$&
        48.2&32.7\\

        f&\checkmark&\checkmark&$(C, 32, 2)$&$4$&$(32, C, 2)$&
        44.7&29.7\\

        \midrule

        g&\checkmark&\checkmark&$(C, 16, 2)$&$1$&$(16, C, 2)$&
        67.7&49.8\\

        h&\checkmark&\checkmark&$(C, 32, 4)$&$1$&$(32, C, 4)$&
        \textbf{68.1}&50.1\\

        i&\checkmark&\checkmark&$(C, 8, 4)$&$1$&$(8, C, 4)$&
        67.7&49.7\\

        %==============================================================================================================
        \bottomrule
    \end{tabular}
    % \end{adjustbox}
    % \end{center}
    \label{table_voc2007_abl_aeopg}
\end{table*}
%%%%%%%%%%%%%%%%%%%%%%%%%%%%%%%%%%%%%%%%%%%
%%
\subsubsection{Step-wise entropy minimization}
The proposed SEM is designed to progressively select high-confidence object proposals as positive samples to refine both detection scores and coordinates.
Specifically, SEM has two key hyper-parameters, \ie, $ n^\mathrm{f} $ and $ {\Lambda} $.
$ n^\mathrm{f} $ denotes the number of branches and $ {\Lambda} $ controls the smoothness of refinement.
We vary both hyper-parameters in the rows~(e-g) of Tab.~\ref{table_voc2007_abl} to discuss their impacts.
We first set $ n^\mathrm{f} $ to $3$ following the setting in the previous work~\cite{Tang2017}, and set IoU threshold $ {\Lambda} $ to $\{0.30, 0.40, 0.50\}$ in ascending order.
Row~(e) shows that SEM significantly achieves absolute gains of $9.5\%$ CorLoc and $9.5\%$ $m$AP over without SEM, \ie, \emph{vs.} row (c).
Compared to the vanilla instance refinement~\cite{Tang2017} that brings $6.4\%$ $m$AP improvement, the proposed SEM achieves larger performance-boosting, demonstrating the effectiveness of step-wise learning.
The benefits of SEM are mainly from:
First, the proposed step-wise learning paradigm progressively selects high-confidence object proposals as positive samples for refining.
Second, each refinement branch has a bounding-box regressor to refine bounding-box coordinates step-wisely.
Rows~(e-f) show that the models quickly saturate at $ n^\mathrm{f} = 4$, which is consistent with the results in~\cite{Tang2017}.
%%%%%%%%%%%%%%%%%%%%%%%%%%%%%%%%%%%%%%%%%%%%%%%%%%%
%%
\begin{table*}[!ht]
    \centering
    \caption{
        Ablation study of CCR on PASCAL VOC 2007 in terms of CorLoc~(\%) and $m$AP~(\%).
    }
    %\footnotesize
%    \begin{center}
%        \begin{adjustbox}{max width=1.0\linewidth}
            \begin{tabular}{c|c|c|c|c|c|c|cc}
                \toprule

                \multirow{3}{*}{}&\multicolumn{6}{c|}{CCR}&\multirow{3}{*}{CorLoc}&\multirow{3}{*}{$m$AP}\\

                \cmidrule{2-7}

                &$| \mathcal{T} |$&color jitter&gray scale&blur&cutout&$\beta$\\

                %==============================================================================================================
                \midrule

                &--&--&--&--&--&--&
                68.0&50.2\\

                \midrule

                a&2&\checkmark&--&--&--&--&
                70.6&52.2\\

                b&2&\checkmark&\checkmark&--&--&--&
                70.6&52.2\\

                c&2&\checkmark&\checkmark&\checkmark&--&--&
                70.8&52.7\\

                d&2&\checkmark&\checkmark&\checkmark&\checkmark&--&
                71.1&52.9\\

                \midrule

                e&3&\checkmark&\checkmark&\checkmark&\checkmark&--&
                71.2&52.6\\

                f&4&\checkmark&\checkmark&\checkmark&\checkmark&--&
                71.2&53.0\\

                g&5&\checkmark&\checkmark&\checkmark&\checkmark&--&
                71.0&52.4\\

                \midrule

                h&2&\checkmark&\checkmark&\checkmark&\checkmark&0.996&
                70.6&51.8   \\

                i&2&\checkmark&\checkmark&\checkmark&\checkmark&0.9996&
                \textbf{71.5}&\textbf{53.4}\\

                j&2&\checkmark&\checkmark&\checkmark&\checkmark&0.99996&
                71.2&53.3\\

                k&2&\checkmark&\checkmark&\checkmark&\checkmark&0.999996&
                71.2&52.6\\

                %==============================================================================================================
                \bottomrule
            \end{tabular}
%        \end{adjustbox}
%    \end{center}
    \label{table_voc2007_abl_crst}
\end{table*}
%%%%%%%%%%%%%%%%%%%%%%%%%%%%%%%%%%%%%%%%%%
%%
\subsubsection{Multi-rate resampling pyramid}
MRRP aims to aggregate multi-scale information for detectors to exploit context and achieve better performance in large scale variations.
We ablate two key hyper-parameters $n^\mathrm{pls}$ and $ {\alpha}^\mathrm{vdr}$ in Tab.~\ref{table_voc2007_abl}.
We also set $n^\mathrm{stg} = 1$ to only use the last stage of backbone to provide multi-scale information, as larger strides lead to a larger difference in receptive fields as needed~\cite{Li2019g}.
Row~(h) shows that the proposed MRRP further improves both the localization and detection performance with gains of $1.6 \%$ CorLoc and $1.2 \%$ $m$AP, respectively.
Note that the improvement of MRRP is stacked on the multi-scale training and testing, which already provide $3 \sim 4 \%$ absolute gains in CorLoc and $m$AP~\cite{Tang2018b}.
This demonstrates that MRRP has leveraged useful information from network-embedded feature hierarchy, which does not learn by multi-scale image pyramids.
Rows~(i-k) vary hyper-parameters $n^\mathrm{pls}$ and $ {\alpha}^\mathrm{vdr}$ with more scales.
The results show that the proposed MRRP further improves both the localization and detection performance robustly.
When dilation rates are too large, the performance tends to decrease slightly.
To explain, large receptive fields lead to inferior classification capacity by introducing redundant context for small objects.
%%%%%%%%%%%%%%%%%%%%%%%%%%%%%%%%%%%%%%%%%%%%%%%%%%%%%%%%%%%
%%
\begin{table*}[t]
    %\vspace{-10pt}
    \centering
    \caption{Comparison with the state-of-the-art methods on PASCAL VOC 2007 in terms of AP (\%) on \textit{test}.
    }
    %    \vspace{-10pt}
    %    \footnotesize
    \large
    \begin{center}
        \begin{adjustbox}{max width=1.0\linewidth}
            \begin{tabular}{l|c|cccccccccccccccccccc|c}
                \toprule

                \raisebox{+.5\height}{Method} & \raisebox{+.5\height}{Backbone} & 
                {\includegraphics{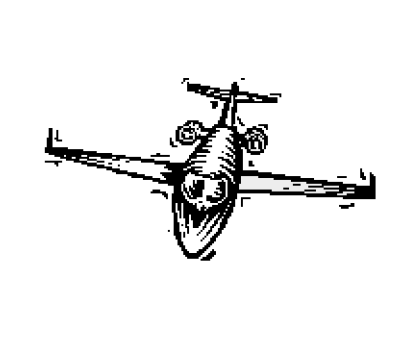}} &
                {\includegraphics{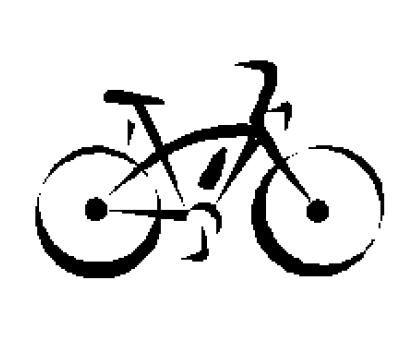}} &
                {\includegraphics{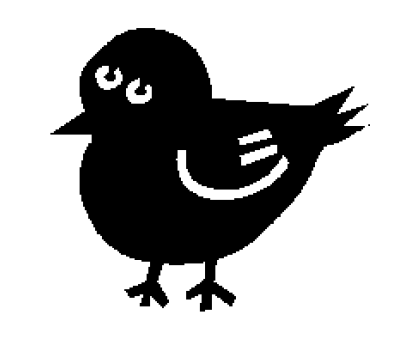}} &
                {\includegraphics{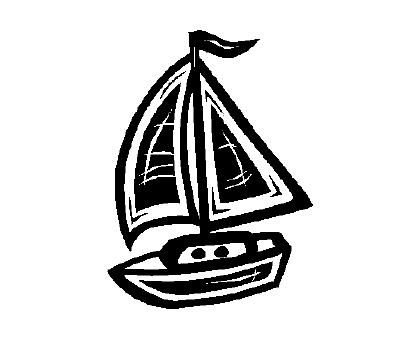}} &
                {\includegraphics{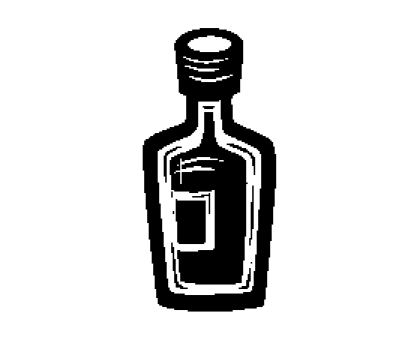}} &
                {\includegraphics{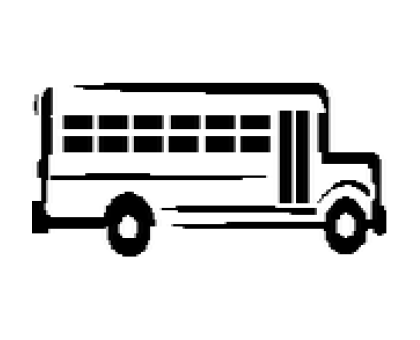}} &
                {\includegraphics{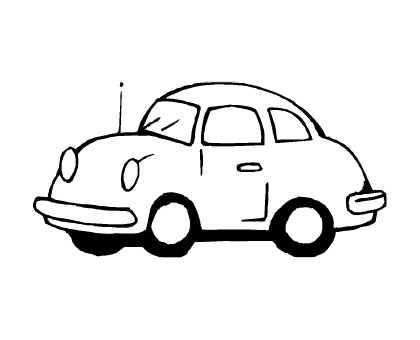}} &
                {\includegraphics{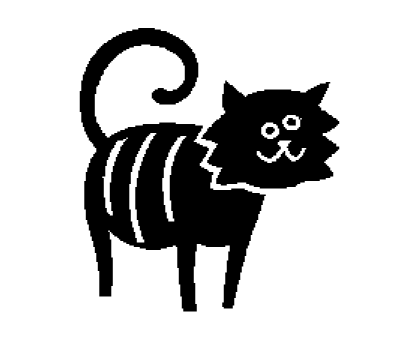}} &
                {\includegraphics{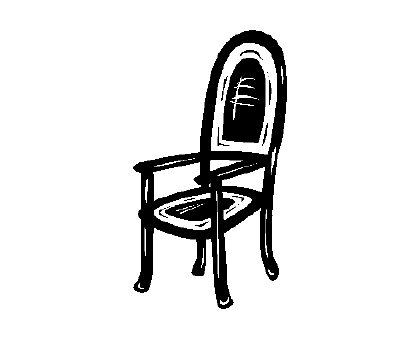}} &
                {\includegraphics{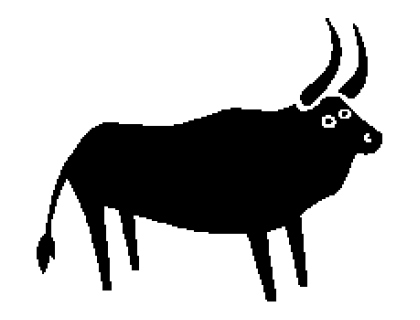}} &
                {\includegraphics{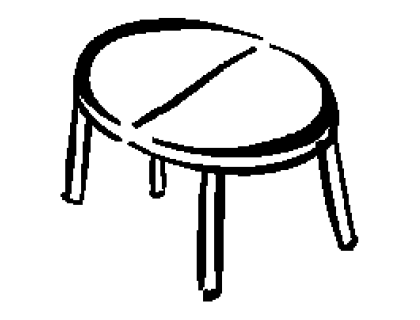}} &
                {\includegraphics{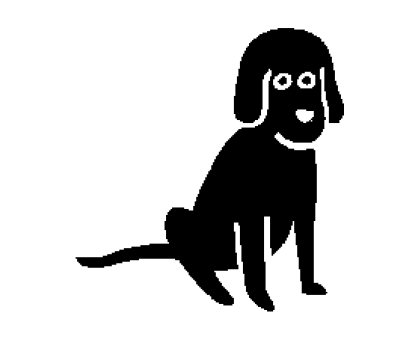}} &
                {\includegraphics{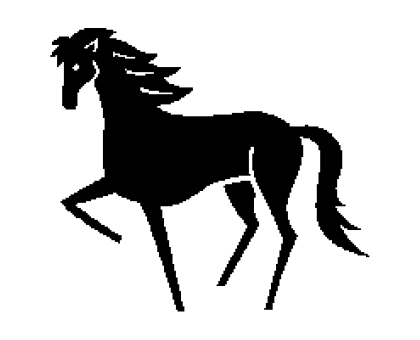}} &
                {\includegraphics{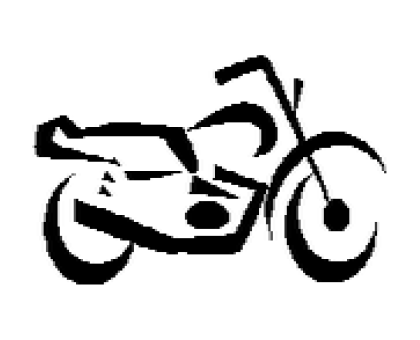}} &
                {\includegraphics{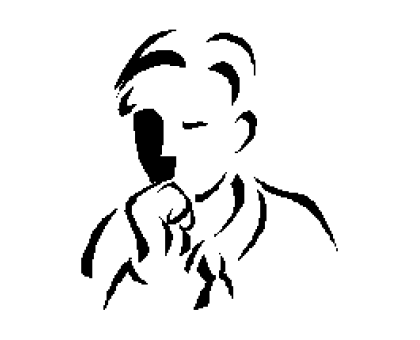}} &
                {\includegraphics{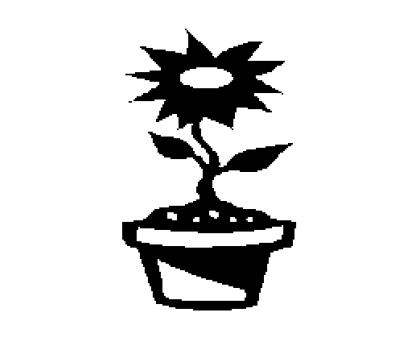}} &
                {\includegraphics{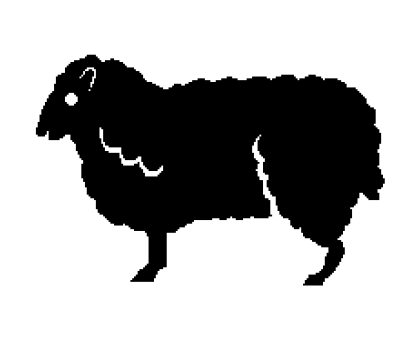}} &
                {\includegraphics{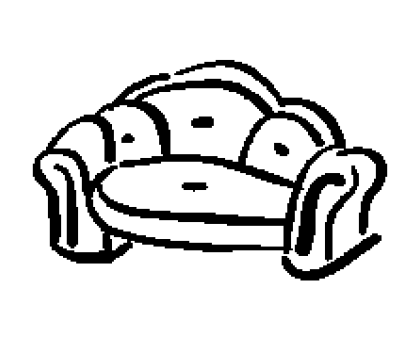}} &
                {\includegraphics{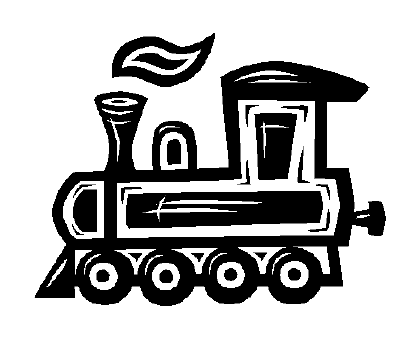}} &
                {\includegraphics{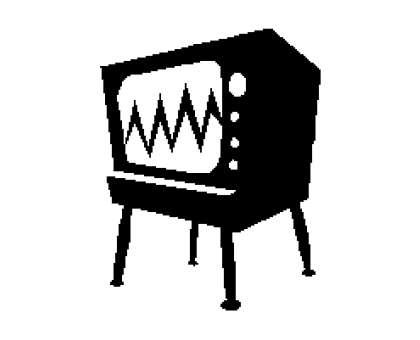}} & \raisebox{+.5\height}{Av.} \\
                
                \midrule
                \multicolumn{23}{c}{WSOD \textbf{with} external object proposal modules or additional data}\\
                \midrule%==============================================================================================================

                WSDDN\hfilll\cite{Bilen2016}&VGG16&39.4&50.1&31.5&16.3&12.6&64.5&42.8&42.6&10.1&35.7&24.9&38.2&34.4&55.6&9.4&14.7&30.2&40.7&54.7&46.9&34.8\\
                
                ContextLocNet\hfilll\cite{Kantorov2016}&VGG-F&57.1&52.0&31.5&7.6&11.5&55.0&53.1&34.1&1.7&33.1&49.2&42.0&47.3&56.6&15.3&12.8&24.8&48.9&44.4&47.8&36.3\\
                
                %Beam Search\hfilll\cite{Bency2016}&VGG16&-&-&-&-&-&-&-&-&-&-&-&-&-&-&-&-&-&-&-&-&25.7\\
                
                %OM+MIL\hfilll\cite{Li2016}&AlexNet&37.2&35.7&25.8&13.8&12.7&36.2&42.4&22.3&14.3&24.2&9.4&13.1&27.9&38.9&3.7&18.7&20.1&16.3&36.1&18.4&23.4\\
                
                %                OPG\hfilll\cite{Shen2018OPG}&AlexNet&32.5&43.8&20.0&11.6&9.1&42.9&21.0&49.2&4.5&15.2&30.6&29.0&68.9&26.8&9.8&12.4&18.2&28.9&45.7&20.5&27.0\\
                %                OPG\hfilll\cite{Shen2018OPG}&VGG16&48.9&49.9&25.6&14.0&6.1&47.1&22.5&52.7&3.4&19.6&33.2&33.3&55.3&30.2&9.9&9.1&14.8&25.8&50.8&24.7&28.8\\
                
                WCCN\hfilll\cite{Diba2017}&VGG16&49.5&60.6&38.6&29.2&16.2&70.8&56.9&42.5&10.9&44.1&29.9&42.2&47.9&64.1&13.8&23.5&45.9&54.1&60.8&54.5&42.8\\
                
                Jie~\etal\hfilll\cite{Jie2017}&VGG16&52.2&47.1&35.0&26.7&15.4&61.3&66.0&54.3&3.0&53.6&24.7&43.6&48.4&65.8&6.6&18.8&51.9&43.6&53.6&62.4&41.7 \\
                
                %SPAM-CAM\hfilll\cite{Gudi2017}&VGG16&-&-&-&-&-&-&-&-&-&-&-&-&-&-&-&-&-&-&-&-&27.5\\
                
                TST\hfilll\cite{Shi2017a}&AlexNet&-&-&-&-&-&-&-&-&-&-&-&-&-&-&-&-&-&-&-&-&33.8\\
                
                %                SGWSOD\hfilll\cite{Lai2017a}&VGG-F&45.9&59.6&26.4&24.7&11.4&61.2&56.5&49.3&4.9&35.6&24.1&45.2&56.0&56.5&22.7&19.8&34.7&44.7&50.1&48.3&38.9\\
                %                SGWSOD\hfilll\cite{Lai2017a}&VGG-M-1024&45.8&56.1&29.1&26.4&10.5&63.1&59.0&50.3&7.1&34.7&31.4&37.0&49.6&60.1&20.2&17.0&41.3&45.4&51.7&51.7&39.4\\
%                SGWSOD\hfilll\cite{Lai2017a}&VGG16&48.4&61.5&33.3&30.0&15.3&72.4&62.4&59.1&10.9&42.3&34.3&53.1&48.4&65.0&20.5&16.6&40.6&46.5&54.6&55.1&43.5\\
                
                TS$^C2$\hfilll\cite{Wei2018TS2C}&VGG16&59.3&57.5&43.7&27.3&13.5&63.9&61.7&59.9&24.1&46.9&36.7&45.6&39.9&62.6&10.3&23.6&41.7&52.4&58.7&56.6&44.3\\

                OICR\hfilll\cite{Tang2017}&VGG16&58.0&62.4&31.1&19.4&13.0&65.1&62.2&28.4&24.8&44.7&30.6&25.3&37.8&65.5&15.7&24.1&41.7&46.9&64.3&62.6&41.2\\
                
                %K-EM\hfilll\cite{Yan2017}&VGG16&59.8&64.6&47.8&28.8&21.4&67.7&70.3&61.2&17.2&51.5&34.0&42.3&48.8&65.9&9.3&21.1&53.6&51.4&54.7&50.7&46.1\\
                
                MELM\hfilll\cite{Wan}&VGG16&55.6&66.9&34.2&29.1&16.4&68.8&68.1&43.0&25.0&65.6&45.3&53.2&49.6&68.6&2.0&25.4&52.5&56.8&62.1&57.1&47.3\\
                
                ZLDN\hfilll\cite{Zhang2018b}&VGG16&55.4&68.5&50.1&16.8&20.8&62.7&66.8&56.5&2.1&57.8&47.5&40.1&69.7&68.2&21.6&27.2&53.4&56.1&52.5&58.2&47.6\\
                
                %                GAL-fWSD300\hfilll\cite{Shen2018GAL}&VGG16&52.0&60.5&44.6&26.1&20.6&63.1&66.2&65.3&15.0&50.1&52.8&56.7&21.3&63.4&36.8&22.7&47.9&51.7&68.9&54.1&47.0\\
%                GAL-fWSD512\hfilll\cite{Shen2018GAL}&VGG16&58.4&63.8&45.8&24.0&22.7&67.7&65.7&58.9&15.0&58.1&47.0&53.7&23.8&64.3&36.2&22.3&46.7&50.3&70.8&55.1&47.5\\
                
                %                ML-LocNet\hfilll\cite{Zhang2018g}&VGG-S&57.0&64.0&42.6&22.1&17.9&59.3&64.0&39.5&2.2&47.6&55.0&38.9&66.4&68.1&30.6&23.5&43.2&44.1&55.4&46.3&44.4\\
                %                ML-LocNet\hfilll\cite{Zhang2018g}&VGG-M&57.2&64.6&44.5&26.3&21.1&65.7&67.1&56.4&16.1&51.7&50.5&37.2&64.2&69.4&24.5&25.4&51.9&51.3&56.9&42.2&47.2\\
%                ML-LocNet\hfilll\cite{Zhang2018g}&VGG16&59.3&68.9&45.7&29.0&24.5&64.8&68.4&59.3&18.6&49.1&50.2&43.1&65.8&70.2&19.9&24.3&48.1&54.2&62.8&41.8&48.4\\
                
                WSRPN\hfilll\cite{Tang}&VGG16&57.9&70.5&37.8&5.7&21.0&66.1&69.2&59.4&3.4&57.1&57.3&35.2&64.2&68.6&32.8&28.6&50.8&49.5&41.1&30.0&45.3\\
                
                PCL\hfilll\cite{Tang2018b}&VGG16&54.4&69.0&39.3&19.2&15.7&62.9&64.4&30.0&25.1&52.5&44.4&19.6&39.3&67.7&17.8&22.9&46.6&57.5&58.6&63.0&43.5\\
                
                Kosugi~\etal\hfilll\cite{Kosugi2019}&VGG16&61.5&64.8&43.7&26.4&17.1&67.4&62.4&67.8&25.4&51.0&33.7&47.6&51.2&65.2&19.3&24.4&44.6&54.1&65.6&59.5&47.6\\
                
                %                C-MIL\hfilll\cite{Wan2019}&VGG-F&54.5&55.5&34.4&20.3&16.7&53.4&59.2&44.6&8.4&46.0&40.2&40.8&47.7&63.2&22.8&23.2&39.4&44.3&53.8&52.3&40.7\\
                C-MIL\hfilll\cite{Wan2019}&VGG16&62.5&58.4&49.5&32.1&19.8&70.5&66.1&63.4&20.0&60.5&52.9&53.5&57.4&68.9&8.4&24.6&51.8&58.7&66.7&63.5&50.5\\
                
                Pred Net\hfilll\cite{Arun2018a}&VGG16&66.7&69.5&52.8&31.4&24.7&74.5&74.1&67.3&14.6&53.0&46.1&52.9&69.9&70.8&18.5&28.4&54.6&60.7&67.1&60.4&52.9\\
                
                %                WSDDN + W-RPN\hfilll\cite{Singh}&VGG16&55.9&52.6&27.4&20.7&7.8&63.6&54.8&55.7&4.9&37.6&35.6&59.4&52.0&54.8&19.6&12.9&31.9&44.2&57.4&39.2&39.4\\
                OICR W-RPN\hfilll\cite{Singh}&VGG16&-&-&-&-&-&-&-&-&-&-&-&-&-&-&-&-&-&-&-&-&46.9\\
                
%                SDCN\hfilll\cite{Li2019a}&VGG16&59.8&67.1&32.0&34.7&22.8&67.1&63.8&67.9&22.5&48.9&47.8&60.5&51.7&65.2&11.8&20.6&42.1&54.7&60.8&64.3&48.3\\
                
                %Sona~\etal\hfilll\cite{Sona}&VGG16&62.1&55.7&42.0&31.1&17.2&67.6&65.2&50.8&20.4&51.5&36.3&34.1&46.2&65.8&12.3&21.9&48.8&55.4&60.2&65.7&45.4\\
                
                WSOD$^2$\hfilll\cite{Zeng2019}&VGG16&65.1&64.8&57.2&39.2&24.3&69.8&66.2&61.0&29.8&64.6&42.5&60.1&71.2&70.7&21.9&28.1&58.6&59.7&52.2&64.8&53.6\\
                
                GAM+REG\hfilll\cite{Yanga}&VGG16&55.2&66.5&40.1&31.1&16.9&69.8&64.3&67.8&27.8&52.9&47.0&33.0&60.8&64.4&13.8&26.0&44.0&55.7&68.9&65.5&48.6\\
                %                PCL+GAM+REG\hfilll\cite{Yanga}&VGG16&57.6&70.8&50.7&28.3&27.2&72.5&69.1&65.0&26.9&64.5&47.4&47.7&53.5&66.9&13.7&29.3&56.0&54.9&63.4&65.2&51.5\\
                
                C-MIDN\hfilll\cite{CMIDN}&VGG16&53.3&71.5&49.8&26.1&20.3&70.3&69.9&68.3&28.7&65.3&45.1&64.6&58.0&71.2&20.0&27.5&54.9&54.9&69.4&63.5&52.6\\
                
                %                OIM\hfilll\cite{Lin2020}&VGG16&62.2&67.2&48.0&29.6&23.5&68.7&69.3&64.3&22.8&59.6&39.6&30.7&42.7&69.8&3.1&23.3&57.9&55.4&63.4&63.5&48.2\\
%                OIM+IR\hfilll\cite{Lin2020}&VGG16&55.6&67.0&45.8&27.9&21.1&69.0&68.3&70.5&21.3&60.2&40.3&54.5&56.5&70.1&12.5&25.0&52.9&55.2&65.0&63.7&50.1\\
                
                Ren~\etal\hfilll\cite{Ren2020}&VGG16&68.8&77.7&57.0&27.7&28.9&69.1&74.5&67.0&32.1&73.2&48.1&45.2&54.4&73.7&35.0&29.3&64.1&53.8&65.3&65.2&54.9\\
                
%                Zeni~\etal\hfilll\cite{Zeni2020}&VGG16&68.6&62.4&55.5&27.2&21.4&71.1&71.6&56.7&24.7&60.3&47.4&56.1&46.4&69.2&2.7&22.9&41.5&47.7&71.1&69.8&49.7\\
                
                PG-PS\hfilll\cite{Cheng2020}&VGG16&63.0&64.4&50.1&27.5&17.1&70.6&66.0&71.1&25.8&55.9&43.2&62.7&65.9&64.1&10.2&22.5&48.1&53.8&72.2&67.4&51.1\\
                
%                IM-CFB\hfilll\cite{Yin2021}&VGG16&64.1&74.6&44.7&29.4&26.9&73.3&72.0&71.2&28.1&66.7&48.1&63.8&55.5&68.3&17.8&27.7&54.4&62.7&70.5&66.6&54.3\\
                
%                P-MIDN-MGSC\hfilll\cite{Xu2021}&VGG16&-&-&-&-&-&-&-&-&-&-&-&-&-&-&-&-&-&-&-&-&53.9\\
                
                Zhang~\etal\hfilll\cite{Zhang2020l}&VGG16&62.2&61.1&51.1&33.8&18.0&66.7&66.5&65.0&18.5&59.4&44.8&60.9&65.6&66.9&24.7&26.0&51.0&53.2&66.0&62.2&51.2\\
                
                %                CASD\hfilll\cite{Huang2020}&VGG16&-&-&-&-&-&-&-&-&-&-&-&-&-&-&-&-&-&-&-&-&56.8\\
                
%                Yang~\etal\hfilll\cite{Yang2020a}&VGG16&59.4&66.4&45.8&21.5&22.1&70.1&67.3&66.1&24.2&58.8&48.5&60.5&62.4&66.7&17.9&26.0&47.5&57.5&60.5&63.5&50.6\\
                
                SLV\hfilll\cite{Chen2020}&VGG16&65.6&71.4&49.0&37.1&24.6&69.6&70.3&70.6&30.8&63.1&36.0&61.4&65.3&68.4&12.4&29.9&52.4&60.0&67.6&64.5&53.5\\
                
%                PML-LocNet\hfilll\cite{Zhang2020a}&VGG16&60.0&69.8&47.0&29.4&24.0&64.1&67.6&57.1&10.2&50.5&54.7&47.3&67.9&68.7&27.5&24.4&50.5&55.7&59.2&42.3&48.9\\
                
                OICR\hfilll\cite{Tang2017}&WSR18&61.3&54.5&52.4&30.1&34.9&68.9&65.0&75.0&22.5&57.4&19.7&66.6&64.8&64.9&16.8&22.3&53.2&54.9&69.9&64.8&51.0\\
                PCL\hfilll\cite{Tang2018b}&WSR18&54.4&69.5&48.7&29.7&33.2&70.7&69.7&57.2&11.5&62.4&37.2&39.3&66.3&67.5&23.7&30.9&60.1&52.0&65.3&55.3&50.2\\
                C-MIL\hfilll\cite{Wan2019}&WSR18&57.0&54.9&43.6&39.9&32.2&70.9&69.8&75.2&14.2&59.9&28.5&66.3&67.5&65.3&37.6&21.8&56.7&49.8&71.1&68.9&52.6\\
                GAM+REG\hfilll\cite{Yanga}&WSR101&67.3&72.1&55.8&31.8&31.3&71.6&70.0&76.7&19.4&58.7&21.1&68.5&74.6&69.9&19.1&18.8&48.4&55.1&71.9&53.2&52.8\\
                
                %==============================================================================================================
                \midrule
                \multirow{3}{*}{\OurMethodx}
                &VGG16&57.2&75.6&53.8&31.1&38.6&76.7&74.0&69.7&3.4&73.6&47.9&72.1&75.8&74.1&8.5&27.4&59.2&50.3&71.6&70.2&55.5\\
                &WSR18&69.0&67.3&62.0&42.3&44.8&77.8&75.8&76.3&18.5&75.5&41.3&78.3&75.9&75.1&22.7&26.7&56.1&60.8&74.7&68.4&\textbf{59.5}\\
%                &WSR18&66.5&68.8&59.6&35.9&40.3&73.7&74.7&76.7&25.7&72.9&41.2&77.7&77.7&74.8&22.2&26.1&56.4&51.2&74.1&63.7&58.0\\
                &WSR50&67.3&61.1&63.5&46.4&38.1&74.6&76.2&80.8&7.0&71.7&37.5&77.0&76.8&72.7&8.9&22.5&53.4&67.9&75.7&63.8&57.2\\
                %&ResNet101&\\ %TODO

                \midrule
                \multicolumn{23}{c}{WSOD \textbf{without} external object proposal modules or additional data}\\
                \midrule%==============================================================================================================
                
                Beam Search\hfilll\cite{Bency2016}&VGG16&-&-&-&-&-&-&-&-&-&-&-&-&-&-&-&-&-&-&-&-&25.7\\
                
                OM+MIL\hfilll\cite{Li2016}&AlexNet&37.2&35.7&25.8&13.8&12.7&36.2&42.4&22.3&14.3&24.2&9.4&13.1&27.9&38.9&3.7&18.7&20.1&16.3&36.1&18.4&23.4\\
                
                %                OPG\hfilll\cite{Shen2018OPG}&AlexNet&32.5&43.8&20.0&11.6&9.1&42.9&21.0&49.2&4.5&15.2&30.6&29.0&68.9&26.8&9.8&12.4&18.2&28.9&45.7&20.5&27.0\\
                OPG\hfilll\cite{Shen2018OPG}&VGG16&48.9&49.9&25.6&14.0&6.1&47.1&22.5&52.7&3.4&19.6&33.2&33.3&55.3&30.2&9.9&9.1&14.8&25.8&50.8&24.7&28.8\\
                
                SPAM-CAM\hfilll\cite{Gudi2017}&VGG16&-&-&-&-&-&-&-&-&-&-&-&-&-&-&-&-&-&-&-&-&27.5\\
                
                %==============================================================================================================
                %                \midrule
                %                \multirow{3}{*}{UWSOD}
                %                &VGG16&57.7&72.7&46.4&24.3&11.2&60.4&72.3&29.2&14.6&58.7&29.1&59.4&72.6&68.6&1.4&23.7&35.6&40.3&51.8&49.8&44.0\\
                %                &WSR18&59.1&68.9&47.1&18.3&28.2&65.1&72.3&31.4&7.6&62.1&44.9&46.6&68.1&70.4&18.9&23.5&49.9&40.5&66.2&10.5&45.0\\
                %                &ResNet50&\\ %TODO
                %                %&ResNet101&\\ %TODO
                
                \midrule
                \multirow{3}{*}{\OurMethod}
                &VGG16&63.23&77.08&52.49&38.74&21.33&74.90&72.20&26.62&6.91&72.25&38.41&41.68&79.10&78.63&14.65&29.70&52.55&48.22&67.26&58.7&50.7\\
                &WSR18&69.7&77.8&55.2&37.2&36.1&73.7&72.6&44.3&8.7&69.8&48.1&61.6&76.1&73.1&24.7&27.1&54.6&57.9&68.8&30.6&{53.4}\\
                &WSR50&70.2&67.7&62.0&45.7&33.9&73.4&74.0&85.2&3.3&68.1&50.4&79.6&77.3&72.0&12.5&10.1&58.1&62.4&70.2&15.5&\textbf{54.6}\\

                \midrule
                \multicolumn{23}{c}{FSOD}\\
                \midrule
                
                Fast RCNN\hfilll\cite{FASTRCNN}&VGG16&74.5&78.3&69.2&53.2&36.6&77.3&78.2&82.0&40.7&72.7&67.9&79.6&79.2&73.0&69.0&30.1&65.4&70.2&75.8&65.8&66.9\\
                Faster RCNN\hfilll\cite{FASTERRCNN}&VGG16&70.0&80.6&70.1&57.3&49.9&78.2&80.4&82.0&52.2&75.3&67.2&80.3&79.8&75.0&76.3&39.1&68.3&67.3&81.1&67.6&69.9\\
                
                %                \midrule%==============================================================================================================
                %                \multicolumn{23}{c}{WSOD with Cls-agnostic GT-bbox Known}\\
                %                \midrule
                %                
                %                GAM + REG\hfilll\cite{Yanga}&VGG16&66.1&64.0&55.6&25.3&37.2&75.6&69.9&53.6&17.4&63.2&62.2&49.0&65.1&69.3&50.8&18.9&52.1&54.7&69.5&66.2&54.3\\
                %                
                %                Ren~\etal\hfilll\cite{Ren2020}&VGG16&70.2&76.1&57.7&46.2&42.3&73.8&72.0&78.5&35.3&69.4&46.0&72.1&76.5&72.4&64.7&30.5&65.1&59.2&69.3&66.0&62.2\\
                %                
                %                
                %                \midrule
                %                \multirow{2}{*}{UWSOD}
                %                &VGG16&{74.0}&77.6&{66.2}&{57.8}&{51.0}&{78.0}&{74.7}&{81.5}&{43.2}&70.5&{67.6}&{77.8}&{81.2}&71.1&{67.7}&{36.7}&59.0&{66.1}&{78.6}&{72.0}&{67.7}\\
                %                &WSR18&73.2&77.6&69.6&64.4&50.7&81.4&77.7&83.9&44.1&71.0&69.5&79.3&81.6&72.3&70.0&42.0&63.4&69.1&82.2&70.9&69.7\\
                %                %&ResNet50&\\ %TODO
                %                %&ResNet101&\\ %TODO
                
                %==============================================================================================================
                \bottomrule
            \end{tabular}
        \end{adjustbox}
    \end{center}
    %    \vspace{-20pt}
    \label{table_voc2007_sota_map}
\end{table*}
%%%%%%%%%%%%%%%%%%%%%%%%%%%%%%%%%%
%%
\begin{table*}[t]
    \caption{Comparison with the state-of-the-art methods on PASCAL VOC 2007 in terms of CorLoc (\%) on \textit{train+val}.}
    %    \vspace{-10pt}
    %    \footnotesize
    \large
    \begin{center}
        \begin{adjustbox}{max width=1.0\linewidth}
            \begin{tabular}{l|c|cccccccccccccccccccc|c}
                \toprule             
                %                Method&Backbone&aero&bicy&bird&boa&bot&bus&car&cat&cha&cow&dtab&dog&hors&mbik&pers&plnt&she&sofa&trai&tv&Av.\\
                
                \raisebox{+.5\height}{Method} & \raisebox{+.5\height}{Backbone} & 
                {\includegraphics{main/figure/pascalvoc/airplane.png}} &
                {\includegraphics{main/figure/pascalvoc/bicycle.png}} &
                {\includegraphics{main/figure/pascalvoc/bird.png}} &
                {\includegraphics{main/figure/pascalvoc/boat.png}} &
                {\includegraphics{main/figure/pascalvoc/bottle.png}} &
                {\includegraphics{main/figure/pascalvoc/bus.png}} &
                {\includegraphics{main/figure/pascalvoc/car.png}} &
                {\includegraphics{main/figure/pascalvoc/cat.png}} &
                {\includegraphics{main/figure/pascalvoc/chair.png}} &
                {\includegraphics{main/figure/pascalvoc/cow.png}} &
                {\includegraphics{main/figure/pascalvoc/table.png}} &
                {\includegraphics{main/figure/pascalvoc/dog.png}} &
                {\includegraphics{main/figure/pascalvoc/horse.png}} &
                {\includegraphics{main/figure/pascalvoc/motorbike.png}} &
                {\includegraphics{main/figure/pascalvoc/person.png}} &
                {\includegraphics{main/figure/pascalvoc/pottedplant.png}} &
                {\includegraphics{main/figure/pascalvoc/sheep.png}} &
                {\includegraphics{main/figure/pascalvoc/sofa.png}} &
                {\includegraphics{main/figure/pascalvoc/train.png}} &
                {\includegraphics{main/figure/pascalvoc/tv.png}} & \raisebox{+.5\height}{Av.} \\

                \midrule
                \multicolumn{23}{c}{WSOD \textbf{with} external object proposal modules or additional data}\\
                \midrule%==============================================================================================================

                WSDDN\hfilll\cite{Bilen2016}&VGG16&65.1&58.8&58.5&33.1&39.8&68.3&60.2&59.6&34.8&64.5&30.5&43.0&56.8&82.4&25.5&41.6&61.5&55.9&65.9&63.7&53.5\\
                
                ContextLocNet\hfilll\cite{Kantorov2016}&VGG-F&83.3&68.6&54.7&23.4&18.3&73.6&74.1&54.1&8.6&65.1&47.1&59.5&67.0&83.5&35.3&39.9&67.0&49.7&63.5&65.2&55.1\\
                
                %                OM + MIL\hfilll\cite{Li2016}&AlexNet&64.3&54.3&42.7&22.7&34.4&58.1&74.3&36.2&24.3&50.4&11.0&29.2&50.5&66.1&11.3&42.9&39.6&18.3&54.0&39.8&41.2\\
                
                %                OPG\hfilll\cite{Shen2018OPG}&AlexNet&44.5&33.7&36.1&16.6&7.0&38.2&42.0&57.6&8.5&30.5&29.0&43.2&55.4&56.0&27.6&18.8&36.5&38.9&49.4&14.5&34.2\\
                %                OPG\hfilll\cite{Shen2018OPG}&VGG16&57.1&43.2&53.9&23.8&12.3&47.9&48.8&69.1&16.6&47.5&39.0&61.3&54.7&60.8&32.1&22.0&49.0&44.1&59.4&27.7&43.5\\
                
                WCCN\hfilll\cite{Diba2017}&VGG16&83.9&72.8&64.5&44.1&40.1&65.7&82.5&58.9&33.7&72.5&25.6&53.7&67.4&77.4&26.8&49.1&68.1&27.9&64.5&55.7&56.7\\
                
                Jie \etal\hfilll\cite{Jie2017}&VGG16&72.7&55.3&53.0&27.8&35.2&68.6&81.9&60.7&11.6&71.6&29.7&54.3&64.3&88.2&22.2&53.7&72.2&52.6&68.9&75.5&56.1\\
                
                TST\hfilll\cite{Shi2017a}&AlexNet&--&--&--&--&--&--&--&--&--&--&--&--&--&--&--&--&--&--&--&--&59.5\\
                
                %                SGWSOD\hfilll\cite{Lai2017a}&VGG-F&74.4&74.1&54.2&44.2&38.1&78.0&82.9&62.3&21.6&71.6&31.0&59.1&74.2&85.7&43.6&49.8&72.9&62.5&65.5&69.9&60.8\\
                %                SGWSOD\hfilll\cite{Lai2017a}&VGG-M-1024&74.8&72.0&47.0&40.3&37.7&74.7&84.4&63.8&25.8&66.7&38.5&48.0&68.3&83.7&40.7&52.2&74.0&59.8&66.7&72.7&59.6\\
%                SGWSOD\hfilll\cite{Lai2017a}&VGG16&71.0&76.5&54.9&49.7&54.1&78.0&87.4&68.8&32.4&75.2&29.5&58.0&67.3&84.5&41.5&49.0&78.1&60.3&62.8&78.9&62.9\\
                
                TS$^2$C\hfilll\cite{Wei2018TS2C}&VGG16&84.2&74.1&61.3&52.1&32.1&76.7&82.9&66.6&42.3&70.6&39.5&57.0&61.2&88.4&9.3&54.6&72.2&60.0&65.0&70.3&61.0\\

                OICR\hfilll\cite{Tang2017}&VGG16&81.7&80.4&48.7&49.5&32.8&81.7&85.4&40.1&40.6&79.5&35.7&33.7&60.5&88.8&21.8&57.9&76.3&59.9&75.3&81.4&60.6\\
                
                %K-EM\hfilll\cite{Yan2017}&VGG16&79.8&77.8&66.7&50.3&57.0&80.1&89.9&71.5&29.9&75.9&30.5&58.9&73.2&90.2&25.4&51.8&80.2&60.3&72.4&78.9&65.0\\
                
                MELM\hfilll\cite{Wan}&VGG16&--&--&--&--&--&--&--&--&--&--&--&--&--&--&--&--&--&--&--&--&61.4\\
                
                ZLDN\hfilll\cite{Zhang2018b}&VGG16&74.0&77.8&65.2&37.0&46.7&75.8&83.7&58.8&17.5&73.1&49.0&51.3&76.7&87.4&30.6&47.8&75.0&62.5&64.8&68.8&61.2\\
                
                %                GAL-fWSD300\hfilll\cite{Shen2018GAL}&VGG16&76.5&76.1&64.2&48.1&52.5&80.7&86.1&73.9&30.8&78.7&62.0&71.5&46.7&86.1&60.7&47.8&82.3&74.7&83.1&79.3&68.1\\
%                GAL-fWSD512\hfilll\cite{Shen2018GAL}&VGG16&78.6&81.9&63.6&40.3&48.8&80.7&85.3&76.3&30.3&78.0&54.5&65.3&48.4&86.5&56.3&46.9&76.0&68.1&83.9&73.1&66.1\\
                
                %                ML-LocNet\hfilll\cite{Zhang2018g}&VGG-S&76.9&78.2&65.8&39.8&45.9&78.0&85.1&57.0&16.9&70.9&68.5&56.5&77.0&90.6&47.4&52.2&65.6&60.7&75.5&65.2&63.7\\
                %                ML-LocNet\hfilll\cite{Zhang2018g}&VGG-M&78.0&78.5&66.2&43.2&51.5&76.5&86.8&65.7&34.9&69.5&59.7&55.1&79.5&88.1&40.3&58.3&71.5&64.0&77.0&61.6&65.3\\
%                ML-LocNet\hfilll\cite{Zhang2018g}&VGG16&78.6&82.3&68.2&42.0&53.3&78.5&88.5&70.3&36.4&70.2&60.5&58.0&80.5&88.2&38.8&59.2&75.0&69.0&78.2&64.5&67.0\\
                
                WSRPN\hfilll\cite{Tang}&VGG16&77.5&81.2&55.3&19.7&44.3&80.2&86.6&69.5&10.1&87.7&68.4&52.1&84.4&91.6&57.4&63.4&77.3&58.1&57.0&53.8&63.8\\
                
                Kosugi \etal\hfilll\cite{Kosugi2019}&VGG16&85.5&79.6&68.1&55.1&33.6&83.5&83.1&78.5&42.7&79.8&37.8&61.5&74.4&88.6&32.6&55.7&77.9&63.7&78.4&74.1&66.7\\
                
                C-MIL\hfilll\cite{Wan2019}&VGG16&--&--&--&--&--&--&--&--&--&--&--&--&--&--&--&--&--&--&--&--&65.0\\
                
                Pred Net\hfilll\cite{Arun2018a}&VGG16&88.6&86.3&71.8&53.4&51.2&87.6&89.0&65.3&33.2&86.6&58.8&65.9&87.7&93.3&30.9&58.9&83.4&67.8&78.7&80.2&70.9\\
                
                OICR W-RPN\hfilll\cite{Singh}&VGG16&-&-&-&-&-&-&-&-&-&-&-&-&-&-&-&-&-&-&-&-&66.5\\
                
%                SDCN\hfilll\cite{Li2019a}&VGG16&85.8&83.1&56.2&58.5&44.7&80.2&85.0&77.9&29.6&78.8&53.6&74.2&73.1&88.4&18.2&57.5&74.2&60.8&76.1&79.2&66.8\\
                
                WSOD$^2$\hfilll\cite{Zeng2019}&VGG16&87.1&80.0&74.8&60.1&36.6&79.2&83.8&70.6&43.5&88.4&46.0&74.7&87.4&90.8&44.2&52.4&81.4&61.8&67.7&79.9&69.5\\
                
                OICR+GAM+REG \hfilll\cite{Yanga}&VGG16&81.7&81.2&58.9&54.3&37.8&83.2&86.2&77.0&42.1&83.6&51.3&44.9&78.2&90.8&20.5&56.8&74.2&66.1&81.0&86.0&66.8\\
                %                PCL+GAM+REG \hfilll\cite{Yanga}&VGG16&80.0&83.9&74.2&53.2&48.5&82.7&86.2&69.5&39.3&82.9&53.6&61.4&72.4&91.2&22.4&57.5&83.5&64.8&75.7&77.1&68.0\\
                
                C-MIDN\hfilll\cite{CMIDN}&VGG16&-&-&-&-&-&-&-&-&-&-&-&-&-&-&-&-&-&-&-&-&68.7\\
                
%                OIM+IR\hfilll\cite{Lin2020}&VGG16&-&-&-&-&-&-&-&-&-&-&-&-&-&-&-&-&-&-&-&-&67.2\\
                
                Ren \etal\hfilll\cite{Ren2020}&VGG16&87.5&82.4&76.0&58.0&44.7&82.2&87.5&71.2&49.1&81.5&51.7&53.3&71.4&92.8&38.2&52.8&79.4&61.0&78.3&76.0&68.8\\
                
%                Zeni \etal\hfilll\cite{Zeni2020}&VGG16&86.7&73.3&72.4&55.3&46.9&83.2&87.5&64.5&44.6&76.7&46.4&70.9&67.0&88.0&9.6&56.4&69.1&52.4&79.8&82.8&65.7\\
                
                PG-PS\hfilll\cite{Cheng2020}&VGG16&85.4&80.4&69.1&58.0&35.9&82.7&86.7&82.6&45.5&84.9&44.1&80.2&84.0&89.2&12.3&55.7&79.4&63.4&82.1&82.1&69.2\\
                
%                IM-CFB\hfilll\cite{Yin2021}&VGG16&-&-&-&-&-&-&-&-&-&-&-&-&-&-&-&-&-&-&-&-&70.7\\
                
%                P-MIDN-MGSC\hfilll\cite{Xu2021}&VGG16&-&-&-&-&-&-&-&-&-&-&-&-&-&-&-&-&-&-&-&-&69.8\\
                
                Zhang \etal\hfilll\cite{Zhang2020l}&VGG16&86.3&72.9&71.2&59.0&36.3&80.2&84.4&75.6&30.8&83.6&53.2&75.1&82.7&87.1&37.7&54.6&74.2&59.1&79.8&78.9&68.1\\
                
                %                CASD\hfilll\cite{Huang2020}&VGG16&-&-&-&-&-&-&-&-&-&-&-&-&-&-&-&-&-&-&-&-&70.4\\
                
%                Yang \etal\hfilll\cite{Yang2020a}&VGG16&85.4&79.2&65.2&47.9&42.4&84.3&83.3&76.2&37.8&79.5&47.9&71.4&83.7&90.8&25.8&57.9&71.1&64.5&75.3&80.6&67.5\\
                
                SLV\hfilll\cite{Chen2020}&VGG16&84.6&84.3&73.3&58.5&49.2&80.2&87.0&79.4&46.8&83.6&41.8&79.3&88.8&90.4&19.5&59.7&79.4&67.7&82.9&83.2&71.0\\

                OICR\hfilll\cite{Tang2017}&WSR18&82.1&60.3&81.1&49.3&67.6&81.4&87.2&84.0&33.4&76.8&21.6&78.8&87.0&87.5&30.8&52.6&81.2&66.6&81.8&82.8&68.7\\
                PCL\hfilll\cite{Tang2018b}&WSR18&76.7&81.9&74.4&48.1&53.9&84.5&87.7&86.5&25.4&68.1&36.0&67.4&84.8&86.6&52.5&51.1&81.2&54.9&78.7&62.5&67.1\\
                C-MIL\hfilll\cite{Wan2019}&WSR18&80.3&64.6&68.3&53.0&56.8&84.5&89.1&86.5&28.1&72.4&28.8&77.3&84.1&79.1&56.8&51.8&85.4&62.1&81.1&80.4&68.5\\
                GAM+REG\hfilll\cite{Yanga}&WSR101&88.8&86.6&66.6&57.0&48.5&78.6&91.1&91.3&34.3&88.8&29.1&78.9&90.5&89.6&34.1&41.0&77.0&74.5&87.3&66.4&70.1\\
                
                %==============================================================================================================
                \midrule
                \multirow{3}{*}{\OurMethodx}
                &VGG16&73.0&88.2&67.3&56.0&41.0&85.4&92.6&79.9&21.7&86.1&60.2&81.2&89.9&89.6&19.5&42.0&79.2&79.7&83.6&75.8&69.6\\
                &WSR18&88.4&81.9&84.4&56.8&62.6&89.7&91.0&88.3&42.9&85.5&48.5&85.7&91.4&92.5&37.0&48.1&79.2&83.8&86.6&76.6&\textbf{75.0}\\
%                &WSR18&86.5&81.9&79.3&64.0&61.9&87.6&92.6&86.8&51.6&90.3&39.8&88.5&91.2&91.2&35.7&56.3&75.0&71.2&90.3&75.0&74.8\\
                &WSR50&82.5&72.4&75.3&73.0&54.3&85.4&91.1&89.1&29.0&90.3&40.8&90.8&89.9&87.2&20.2&40.2&77.1&78.8&91.0&71.9&71.5\\
                %                %&ResNet101&\\ %TODO
                
                \midrule
                \multicolumn{23}{c}{WSOD \textbf{without} external object proposal modules or additional data}\\
                \midrule%==============================================================================================================
                
                Shi \etal\hfilll\cite{Shi2017}&--&67.3&54.4&34.3&17.8&1.3&46.6&60.7&68.9&2.5&32.4&16.2&58.9&51.5&64.6&18.2&3.1&20.9&34.7&63.4&5.9&36.2\\
                
                OM + MIL\hfilll\cite{Li2016}&AlexNet&64.3&54.3&42.7&22.7&34.4&58.1&74.3&36.2&24.3&50.4&11.0&29.2&50.5&66.1&11.3&42.9&39.6&18.3&54.0&39.8&41.2\\
                
                %OPG\hfilll\cite{Shen2018OPG}&AlexNet&44.5&33.7&36.1&16.6&7.0&38.2&42.0&57.6&8.5&30.5&29.0&43.2&55.4&56.0&27.6&18.8&36.5&38.9&49.4&14.5&34.2\\
                OPG\hfilll\cite{Shen2018OPG}&VGG16&57.1&43.2&53.9&23.8&12.3&47.9&48.8&69.1&16.6&47.5&39.0&61.3&54.7&60.8&32.1&22.0&49.0&44.1&59.4&27.7&43.5\\
                
                %==============================================================================================================
                %                \midrule
                %                \multirow{2}{*}{UWSOD}
                %                &VGG16&77.8&85.8&66.0&56.0&39.1&74.2&91.4&41.4&30.3&81.9&33.0&78.9&90.5&85.6&7.6&46.4&68.8&67.0&76.1&61.7&63.0\\
                %                &WSR18&80.4&85.3&79.4&42.0&65.5&78.4&90.7&49.7&18.8&73.9&48.5&63.1&87.8&90.8&37.4&47.4&77.1&54.1&81.9&23.4&63.8\\
                %                %&ResNet50&\\ %TODO
                %                %&ResNet101&\\ %TODO
                
                \midrule
                \multirow{3}{*}{\OurMethod}
                &VGG16&82.5&89.8&74.7&67.0&45.7&84.3&90.5&39.7&27.2&90.3&40.8&63.3&89.9&92.8&28.7&58.9&79.2&69.5&82.8&62.5&68.0\\
                &WSR18&80.2&90.6&74.7&69.0&53.3&88.8&92.6&59.8&32.6&91.7&45.6&77.1&91.9&88.8&40.6&58.0&81.4&75.4&85.8&51.6&\textbf{71.5}\\
                &WSR50&86.6&82.8&85.0&66.7&67.6&84.5&91.8&92.0&16.5&79.7&59.8&90.2&95.7&93.3&26.2&15.8&81.3&74.8&82.7&39.8&70.6\\
                \hline
            \end{tabular}
        \end{adjustbox}
    \end{center}
    %    \vspace{-20pt}
    \label{table_voc2007_sota_corloc}
\end{table*}

\begin{table*}[t]
    \centering
    \caption{
        Comparison with the state-of-the-art methods on PASCAL VOC 2012 and MS COCO.
    }
    %\footnotesize
    %    \begin{center}
        %        \begin{adjustbox}{max width=0.5\textwidth}
            \begin{tabular}{l|c|cc|ccc}
                \toprule

                %                \multirow{3}{*}{Method}&\multirow{3}{*}{Bakcbone}&\multicolumn{2}{c|}{VOC 2012}&\multicolumn{3}{c}{MS COCO}\\
                %                &&\multirow{2}{*}{$m$AP}&\multirow{2}{*}{CorLoc}&\multicolumn{3}{c}{AP@IoU}\\
                %                &&&&0.5:0.95&0.5&0.75\\

                \multirow{2}{*}{Method}&\multirow{2}{*}{Bakcbone}&\multicolumn{2}{c|}{VOC 2012}&\multicolumn{3}{c}{MS COCO}\\
                &&\multirow{1}{*}{$m$AP}&\multirow{1}{*}{CorLoc}&AP&AP$_{50}$&AP$_{75}$\\

                \midrule
                \multicolumn{7}{c}{WSOD \textbf{with} external object proposal modules or additional data}\\
                \midrule%==============================================================================================================

                %                \multicolumn{9}{c}{Object Discovery}\\
                %                \hline

                %                Siva~\etal\hfilll\cite{siva2012defence}&&--&30.2&--&--&--&--&--\\

                %                Shi~\etal\hfilll\cite{Shi2017}&&--&36.2&--&--&--&--&--\\

                %                Rochan~\etal\hfilll\cite{Rochan2015}&AlexNet&--&58.8&--&--&--&--&--\\

                %                RMI-SVM\hfilll\cite{Wang2015}&AlexNet&--&40.2&--&--&--&--&--\\

                %                STL\hfilll\cite{Bazzani2016}&AlexNet&--&54.4&--&--&--&--&--\\

                %                Song~\etal\hfilll\cite{Song2014Weakly}&AlexNet&24.6&--&--&--&--&--&--\\

                %                Bilen~\etal\hfilll\cite{Bilen2014}&AlexNet&26.4&--&--&--&--&--&--\\

                %                Cover+SLSVM\hfilll\cite{song2014learning}&AlexNet&22.7&--&--&--&--&--&--\\

                %                LCL-pLSA\hfilll\cite{Wang2014}&AlexNet&30.9&48.5&--&--&--&--&--\\
                %                LCL+Context\hfilll\cite{Wang2014}&AlexNet&31.6&--&--&--&--&--&--\\

                %                Bilen~\etal\hfilll\cite{Bilen2015}&AlexNet&27.7&43.7&--&--&--&--&--\\

                %                Multi-Fold MIL\hfilll\cite{Cinbis2015}&AlexNet&30.2&52.0&--&--&--&--&--\\

                %                Teh\hfilll\cite{Teh2016}&AlexNet&34.5&64.6&--&--&--&--&--\\

                %                WSDDN\hfilll\cite{Bilen2016}&VGG-F&34.5&54.2&--&--&--&--&--\\
                %                WSDDN\hfilll\cite{Bilen2016}&VGG-M&34.9&56.1&--&--&--&--&--\\
                WSDDN\hfilll\cite{Bilen2016}&VGG16&--&--&9.5&19.2&8.2\\

                ContextLocNet\hfilll\cite{Kantorov2016}&VGG-F&35.3&54.8&11.1&22.1&10.7\\

                %                Beam Search\hfilll\cite{Bency2016}&VGG16&25.7&--&26.5&--&--&--&--\\

                %                OM+MIL\hfilll\cite{Li2016}&AlexNet&23.4&41.2&29.1&--&--&--&--\\

                %                OPG\hfilll\cite{Shen2018OPG}&AlexNet&27.0&34.2&--&--&--&--&--\\
                %                OPG\hfilll\cite{Shen2018OPG}&VGG16&28.8&43.5&--&--&--&--&--\\

                WCCN\hfilll\cite{Diba2017}&VGG16&37.9&--&--&--&--\\

                %                Jiang~\etal\hfilll\cite{Jiang2017}&AlexNet&37.4&57.3&33.6&--&--&--&--\\

                Jie~\etal\hfilll\cite{Jie2017}&VGG16&38.3&58.8&--&--&--\\

                %                SPAM-CAM\hfilll\cite{Gudi2017}&VGG16&27.5&--&--&--&--&--&--\\

                %                TST\hfilll\cite{Shi2017a}&AlexNet&33.8&59.5&--&--&--&--&--\\

                %                SGWSOD\hfilll\cite{Lai2017a}&VGG-F&38.9&60.8&--&--&--&--&--\\
                %                SGWSOD\hfilll\cite{Lai2017a}&VGG-M-1024&39.4&59.6&--&--&--&--&--\\
                %                SGWSOD\hfilll\cite{Lai2017a}&VGG16&39.6&62.9&--&--&--\\

                TS$^2$C\hfilll\cite{Wei2018TS2C}&VGG16&40.0&64.4&--&--&--\\

                %                CSC C5\hfilll\cite{Shen2019CSC}&VGG-F&38.4&60.4&--&--&--&--&--\\
                %                CSC C5\hfilll\cite{Shen2019CSC}&VGG-M&39.3&60.6&--&--&--&--&--\\
                %                CSC C5\hfilll\cite{Shen2019CSC}&VGG16&37.1&61.4&\textbf{12.9}&\textbf{23.8}&\textbf{13.2}\\

                %                WS-JDS\hfilll\cite{Shen2019WSJDS}&VGG16&39.1&63.5&--&--&--\\

                %                Oquab~\etal\hfilll\cite{Oquab2015}&AlexNet&11.7&--&--&--&--\\

                %                \hline%==============================================================================================================

                %                \multirow{4}{*}{WSDDN}\hfilll\multirow{4}{*}{\cite{Bilen2016}}&VGG16&34.8&53.5&--&--\\
                %                &ResNet18-WS&43.4&63.1&--&--\\
                %                &ResNet50-WS&44.0&63.6&--&--\\
                %                &ResNet101-WS&44.1&64.0&--&--\\
                %
                %                \hline%==============================================================================================================
                %
                %                \multirow{4}{*}{ContextLocNet}\hfilll\multirow{4}{*}{\cite{Kantorov2016}}&VGG-F&36.3&55.1&35.3&54.8\\
                %                &ResNet18-WS&45.4&64.7&42.0$^{\ddag}$&66.7\\
                %                &ResNet50-WS&45.3&65.1&--&--\\ %TODO
                %                &ResNet101-WS&45.9&65.7&--&--\\ %TODO
                %
                %                \hline%==============================================================================================================

                %                \multicolumn{9}{c}{Object Discovery + Instance Refinement}\\
                %                \hline

                %                OICR\hfilll\cite{Tang2017}&VGG-M&37.9&57.3&34.6&60.7&--&--&--\\
                OICR\hfilll\cite{Tang2017}&VGG16&37.9&62.1&--&--&--\\

                %                K-EM\hfilll\cite{Yan2017}&VGG16&46.1&65.0&--&--&--&--&--\\

                MELM\hfilll\cite{Wan}&VGG16&42.4&--&--&--&--\\

                ZLDN\hfilll\cite{Zhang2018b}&VGG16&42.9&61.5&--&--&--\\

                %                GAL-fWSD300\hfilll\cite{Shen2018GAL}&VGG16&47.0&68.1&43.1&67.2&--&--&--\\
                %                GAL-fWSD512\hfilll\cite{Shen2018GAL}&VGG16&47.5&66.1&--&--&--&--&--\\

                %                ML-LocNet\hfilll\cite{Zhang2018g}&VGG-S&44.4&63.7&--&--&--&--&--\\
                %                ML-LocNet\hfilll\cite{Zhang2018g}&VGG-M&47.2&65.3&--&--&--&--&--\\
                %                ML-LocNet\hfilll\cite{Zhang2018g}&VGG16&42.2&66.3&--&16.2&--\\

                WSRPN\hfilll\cite{Tang}&VGG16&40.8&64.9&--&--&--\\

                PCL\hfilll \cite{Tang2018b}&VGG16&--&--&8.5&19.4&--\\

                Kosugi~\etal\hfilll\cite{Kosugi2019}&VGG16&43.4&66.7&--&--&--\\

                %                C-MIL\hfilll\cite{Wan2019}&VGG-F&40.7&--&--&--&--&--&--\\
                C-MIL\hfilll\cite{Wan2019}&VGG16&46.7&67.4&--&--&--\\

                Pred Net\hfilll\cite{Arun2018a}&VGG16&48.4&69.5&--&--&--\\

                %                WSDDN + W-RPN\hfilll\cite{Singh}&VGG16&39.4&--&--&--&--&--&--\\
                OICR W-RPN\hfilll\cite{Singh}&VGG16&43.2&67.5&--&--&--\\

                %                SDCN\hfilll\cite{Li2019a}&VGG16&43.5&67.9&--&--&--\\

                %                Sona~\etal\hfilll\cite{Sona}&VGG16&45.4&--&--&--&--&--&--\\

                WSOD$^2$\hfilll\cite{Zeng2019}&VGG16&47.2&{71.9}&10.8&22.7&--\\

                %                OICR+GAM+REG\hfilll\cite{Yanga}&VGG16&48.6&66.8&--&--&--&--&--\\
                %                PCL+GAM+REG\hfilll\cite{Yanga}&VGG16&51.5&68.0&--&--&--&--&--\\

                C-MIDN\hfilll\cite{CMIDN}&VGG16&50.2&71.2&9.6&21.4&--\\

                %                OIM\hfilll\cite{Lin2020}&VGG16&48.2&--&--&--&--&--&--\\
                %                OIM+IR\hfilll\cite{Lin2020}&VGG16&45.3&67.1&--&--&--\\

                %                Ren~\etal\hfilll\cite{Ren2020}&VGG16&{52.1}&70.9&12.4&25.8&10.5\\

                %                Zeni~\etal\hfilll\cite{Zeni2020}&VGG16&--&66.3&--&--&--\\

                PG-PS\hfilll\cite{Cheng2020}&VGG16&48.3&68.7&--&20.7&--\\

                %                IM-CFB\hfilll\cite{Yin2021}&VGG16&49.4&69.6&--&--&--\\

                %                P-MIDN-MGSC\hfilll\cite{Xu2021}&VGG16&52.8&73.3&13.1&27.4&-\\

                Zhang~\etal\hfilll\cite{Zhang2020l}&VGG16&46.3&68.7&11.1&23.6&--\\

                %                CASD\hfilll\cite{Huang2020}&VGG16&53.6&72.3&12.8&26.4&--\\
                %                CASD\hfilll\cite{Huang2020}&VGG16&--&--&13.9&27.8&--\\

                %                Yang~\etal\hfilll\cite{Yang2020a}&VGG16&48.3&71.6&--&--&--\\

                SLV\hfilll\cite{Chen2020}&VGG16&49.2&69.2&--&--&--\\

                %                PML-LocNet\hfilll\cite{Zhang2020a}&VGG16&43.2&67.5&--&20.9&--\\

                \midrule
                \multirow{2}{*}{\OurMethodx}
                %                &VGG16&&&7.3&19.6&3.9\\
                % &VGG16&53.5&70.8&10.3&22.6&6.9\\ %TODO
                &VGG16&53.0&71.2&10.4&24.9&7.2\\ % by hunterj
                % &WSR18&\textbf{55.9}&\textbf{75.5}&9.6&19.4&6.1\\ %TODO
                &WSR18&\textbf{57.1}&\textbf{78.6}&8.0&20.8&5.0\\ % by hunterj

                \midrule
                \multicolumn{7}{c}{WSOD \textbf{without} external object proposal modules or additional data}\\
                \midrule%==============================================================================================================

                Shi~\etal\hfilll\cite{Shi2017}&--&36.2&--&--&--&--\\

                Beam Search\hfilll\cite{Bency2016}&VGG16&26.5&--&--&--&--\\

                OM+MIL\hfilll\cite{Li2016}&AlexNet&29.1&--&--&--&--\\

                %OPG\hfilll\cite{Shen2018OPG}&AlexNet&27.0&34.2&--&--&--&--&--\\
                % OPG\hfilll\cite{Shen2018OPG}&VGG16&--&--&--&--&--\\

                %                SPAM-CAM\hfilll\cite{Gudi2017}&VGG16&27.5&--&--&--&--&--&--\\

                %                \midrule
                %                \multirow{2}{*}{UWSOD}&VGG16&44.0&63.0&45.1&65.2&2.5&9.3&1.1\\
                %                &WSR18&\textbf{45.0}&\textbf{63.8}&\textbf{46.2}&\textbf{65.7}&3.1&10.1&1.4\\

                \midrule
                \multirow{2}{*}{\OurMethod}
                &VGG16&48.1&68.1&\textbf{6.3}&\textbf{17.9}&\textbf{3.1}\\ %TODO
                % &WSR18&\textbf{50.2}&\textbf{71.8}&5.4&16.2&2.6\\ %TODO
                &WSR18&\textbf{50.3}&\textbf{72.3}&5.8&16.8&3.1\\ % by hunterj

                \midrule%==============================================================================================================
                \multicolumn{7}{c}{FSOD}\\
                \midrule

                Fast RCNN\hfilll\cite{FASTRCNN}&VGG16&65.7&--&18.9&38.6&--\\
                Faster RCNN\hfilll\cite{FASTERRCNN}&VGG16&67.0&--&21.2&41.5&--\\

                %                \midrule%==============================================================================================================
                %                \multicolumn{9}{c}{Upper Bound with GT-bbox Known}\\
                %                \midrule
                %
                %                OICR + GAM + REG\hfilll\cite{Yanga}&VGG16&66.9&--&65.7&--&18.9&38.6&--\\
                %                Ren~\etal\hfilll\cite{Ren2020}&VGG16&66.9&--&65.7&--&18.9&38.6&--\\
                %                UWSOD&VGG16&69.9&--&67.0&--&21.2&41.5&--\\

                %                \midrule%==============================================================================================================
                %                \multicolumn{9}{c}{WSOD with Cls-agnostic GT-bbox Known}\\
                %                \midrule
                %
                %                OICR + GAM + REG\hfilll\cite{Yanga}&VGG16&53.9&82.1&13.7&27.1&12.5\\
                %
                %                Ren~\etal\hfilll\cite{Ren2020}&VGG16&62.1&88.9&14.1&28.9&12.7\\

                %                \midrule
                %                \multirow{2}{*}{UWSOD}&  VGG16  &{67.7}&\textbf{93.3}&{65.3}&91.1&\textbf{15.3}&\textbf{32.4}&\textbf{12.8}\\
                %
                %                &  WSR18  &\textbf{69.7}&{92.5}&\textbf{66.1}&\textbf{92.3}&{13.7}&{27.9}&{12.5}\\

                \bottomrule%==============================================================================================================

            \end{tabular}
            %        \end{adjustbox}
        %    \end{center}
    \label{table_voc2007voc2012mscoco_sota}
\end{table*}
%%%%%%%%%%%%%%%%%%%%%%%%%%%%%%%%%%%%%%%
%%
\begin{table*}[t]
    \centering
    \caption{
        Upper bound of various WSOD methods on PASCAL VOC and MS COCO.
    }
    %\footnotesize
    %    \begin{center}
        %        \begin{adjustbox}{max width=1.0\linewidth}
            \begin{tabular}{l|c|cc|cc|ccc}
                \toprule

                %                \multirow{3}{*}{Method}&\multirow{3}{*}{Bakcbone}&\multicolumn{2}{c|}{VOC 2007}&\multicolumn{2}{c|}{VOC 2012}&\multicolumn{3}{c}{MS COCO}\\
                %                &&\multirow{2}{*}{$m$AP}&\multirow{2}{*}{CorLoc}&\multirow{2}{*}{$m$AP}&\multirow{2}{*}{CorLoc}&\multicolumn{3}{c}{AP@IoU}\\
                %                &&&&&&0.5:0.95&0.5&0.75\\

                \multirow{2}{*}{Method}&\multirow{2}{*}{Bakcbone}&\multicolumn{2}{c|}{VOC 2007}&\multicolumn{2}{c|}{VOC 2012}&\multicolumn{3}{c}{MS COCO}\\
                &&\multirow{1}{*}{$m$AP}&\multirow{1}{*}{CorLoc}&\multirow{1}{*}{$m$AP}&\multirow{1}{*}{CorLoc}&AP&AP$_{50}$&AP$_{75}$\\

                \midrule%==============================================================================================================
                \multicolumn{9}{c}{FSOD}\\
                \midrule

                Fast RCNN\hfilll\cite{FASTRCNN}&VGG16&66.9&--&65.7&--&18.9&38.6&--\\
                Faster RCNN\hfilll\cite{FASTERRCNN}&VGG16&69.9&--&67.0&--&21.2&41.5&--\\

                \midrule%==============================================================================================================
                \multicolumn{9}{c}{GT-bbox Known}\\
                \midrule

                Yang \etal\hfilll\cite{Yanga}&VGG16&66.9&--&65.7&--&18.9&38.6&--\\
                Ren \etal\hfilll\cite{Ren2020}&VGG16&66.9&--&65.7&--&18.9&38.6&--\\

                %                \midrule
                %                UWSOD&VGG16&69.9&--&67.0&--&21.2&41.5&--\\

                \midrule
                \multirow{1}{*}{\OurMethod}&VGG16&69.9&--&67.0&--&21.2&41.5&--\\

                \midrule%==============================================================================================================
                \multicolumn{9}{c}{Cls-agnostic GT-bbox Known}\\
                \midrule

                Yang \etal\hfilll\cite{Yanga}&VGG16&54.2&67.6&53.9&68.1&13.7&27.1&12.5\\

                Ren \etal\hfilll\cite{Ren2020}&VGG16&64.9&73.8&62.1&74.9&15.1&29.9&13.7\\

                %                \midrule
                %                \multirow{2}{*}{UWSOD}&  VGG16  &{67.7}&\textbf{93.3}&{65.3}&91.1&\textbf{15.3}&\textbf{32.4}&\textbf{12.8}\\
                %                &  WSR18  &\textbf{69.7}&{92.5}&\textbf{66.1}&\textbf{92.3}&{13.7}&{27.9}&{12.5}\\

                \midrule
                \multirow{2}{*}{\OurMethod}&  VGG16  &{67.7}&\textbf{93.3}&{65.3}&91.1&\textbf{15.3}&\textbf{32.4}&\textbf{12.8}\\
                &  WSR18  &\textbf{69.7}&{92.5}&\textbf{66.1}&\textbf{92.3}&{13.7}&{27.9}&{12.5}\\

                \bottomrule%==============================================================================================================

            \end{tabular}
            %        \end{adjustbox}
        %    \end{center}
    \label{table_voc2007voc2012mscoco_sota_gt}
\end{table*}
%%%%%%%%%%%%%%%%%%%%%%%%%%%%%%%%%%%%%%%%%%%%%%
%%
\begin{table*}[t]
    \centering
    \caption{
        Adopting HUWSOD to multi-stage WSOD methods on PASCAL VOC and MS COCO.
    }
    %\footnotesize
    % \begin{center}
    % \begin{adjustbox}{max width=1.0\linewidth}
            \begin{tabular}{l|c|cc|cc|ccc}
                \toprule

                %                \multirow{3}{*}{Method}&\multirow{3}{*}{Bakcbone}&\multicolumn{2}{c|}{VOC 2007}&\multicolumn{2}{c|}{VOC 2012}&\multicolumn{3}{c}{MS COCO}\\
                %                &&\multirow{2}{*}{$m$AP}&\multirow{2}{*}{CorLoc}&\multirow{2}{*}{$m$AP}&\multirow{2}{*}{CorLoc}&\multicolumn{3}{c}{AP@IoU}\\
                %                &&&&&&0.5:0.95&0.5&0.75\\

                \multirow{2}{*}{Method}&\multirow{2}{*}{Bakcbone}&\multicolumn{2}{c|}{VOC 2007}&\multicolumn{2}{c|}{VOC 2012}&\multicolumn{3}{c}{MS COCO}\\
                &&\multirow{1}{*}{$m$AP}&\multirow{1}{*}{CorLoc}&\multirow{1}{*}{$m$AP}&\multirow{1}{*}{CorLoc}&AP&AP$_{50}$&AP$_{75}$\\

                \midrule
                \multirow{2}{*}{\OurMethod} &  VGG16  & 55.5 & 69.6 & 53.5 & 70.8 &10.3 & 22.6 & 6.9 \\
                &  WSR18  & 59.5 & 75.0 & 55.9 & 75.5 & 9.6 & 19.4 & 6.1 \\

                \midrule
                \multirow{2}{*}{SoS~\cite{Sui}} &  VGG16  & 60.3 & -- &57.7 & -- & 15.5 & 30.5 & 14.3\\
                &  ResNet50  & 64.4 & -- & 61.9 & -- & 16.6 & 32.8 & 15.2\\

                \midrule
                %\multirow{2}{*}{\OurMethod + SoS~\cite{Sui}}    &  VGG16     & 61.1 & 75.7 & 59.0 & 76.1 & 15.1 & 29.6 & 13.2 \\
                %                                                &  ResNet50  & 64.5 & 80.1 & 62.4 & 80.4 & 16.2 & 31.3 &  14.5 \\

                \multirow{1}{*}{\OurMethod + SoS}    &  ResNet50  & 64.5 & 80.1 & 62.4 & 80.4 & 16.2 & 31.3 &  14.5 \\

                \midrule
                \multirow{1}{*}{BiB~\cite{Vo2022}} &  VGG16  & 65.1 & -- & -- & -- & 17.2 & 34.1 & --\\
                %&  ResNet50  &\\

                \midrule
                \multirow{1}{*}{\OurMethod + BiB} & VGG16 & 65.7 & 82.7 & 63.1 & 83.2 & 17.9 & 36.2 & 16.4\\
                %&  ResNet50  &\\

                \bottomrule%==============================================================================================================

            \end{tabular}
    % \end{adjustbox}
    % \end{center}
    \label{table_voc2007voc2012mscoco_sota_gt_compatibility}
\end{table*}
%%%%%%%%%%%%%%%%%%%%%%%%%%%%%%%%%%%%%%%%%%%%%%
%%
\subsubsection{Autoencoder object proposal generator}
The proposed AEOPG aims to utilize full image feature maps to capture salient objects, which is also end-to-end learnable.
We ablate the impact of two loss functions, \ie, $\mathcal{L}_\mathrm{Encoder}$ and $\mathcal{L}_\mathrm{Decoder}$, the structures of encoder $E$ and decoder $D$ as well as the channels $q$ of the latent representation $E(F)$.
We employ the model \textit{in the row~(j) of Tab.~\ref{table_voc2007_abl} as the baseline}, which combines SSOPG, SEM, and MRRP with the best hyper-parameters.
The network structure of encoder $\mathcal{E}$ and decoder $\mathcal{D}$ is denoted as triplet tuples $( C_\mathrm{in}, C_\mathrm{stride}, C_\mathrm{out} )$,
%, which is described in Section~\ref{sec_AEOPG}.
%
where $C_\mathrm{in}$ is the channel number of backbone outputs, which is set to $512$, $512$, and $2,048$ for VGG16, ResNet18, and ResNet50, respectively;
$C_\mathrm{stride}$ denotes the channel reduction/increase ratio between the successive convolutional layers in encoder/decoder;
And $C_\mathrm{out}$ is the channel number of the last feature maps that project to the latent representation.
Rows~(a-c) of Tab.~\ref{table_voc2007_abl_aeopg} show that both $\mathcal{L}_\mathrm{Encoder}$ and $\mathcal{L}_\mathrm{Decoder}$ improve detection and localization.
And a complete autoencoder brings the largest performance gains, which demonstrates the effectiveness of AEOPG.
In the rows~(d-f), we vary the channels of latent representations and observe that single-channel feature maps achieve the best performance with $68.0 \%$ CorLoc and $50.2\%$ $m$AP, respectively.
To explain, when the number of channels is large than $1$, the information of objects may be scattered into different channels.
Therefore, encoding the high-level semantic information into a single feature map aggregates salient cues from different channels.
To further ablate AEOPG with different structures, rows~(g-i) show that AEOPG is insensitive to the structure of the autoencoder.
%%%%%%%%%%%%%%%%%%%%%%%%%%%%%%%%%%%%%%%%%%%%%%%%%%%%%%%%
\begin{figure}[ht]
    %\scriptsize
    \begin{center}
        \includegraphics[width=0.45\textwidth]{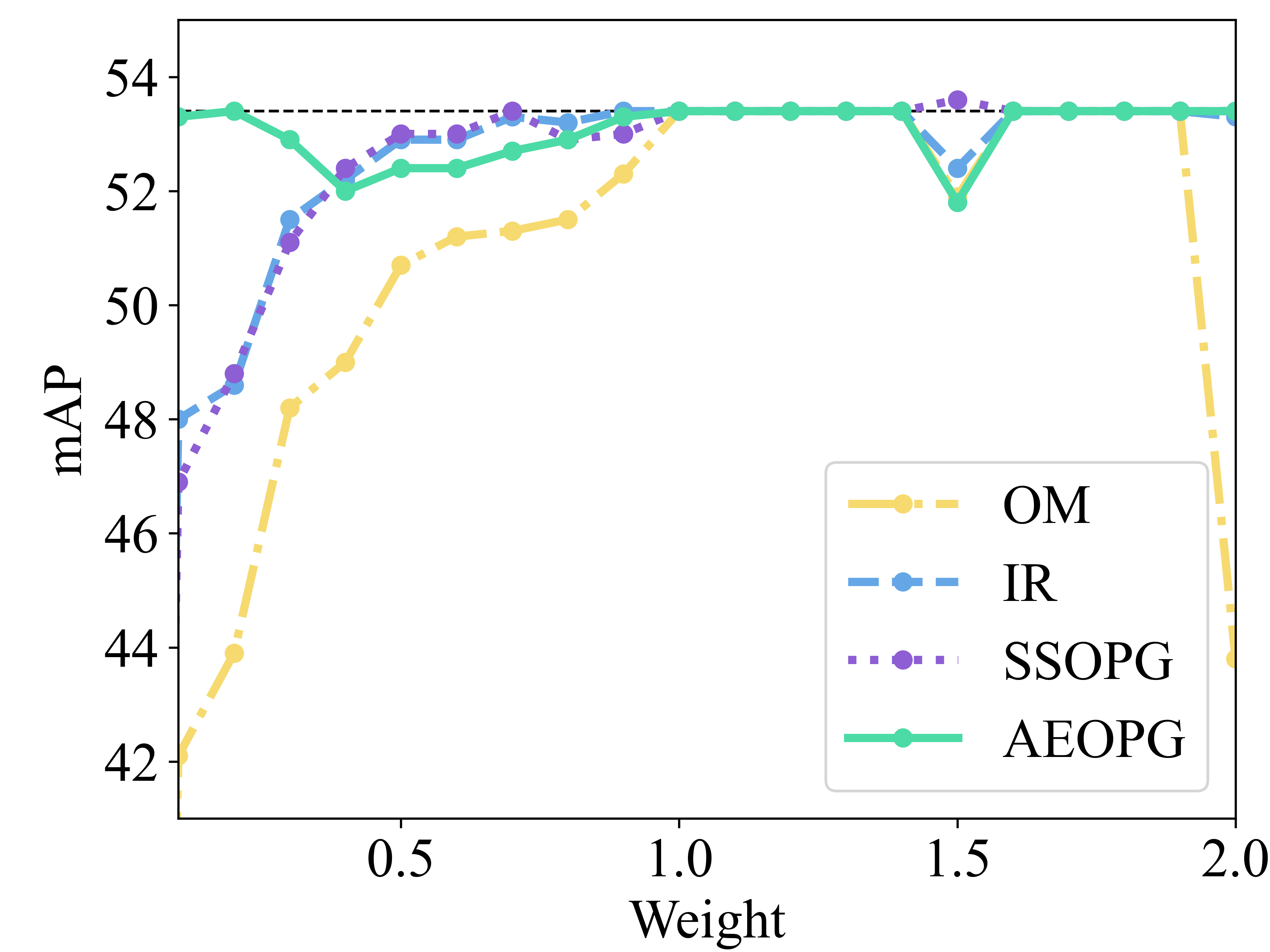}
    \end{center}
    \caption{
        Ablation study of loss weights on PASCAL VOC 2007 \textit{test} in terms of $m$AP~(\%).
    }
    \label{fig_vis_loss_weight}
\end{figure}
%%%%%%%%%%%%%%%%%%%%%%%%%%%%%%%%%%%%%%
%%
\subsubsection{Consistency-regularized self-training}
The proposed CCR aims to enforce consistent predictions produced from stochastic augmentations of the same image.
We analyze the influence of augmentation number $\mathcal{T}$, types of augmentations, and EMA decay rate $\beta$.
We employ the model \textit{in the row~(c) of Tab.~\ref{table_voc2007_abl_aeopg} as the baseline}, which uses the best hyper-parameters of other modules in terms of $m$AP.
In the rows~(a-d) of Tab.~\ref{table_voc2007_abl_crst}, different data augmentation policies have certain improvements.
And combing all data augmentation policies achieves the most gains, as the large diversity brings strong consistency constraints.
Rows~(e-g) report that increasing $\mathcal{T}$ has slight performance improvement, as $\mathcal{T}$ mainly affects the training batch size.
Rows~(h-k) show that the default EMA decay rate $\beta$ improves the final performance by $0.4\%$ CorLoc and $0.5\%$ $m$AP, respectively.
The good $\beta$ values span roughly an order of magnitude, the performance degrades quickly outside these ranges.
%%%%%%%%%%%%%%%%%%%%%%%%%%%%%%%%%%%%%%%%%%%%%%%
\subsubsection{Influence of loss weight}
We also ablate each objective function with different weights, which range in $[0, 1]$.
When tuning the weight of each loss, we keep others as default, \ie, $1$.
We plot the weight-$m$AP trade-off curve in Fig.~\ref{fig_vis_loss_weight}.
As expected, gradually reducing the loss weights usually causes performance drops.
The plot also shows that the overall performance is robust to the loss weight of AEOPG, which mainly provides high-quality proposals for the salient object.
However, we observe that SSOPG is more sensitive to loss weight than AEOPG, as proposal learning is a vital component of WSOD.
Surprisingly, OM loss has significant impact on the final performance among all losses.
Although all loss weights are equally set to $1$ as default, we empirically find that grid search over those hyper-parameter spaces can find better-performing configurations for \OurMethod.
%%%%%%%%%%%%%%%%%%%%%%%%%%%%%%%%%%%%%%%%%%%%%%%%%%%%%%%%%%%%%%
\subsection{Comparison with the State-of-the-Art Methods}
To fully compare with other backbones, we categorize the detection results into two groups according to whether they use external object proposal modules or additional data.
We first compare the results on Pascal VOC 2007 in terms of \emph{m}AP in the bottom part of Tab.~\ref{table_voc2007_sota_map}.
\OurMethod significantly outperforms the state-of-the-art methods that \textbf{do not} use external modules or additional data.
This indicates the efficiency of \OurMethod.
Compared to other methods \textbf{with} external object proposals or data, \OurMethod also achieves competitive results.
It demonstrates that \OurMethod significantly reduces the performance gap between WSOD and FSOD methods.
Note that \OurMethod is compatible with external object proposals.
Thus, we further add offline object proposals to \OurMethod only during training, which is denoted as \OurMethodx.
The top part of Tab.~\ref{table_voc2007_sota_map} shows that \OurMethodx significantly outperforms all previous methods.
In Tab.~\ref{table_voc2007_sota_corloc}, we compare the localization performance on \textit{train+val} images in terms of CorLoc.
\OurMethod reports $71.5\%$ CorLoc, which consistently outperforms other methods that \textbf{do not} use external modules or additional data.
\OurMethodx further achieves $75.0\%$ CorLoc and suppress other methods \textbf{with} external object proposals and data.
We also report the performance on Pascal VOC 2012 and MS COCO in Tab.~\ref{table_voc2007voc2012mscoco_sota}.
\OurMethod shows superior results compared to other models and achieves new state-of-the-art performance.
The benefits are mainly from effectively learnable object proposal generation, well-designed self-training scheme.
For AP and AP$_{50}$, our results are only comparable to the state-of-the-art methods, which is mainly because that MS COCO has more complex scenes and a larger category set.
Thus, design principles in traditional proposal methods may further improve end-to-end proposal generators for WSOD.
%%%%%%%%%%%%%%%%%%%%%%%%%%%%%%%%%%%%%%%%%%%%%%%%
%%
\begin{table*}[t]
    % \vspace{-15pt}
    \caption{
        Comparison with the state-of-the-art methods on Cityscapes in terms of AP on val set.
    }
    \footnotesize
    \begin{center}
        \begin{adjustbox}{max width=1.0\linewidth}
            \begin{tabular}{l|c|cccccccc|c}
                \hline                
                Method&Backbone&person&rider&car&truck&bus&train&motorcycle&bicycle&$AP_{50}$\\
                
                \hline
                % \multicolumn{12}{c}{Object Mining}\\
                %==============================================================================================================
                % \hline
                
                % WSDDN\hfilll\cite{Bilen2016}&VGG16&--&--&--&--&--&--&--&--&--&--&--\\
                %23.2 70.4
                % WSDDN\hfilll\cite{Bilen2016}&WSR18&0.0&0.8&0.02&6.6&11.9&2.7&0.9&0.004&2.9&8.8&1.0\\
                WSDDN\hfilll\cite{Bilen2016}&WSR18&0.0&2.7&0.1&20.2&36.1&8.4&2.9&0.0&8.8\\

                % WSDDN+SAM proposals\hfilll\cite{Bilen2016}&WSR18&5.1&16.7&20.1&27.4&41.8&12.9&7.9&5.2&17.2&27.4&18.2\\
                %==============================================================================================================
                % \hline
                % \multicolumn{12}{c}{Object Mining + Instance Refinement}\\
                % \hline
                
                % OICR\hfilll\cite{Tang2017}&VGG16&--&--&--&--&--&--&--&--&--&--&--\\
                % 22.346  82.304
                % OICR\hfilll\cite{Tang2017}&WSR18&0.059&0.000&5.627&3.336&9.795&2.463&0.829&0.237&2.793&10.288&0.723\\               
                OICR\hfilll\cite{Tang2017}&WSR18&0.2&0.0&20.7&12.3&36.1&9.1&3.1&0.9&10.3\\                 
                %==============================================================================================================
                % \hline
                % HUWSOD&VGG16&--&--&--&--&--&--&--&--&--&--&--\\
                %42.1784 127.172
                % HUWSOD&WSR18&0.0&1.333&10.293&8.942&14.752&4.415&2.089&0.355&5.2723&16.0215&2.3128\\
                HUWSOD&WSR18&0.9&4.0&31.2&27.1&44.8&13.4&6.3&1.1&16.1\\
                % \hline
                % \multicolumn{12}{c}{FSOD}\\
                % \hline
  
                % Fast RCNN\hfilll\cite{FASTRCNN}&RN50&12.4&14.4&33.2&28.8&51.1&21.5&13.4&9.3&23.0&37.7&23.8\\
                
                % Faster RCNN\hfilll\cite{FASTERRCNN}&RN50&34.5&40.3&55.1&31.5&56.7&20.5&26.2&30.5&36.9&62.9&38.1\\
                
                \hline
            \end{tabular}
        \end{adjustbox}
    \end{center}
    % \vspace{-15pt}
    \label{table_cityscapes}
\end{table*}
%%%%%%%%%%%%%%%%%%%%%%%%%%%%%%%%%%%%%%%%%%%%%%
%%
%%
\begin{figure*}[t]
    %    \vspace{-20pt}
    %\scriptsize
    \begin{center}
        \begin{subfigure}[t]{1.0\textwidth}
            \begin{center}
                \stackunder[0pt]{$2$ proposals}{\includegraphics[width=0.20\textwidth]{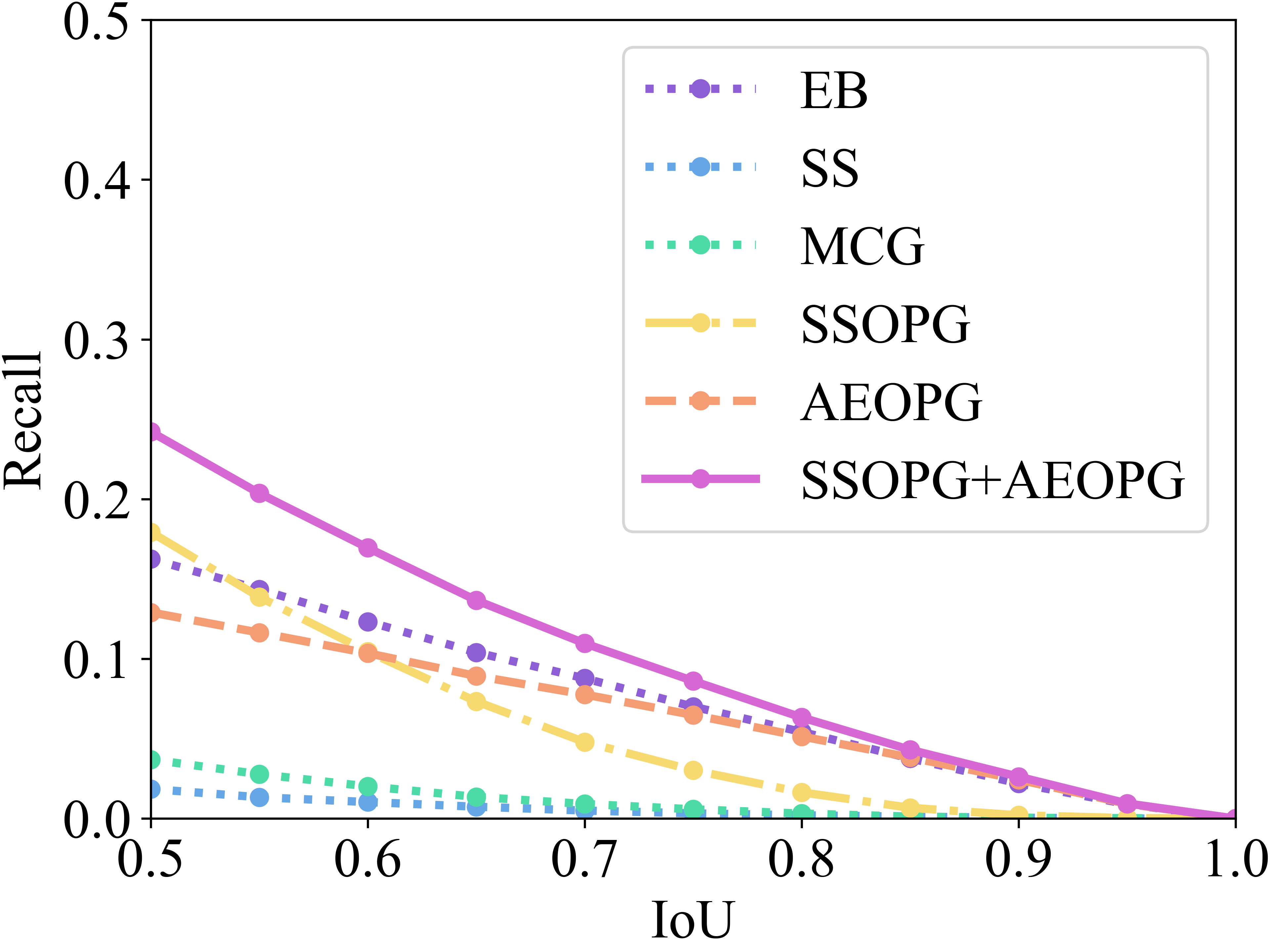}}%
                \stackunder[0pt]{$4$ proposals}{\includegraphics[width=0.20\textwidth]{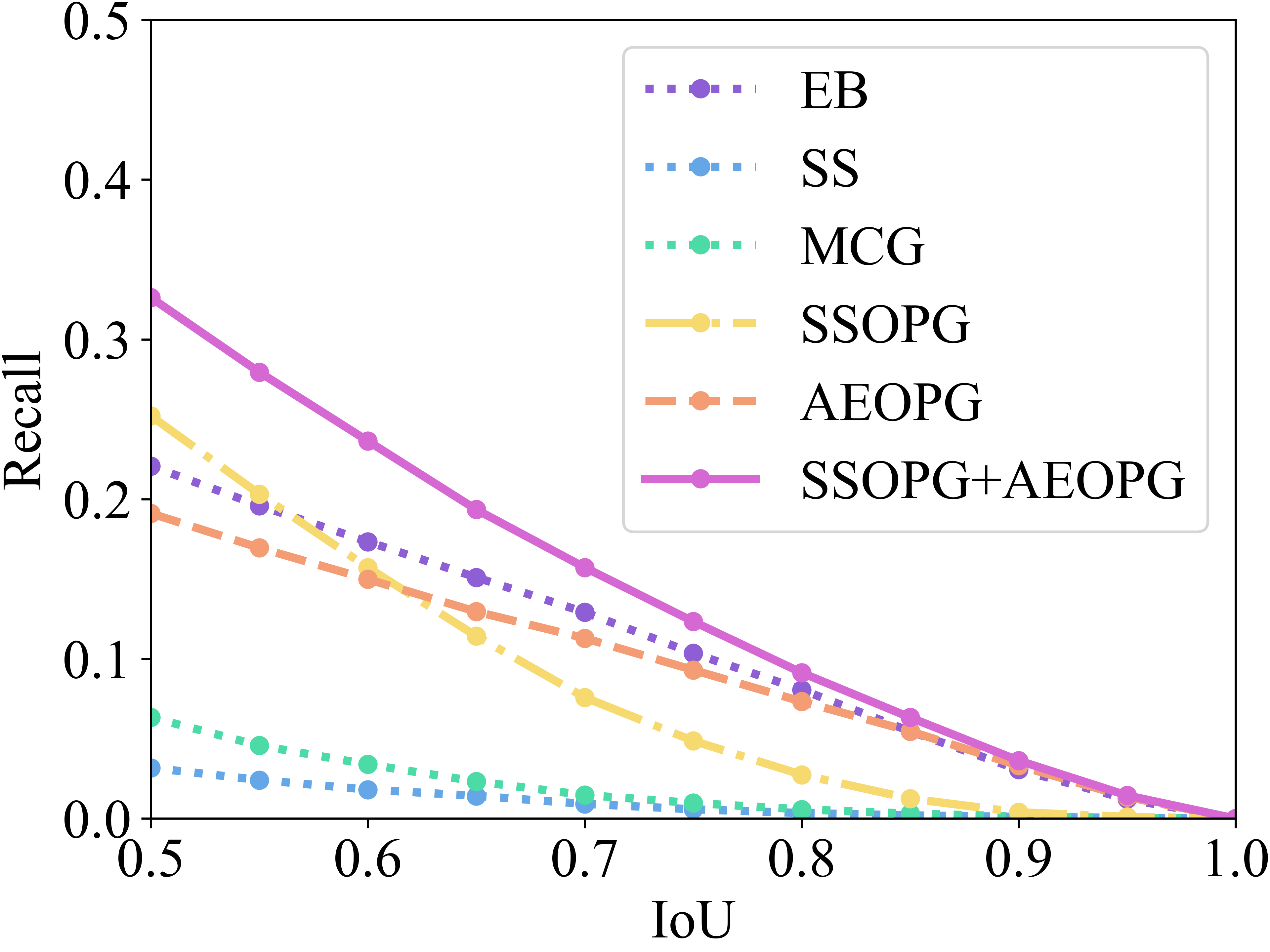}}%
                \stackunder[0pt]{$8$ proposals}{\includegraphics[width=0.20\textwidth]{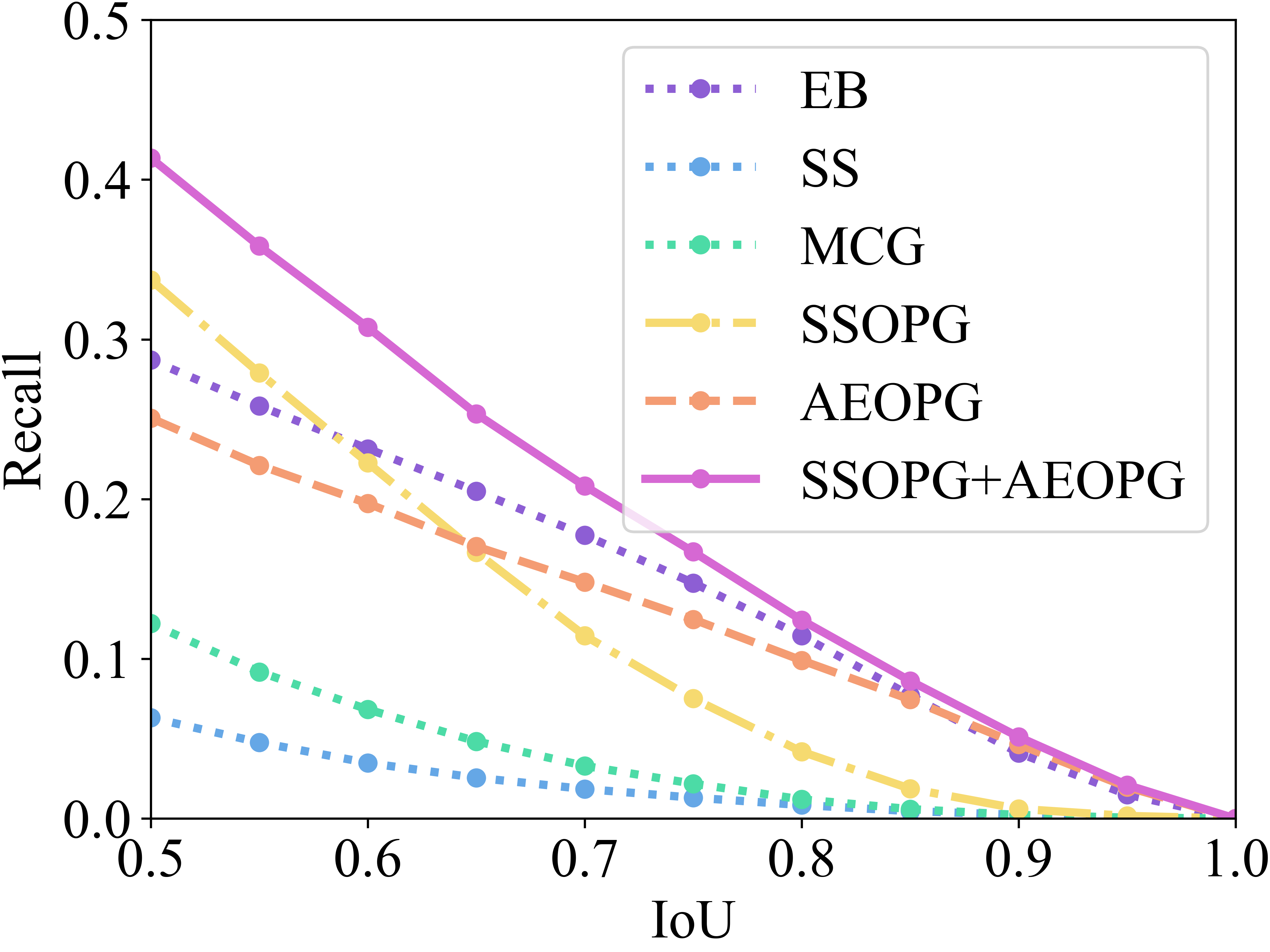}}%
                \stackunder[0pt]{$16$ proposals}{\includegraphics[width=0.20\textwidth]{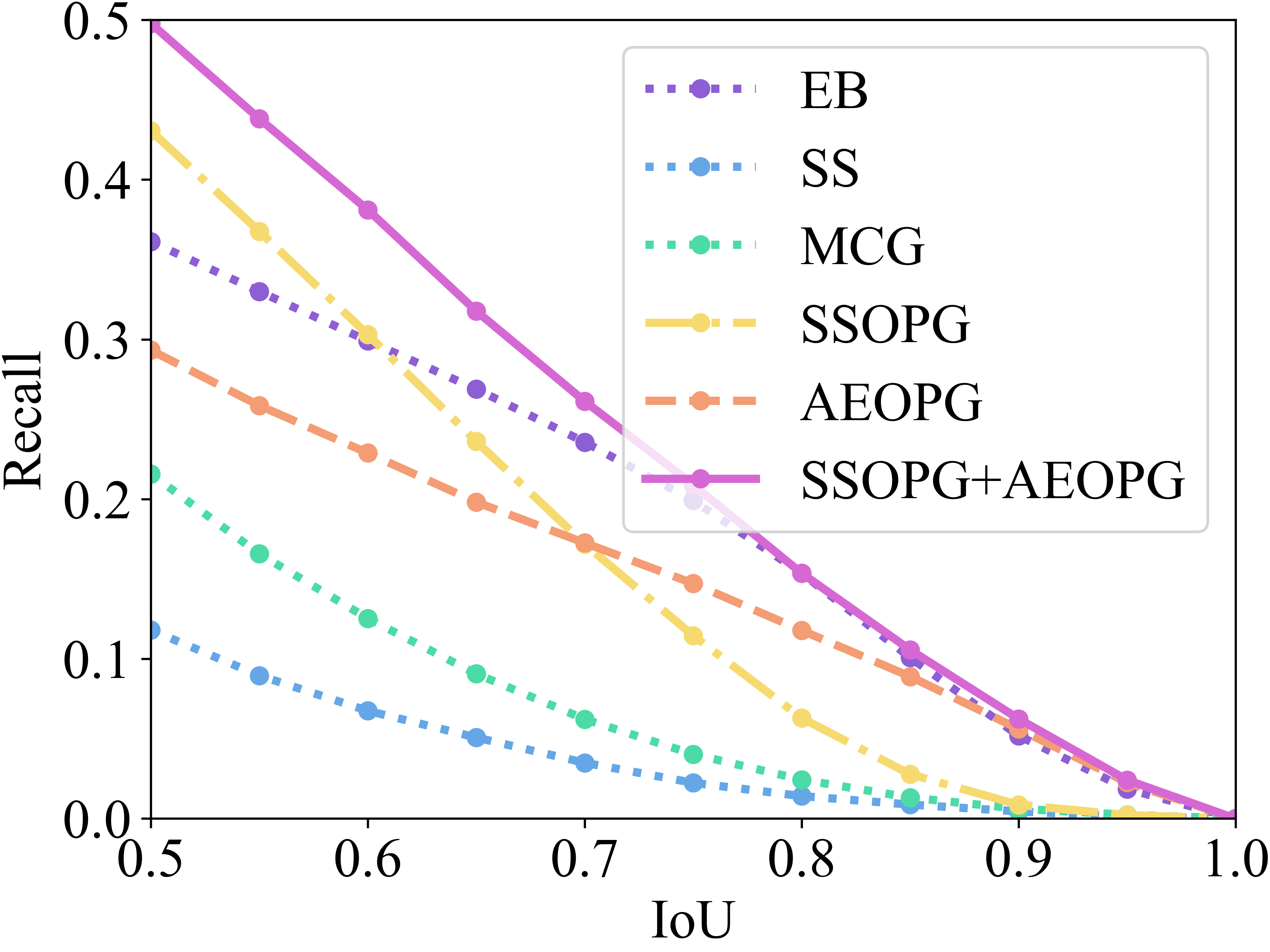}}%
                %\\
                \stackunder[0pt]{$32$ proposals}{\includegraphics[width=0.20\textwidth]{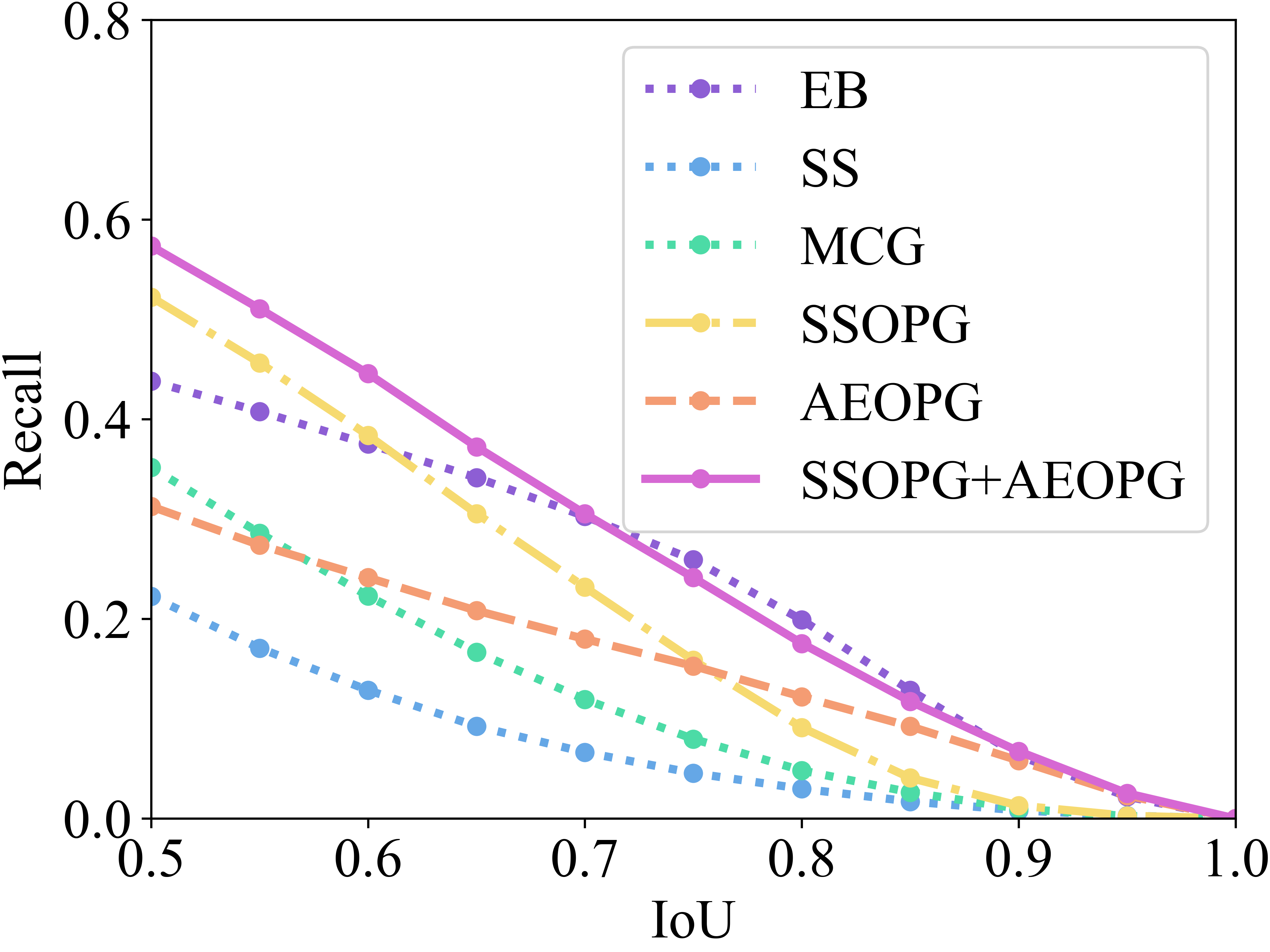}}%
                \\
                \stackunder[0pt]{$64$ proposals}{\includegraphics[width=0.20\textwidth]{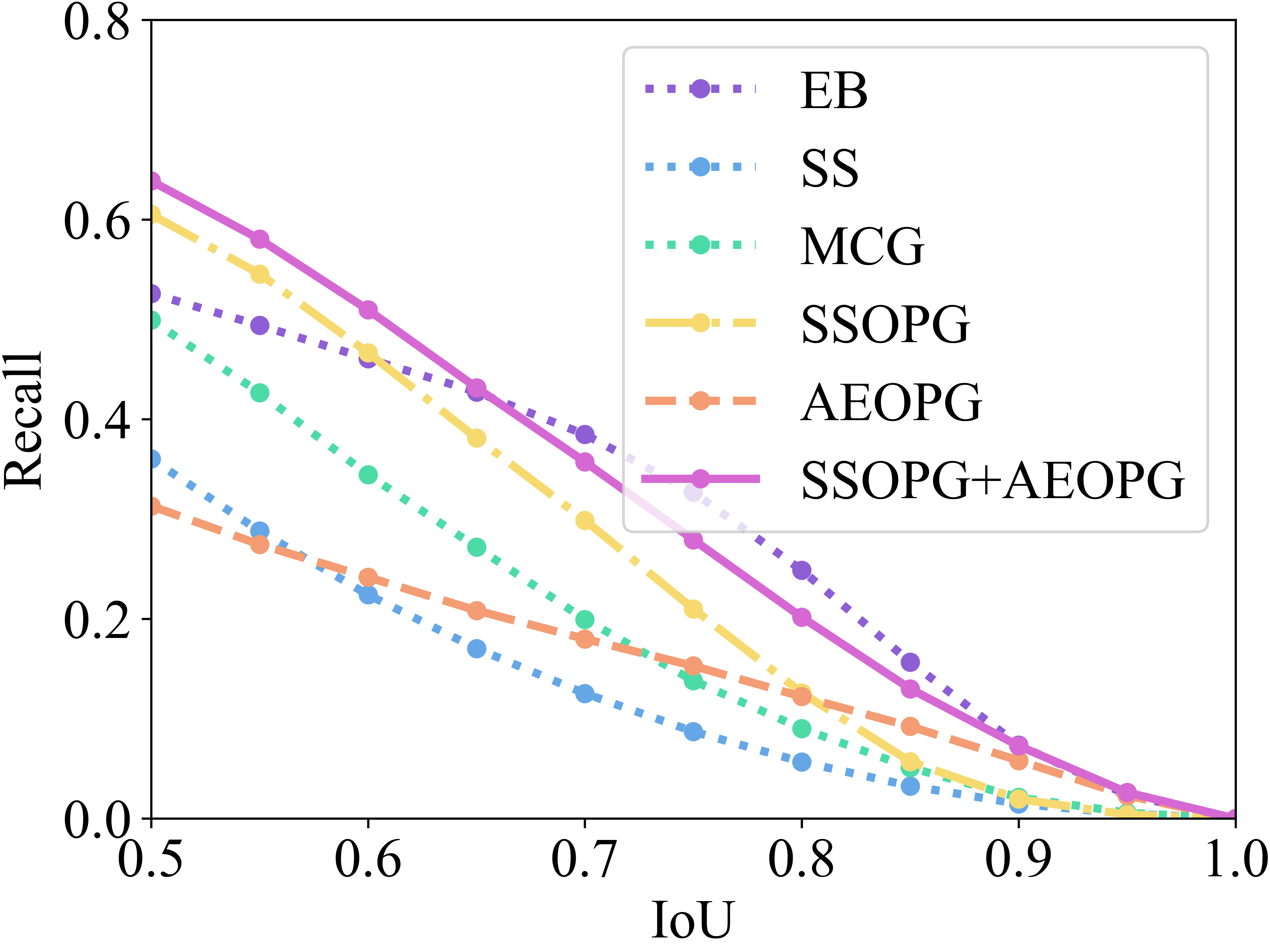}}%
                \stackunder[0pt]{$128$ proposals}{\includegraphics[width=0.20\textwidth]{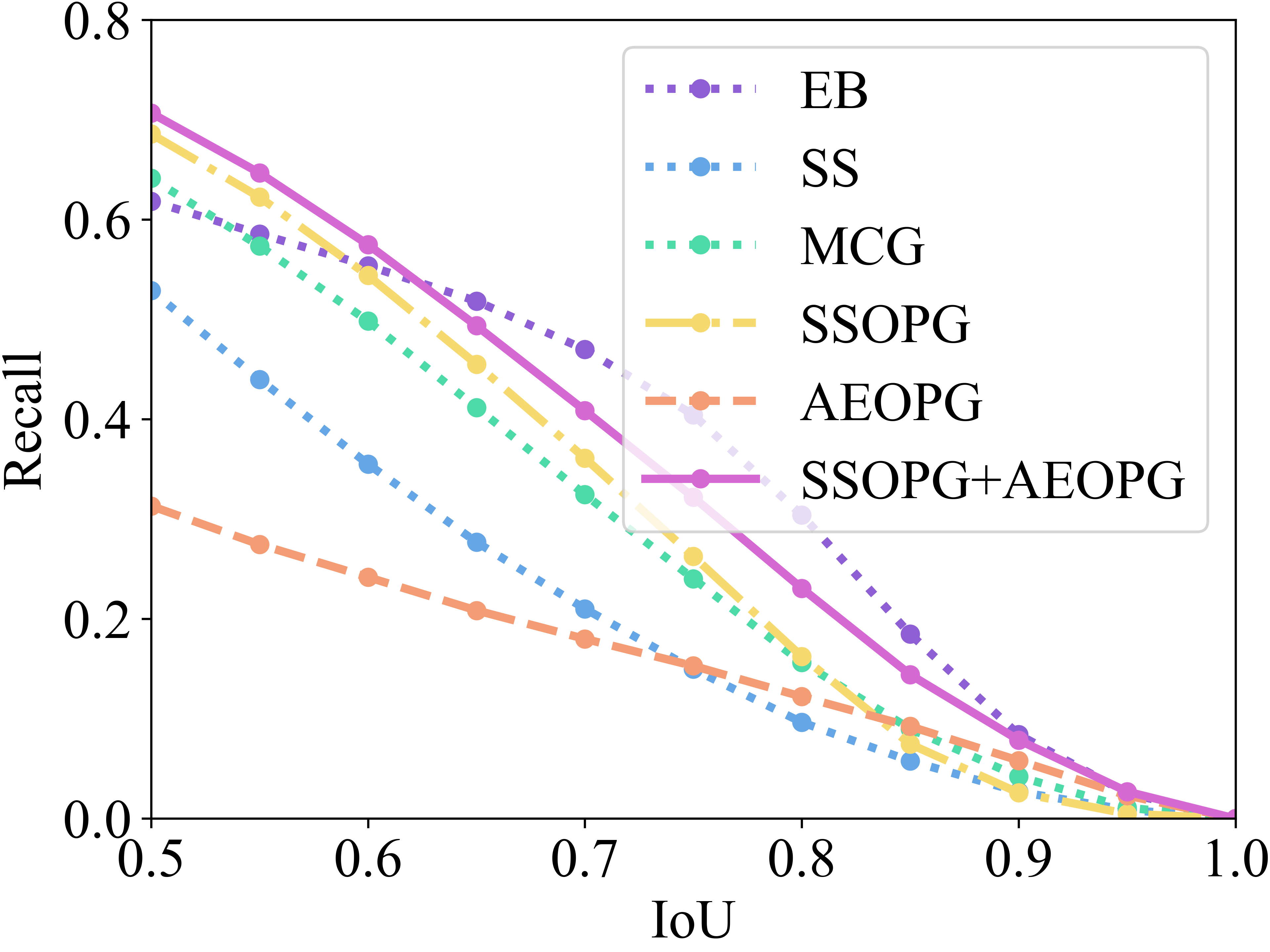}}%
                %\\
                \stackunder[0pt]{$256$ proposals}{\includegraphics[width=0.20\textwidth]{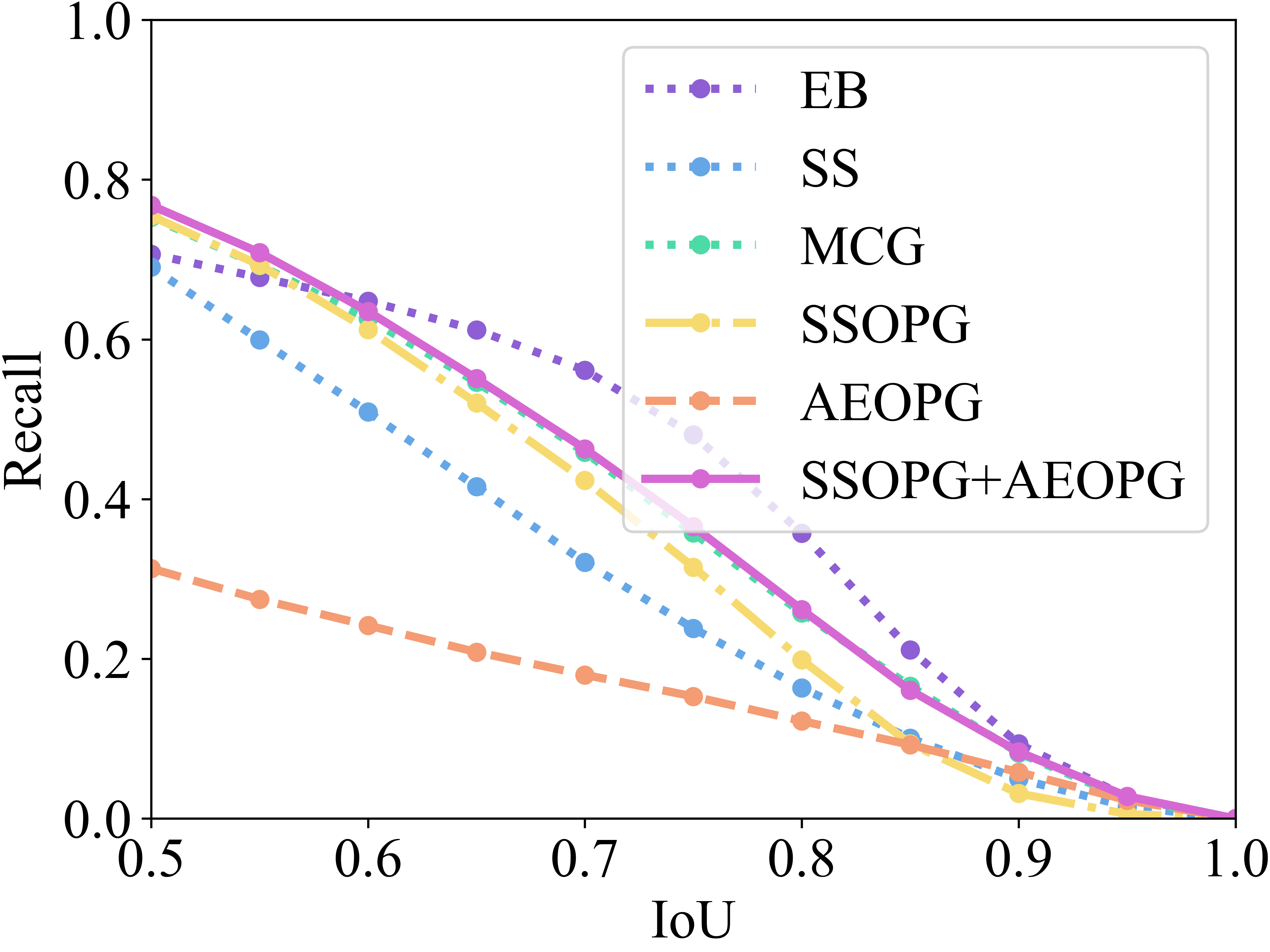}}%
                \stackunder[0pt]{$512$ proposals}{\includegraphics[width=0.20\textwidth]{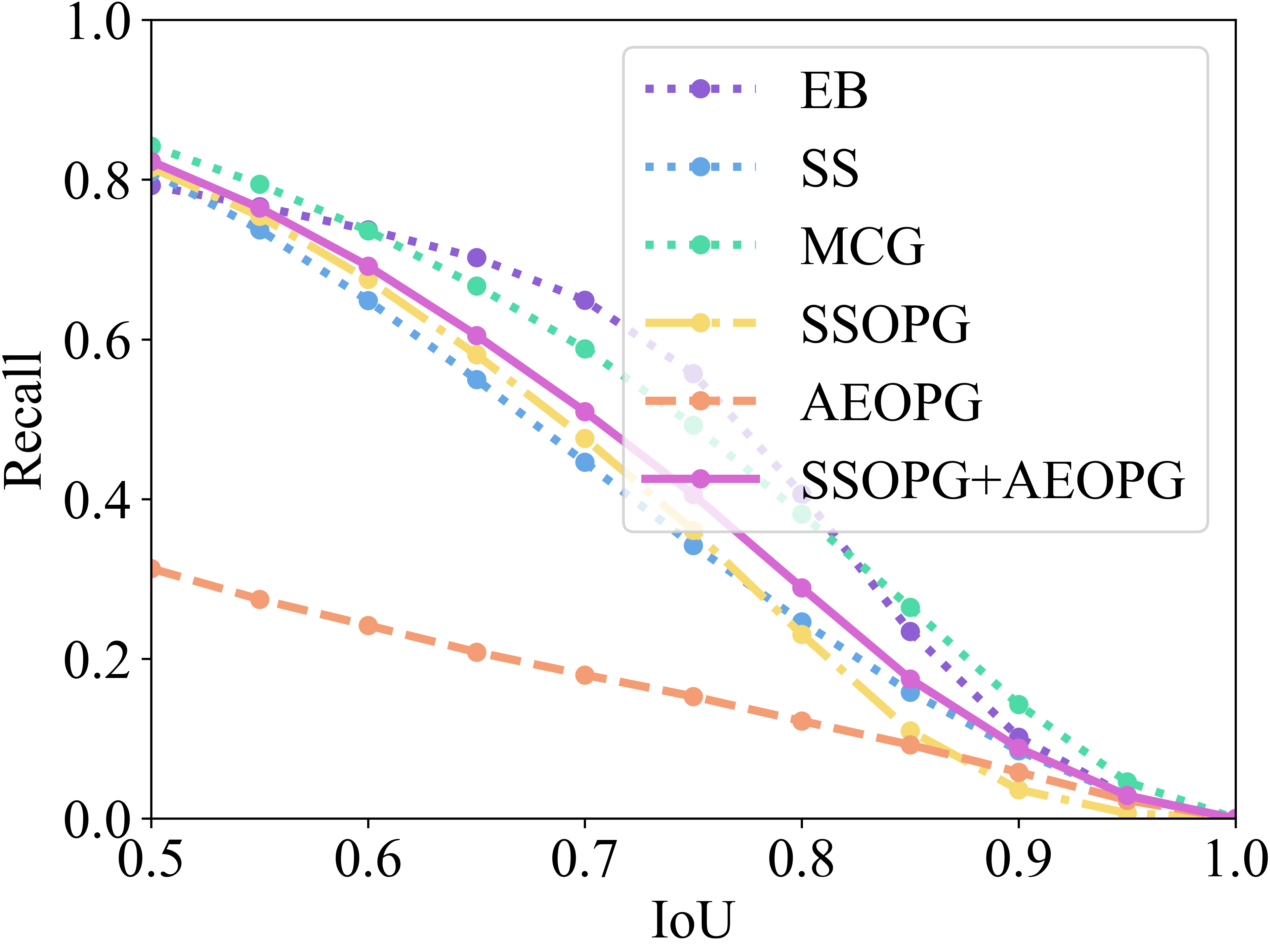}}%
                \stackunder[0pt]{$1,024$ proposals}{\includegraphics[width=0.20\textwidth]{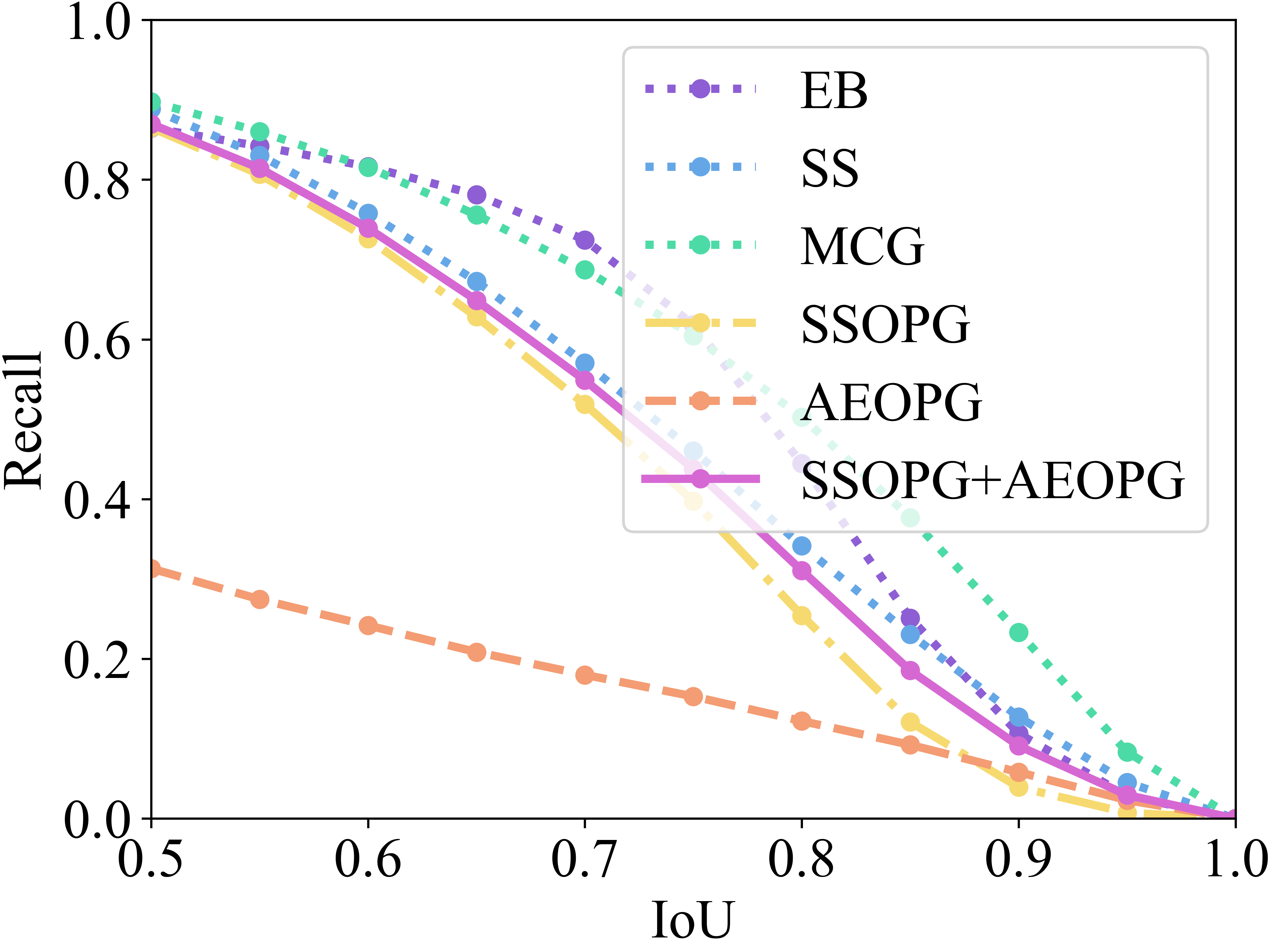}}%
                %\stackunder[0pt]{$2,048$ proposals}{\includegraphics[width=0.20\textwidth]{main/figure/proposal/recall_iou_2048.png}}%
            \end{center}
        \end{subfigure}
    \end{center}
    %    \vspace{-10pt}
    \caption{
        Recall \vs IoU overlap ratio of different object proposal methods on the PASCAL VOC 2007 \textit{test}.
    }
    %    \vspace{-20pt}
    \label{fig_vis_proposal_recall_iou}
\end{figure*}
%%%%%%%%%%%%%%%%%%%%%%%%%%%%%%%%%%%%%%%%%%%%%%%
%%
\begin{figure*}[t]
    %    \vspace{-20pt}
    %\scriptsize
    \begin{center}
        \includegraphics[width=1.00\textwidth]{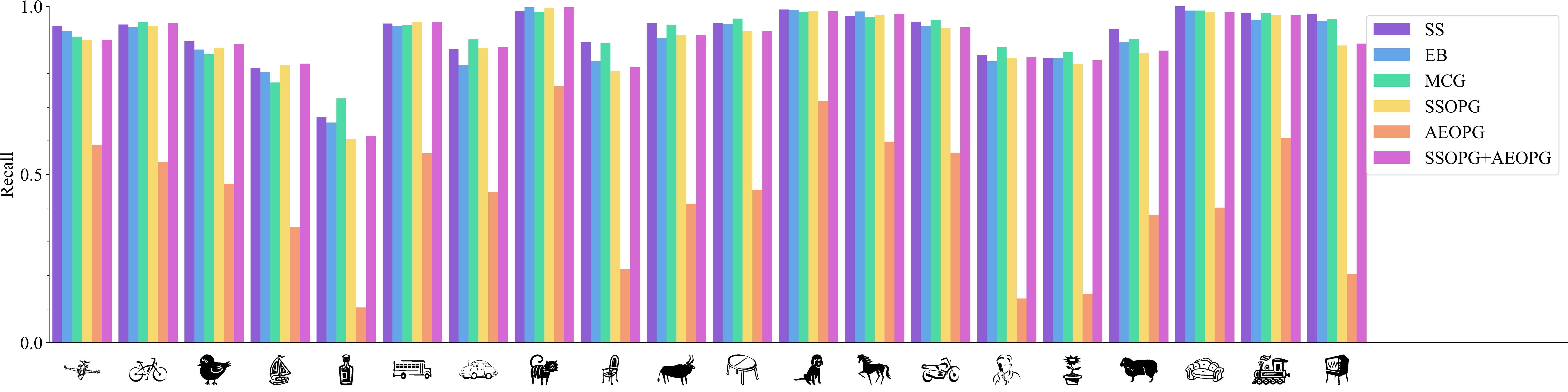}
    \end{center}
    %    \vspace{-10pt}
    \caption{
        Per-category recall of different object proposal methods on the PASCAL VOC 2007 \textit{test}.
    }
    %    \vspace{-20pt}
    \label{fig_vis_proposal_recall_cls}
    \vspace{-15pt}
\end{figure*}
%%%%%%%%%%%%%%%%%%%%%%%%%%%%%%%%%%%%%%%%%
%%
\subsection{Upper-bound Performance}
We further analyze the upper-bound performance of WSOD methods with ground-truth bounding boxes~(GT-bbox Known) setting in Tab.~\ref{table_voc2007voc2012mscoco_sota_gt}.
The upper bound of recent state-of-the-art methods, \eg, works in~\cite{Yanga,Ren2020}, are Fast R-CNN~\cite{FASTRCNN}, as they all heavily rely on external object proposal algorithms.
Different from them, \OurMethod learns to generate object proposals and has upper-bound performance with Faster R-CNN~\cite{FASTERRCNN}.
We further introduce a regression-disentangled learning setting to decouple proposal classification and regression tasks.
In detail, we remove the class annotation from the ground-truth bounding-box labels during training, termed Cls-agnostic GT-bbox Known.
Thus, WSOD still needs to capture coarse object location, while predicted boxes have ground-truth regression supervision to fine-tune themselves.
As shown in the bottom part of Tab.~\ref{table_voc2007voc2012mscoco_sota_gt}, the performance of \OurMethod approaches its upper bound of GT-bbox Known, which also demonstrates that \OurMethod has the ability to achieve fully-supervised accuracy.
We find that the accurate bounding-box localization is one of the main obstacles to reducing the performance gap between \OurMethod and its fully supervised counterpart.
%%%%%%%%%%%%%%%%%%%%%%%%%%%%%%%%%%%%%%%%%%%%%%%%%%%%%
\subsection{Object Proposals in WSOD}
We analyze the recall of object proposals for different methods at different IoU ratios with ground-truth boxes.
Fig.~\ref{fig_vis_proposal_recall_iou} shows that the SSOPG suppresses selective search~\cite{Uijlings2013}~(SS), edge boxes~\cite{Zitnick2014}~(EB) and multiscale combinatorial grouping~\cite{APBMM2014}~(MCG) at IoU $0.5$ in few proposal regimes, \ie, $4 \sim 64$ candidates per image.
At regimes of $256 \sim 1,024$, SSOPG practically achieves similar proposal quality as SS, but has lower quality than EB and MCG.
One possible reason is that SSOPG is overfitting to noisy pseudo labels, which is biased with true distribution in self-supervised framework.
We also observe that the recall of AEOPG is saturated at $32$ proposals per image.
The average proposal number generated by AEOPG is $16.0$ for each image, which reaches $31.4\%$ recall at IoU $0.5$.
A simple ensemble of SSOPG and AEOPG~(SSOPG+AEOPG) achieves consistent improvements across different numbers of proposal regimes.
It demonstrates that SSOPG and AEOPG learn complementary knowledge to hypothesize object locations.
We further investigate the per-category proposal recall at IoU $0.5$ with $1,024$ candidates per image in Fig.~\ref{fig_vis_proposal_recall_cls}.
The ensemble of SSOPG and AEOPG improves the recall of most categories, which achieves more than $95\%$ recall for categories \textit{bicycle}, \textit{bus}, \textit{cat}, \textit{dog}, \textit{horse}, \textit{sofa} and \textit{train} in PASCAL VOC.
The proposed generators obtain the lowest recall for category \textit{bottle} due to its small size.
For most categories, ``SSOPG+AEOPG'' has competitive performance compared to those well-designed traditional proposal methods.
%%%%%%%%%%%%%%%%%%%%%%%%%%%%%%%%%%%%%%%%%%%%%%%%%%
%%
\begin{figure*}[t]
    \small
    \begin{center}
        \begin{subfigure}[b]{1.0\textwidth}
            
            \stackunder[1pt]{$\mathrm{AEOPG}$}
            {\includegraphics[width=0.105\textwidth, height=0.13\textwidth]{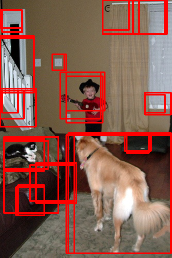}}%
            \hspace{1pt}
            \stackunder[-1pt]{$\mathcal{E}(F)$}{\includegraphics[width=0.105\textwidth, height=0.13\textwidth]{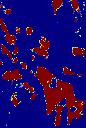}}%
            \hspace{1pt}
            \stackunder[1pt]{$F^q$}{\includegraphics[width=0.105\textwidth, height=0.13\textwidth]{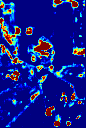}}%
            \hspace{1pt}
            \stackunder[1pt]{$\mathrm{AEOPG}$}{\includegraphics[width=0.105\textwidth, height=0.13\textwidth]{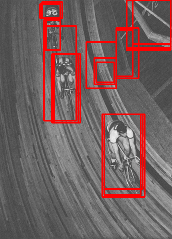}}%
            \hspace{1pt}
            \stackunder[-1pt]{$\mathcal{E}(F)$}{\includegraphics[width=0.105\textwidth, height=0.13\textwidth]{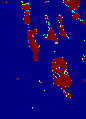}}%
            \hspace{1pt}
            \stackunder[1pt]{$F^q$}{\includegraphics[width=0.105\textwidth, height=0.13\textwidth]{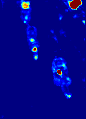}}%
            \hspace{1pt}
            \stackunder[1pt]{$\mathrm{AEOPG}$}{\includegraphics[width=0.105\textwidth, height=0.13\textwidth]{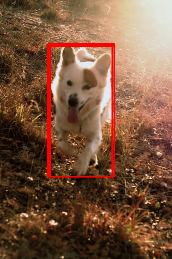}}%
            \hspace{1pt}
            \stackunder[-1pt]{$\mathcal{E}(F)$}{\includegraphics[width=0.105\textwidth, height=0.13\textwidth]{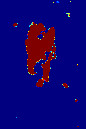}}%
            \hspace{1pt}
            \stackunder[1pt]{$F^q$}{\includegraphics[width=0.105\textwidth, height=0.13\textwidth]{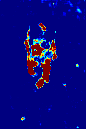}}%
            
            \vspace{2pt}
            
            \includegraphics[width=0.105\textwidth, height=0.06\textwidth]{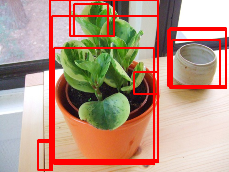}%
            \hspace{1pt}
            \includegraphics[width=0.105\textwidth, height=0.06\textwidth]{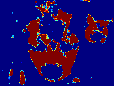}%
            \hspace{1pt}
            \includegraphics[width=0.105\textwidth, height=0.06\textwidth]{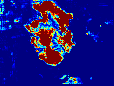}%
            \hspace{1pt}
            \includegraphics[width=0.105\textwidth, height=0.06\textwidth]{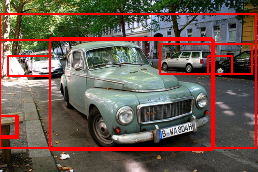}%
            \hspace{1pt}
            \includegraphics[width=0.105\textwidth, height=0.06\textwidth]{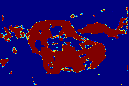}%
            \hspace{1pt}
            \includegraphics[width=0.105\textwidth, height=0.06\textwidth]{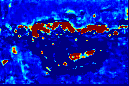}%
            \hspace{1pt}
            \includegraphics[width=0.105\textwidth, height=0.06\textwidth]{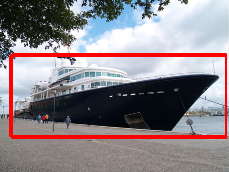}%
            \hspace{1pt}
            \includegraphics[width=0.105\textwidth, height=0.06\textwidth]{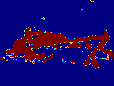}%
            \hspace{1pt}
            \includegraphics[width=0.105\textwidth, height=0.06\textwidth]{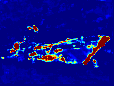}%

            \vspace{2pt}
            
            \includegraphics[width=0.105\textwidth, height=0.06\textwidth]{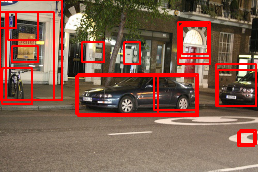}%
            \hspace{1pt}
            \includegraphics[width=0.105\textwidth, height=0.06\textwidth]{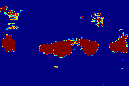}%
            \hspace{1pt}
            \includegraphics[width=0.105\textwidth, height=0.06\textwidth]{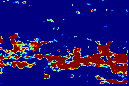}%
            \hspace{1pt}
            \includegraphics[width=0.105\textwidth, height=0.06\textwidth]{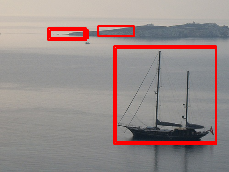}%
            \hspace{1pt}
            \includegraphics[width=0.105\textwidth, height=0.06\textwidth]{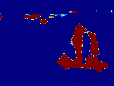}%
            \hspace{1pt}
            \includegraphics[width=0.105\textwidth, height=0.06\textwidth]{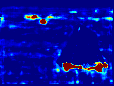}%
            \hspace{1pt}
            \includegraphics[width=0.105\textwidth, height=0.06\textwidth]{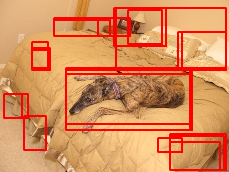}%
            \hspace{1pt}
            \includegraphics[width=0.105\textwidth, height=0.06\textwidth]{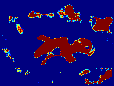}%
            \hspace{1pt}
            \includegraphics[width=0.105\textwidth, height=0.06\textwidth]{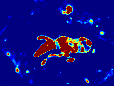}%
            
            \vspace{2pt}
            
            \includegraphics[width=0.105\textwidth, height=0.06\textwidth, height=0.06\textwidth]{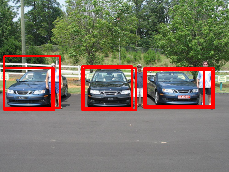}%
            \hspace{1pt}
            \includegraphics[width=0.105\textwidth, height=0.06\textwidth]{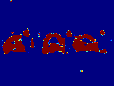}%
            \hspace{1pt}
            \includegraphics[width=0.105\textwidth, height=0.06\textwidth]{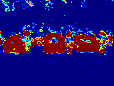}%
            \hspace{1pt}
            \includegraphics[width=0.105\textwidth, height=0.06\textwidth]{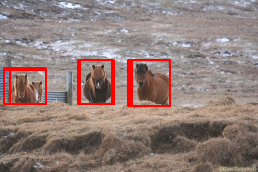}%
            \hspace{1pt}
            \includegraphics[width=0.105\textwidth, height=0.06\textwidth]{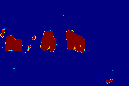}%
            \hspace{1pt}
            \includegraphics[width=0.105\textwidth, height=0.06\textwidth]{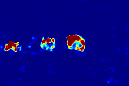}%
            \hspace{1pt}
            \includegraphics[width=0.105\textwidth, height=0.06\textwidth]{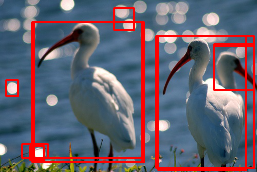}%
            \hspace{1pt}
            \includegraphics[width=0.105\textwidth, height=0.06\textwidth]{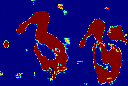}%
            \hspace{1pt}
            \includegraphics[width=0.105\textwidth, height=0.06\textwidth]{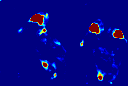}%

            \vspace{2pt}
            
            \includegraphics[width=0.105\textwidth, height=0.06\textwidth]{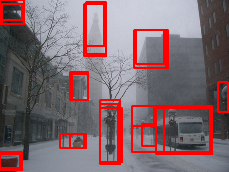}%
            \hspace{1pt}
            \includegraphics[width=0.105\textwidth, height=0.06\textwidth]{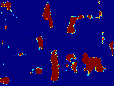}%
            \hspace{1pt}
            \includegraphics[width=0.105\textwidth, height=0.06\textwidth]{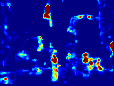}%
            \hspace{1pt}
            \includegraphics[width=0.105\textwidth, height=0.06\textwidth]{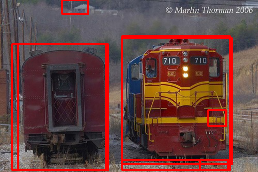}%
            \hspace{1pt}
            \includegraphics[width=0.105\textwidth, height=0.06\textwidth]{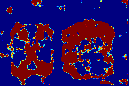}%
            \hspace{1pt}
            \includegraphics[width=0.105\textwidth, height=0.06\textwidth]{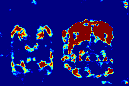}%
            \hspace{1pt}
            \includegraphics[width=0.105\textwidth, height=0.06\textwidth]{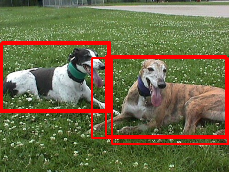}%
            \hspace{1pt}
            \includegraphics[width=0.105\textwidth, height=0.06\textwidth]{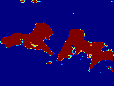}%
            \hspace{1pt}
            \includegraphics[width=0.105\textwidth, height=0.06\textwidth]{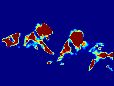}%
            
            \vspace{2pt}
            
            \includegraphics[width=0.105\textwidth, height=0.06\textwidth]{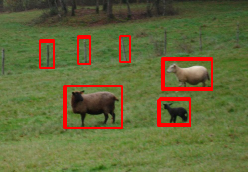}%
            \hspace{1pt}
            \includegraphics[width=0.105\textwidth, height=0.06\textwidth]{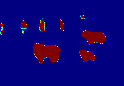}%
            \hspace{1pt}
            \includegraphics[width=0.105\textwidth, height=0.06\textwidth]{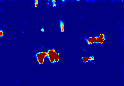}%
            \hspace{1pt}
            \includegraphics[width=0.105\textwidth, height=0.06\textwidth]{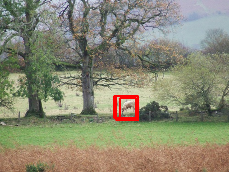}%
            \hspace{1pt}
            \includegraphics[width=0.105\textwidth, height=0.06\textwidth]{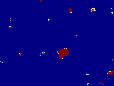}%
            \hspace{1pt}
            \includegraphics[width=0.105\textwidth, height=0.06\textwidth]{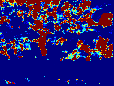}%
            \hspace{1pt}
            \includegraphics[width=0.105\textwidth, height=0.06\textwidth]{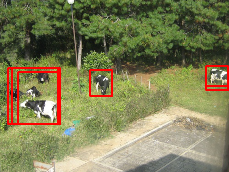}%
            \hspace{1pt}
            \includegraphics[width=0.105\textwidth, height=0.06\textwidth]{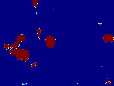}%
            \hspace{1pt}
            \includegraphics[width=0.105\textwidth, height=0.06\textwidth]{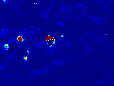}%
            
            \vspace{2pt}

            \includegraphics[width=0.105\textwidth, height=0.06\textwidth]{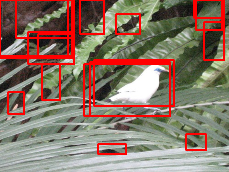}%
            \hspace{1pt}
            \includegraphics[width=0.105\textwidth, height=0.06\textwidth]{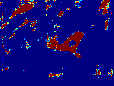}%
            \hspace{1pt}
            \includegraphics[width=0.105\textwidth, height=0.06\textwidth]{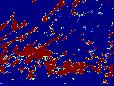}%
            \hspace{1pt}
            \includegraphics[width=0.105\textwidth, height=0.06\textwidth]{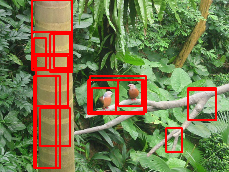}%
            \hspace{1pt}
            \includegraphics[width=0.105\textwidth, height=0.06\textwidth]{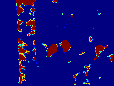}%
            \hspace{1pt}
            \includegraphics[width=0.105\textwidth, height=0.06\textwidth]{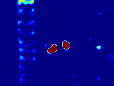}%
            \hspace{1pt}
            \includegraphics[width=0.105\textwidth, height=0.06\textwidth]{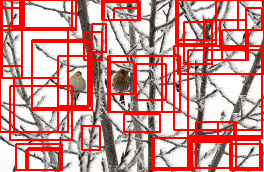}%
            \hspace{1pt}
            \includegraphics[width=0.105\textwidth, height=0.06\textwidth]{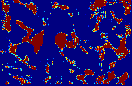}%
            \hspace{1pt}
            \includegraphics[width=0.105\textwidth, height=0.06\textwidth]{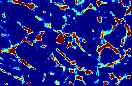}%
            
            \vspace{2pt}

            \includegraphics[width=0.105\textwidth, height=0.06\textwidth]{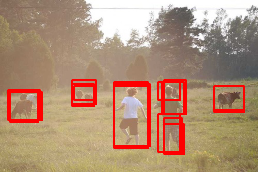}%
            \hspace{1pt}
            \includegraphics[width=0.105\textwidth, height=0.06\textwidth]{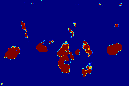}%
            \hspace{1pt}
            \includegraphics[width=0.105\textwidth, height=0.06\textwidth]{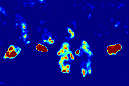}%
            \hspace{1pt}
            \includegraphics[width=0.105\textwidth, height=0.06\textwidth]{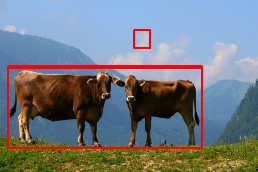}%
            \hspace{1pt}
            \includegraphics[width=0.105\textwidth, height=0.06\textwidth]{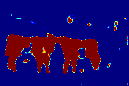}%
            \hspace{1pt}
            \includegraphics[width=0.105\textwidth, height=0.06\textwidth]{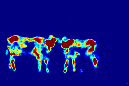}%
            \hspace{1pt}
            \includegraphics[width=0.105\textwidth, height=0.06\textwidth]{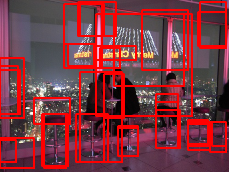}%
            \hspace{1pt}
            \includegraphics[width=0.105\textwidth, height=0.06\textwidth]{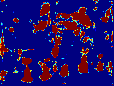}%
            \hspace{1pt}
            \includegraphics[width=0.105\textwidth, height=0.06\textwidth]{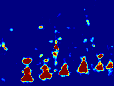}%

        \end{subfigure}
    \end{center}
    \caption{
        Visualization of object proposals, encoded representations~$\mathcal{E}(F)$ and low-rank feature maps~$F^q$ in AEOPG.
    }
    \label{fig_vis_aeopg}
    \vspace{-20pt}
\end{figure*}

\begin{figure*}[t]
    %    \vspace{-20pt}
    %\scriptsize
    \small
    \begin{center}
        \begin{subfigure}[b]{1.0\textwidth}
            \begin{center}
                
                \stackunder[1pt]{Ground Truths}{\includegraphics[width=0.15\textwidth, height=0.10\textwidth]{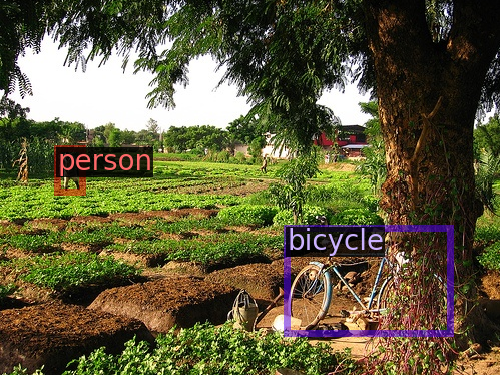}}%
                \hspace{1pt}
                \stackunder[1pt]{WSDDN}{\includegraphics[width=0.15\textwidth, height=0.10\textwidth]{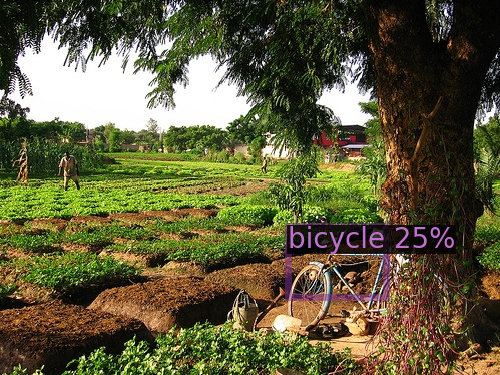}}%
                \hspace{1pt}
                \stackunder[1pt]{PCL}{\includegraphics[width=0.15\textwidth, height=0.10\textwidth]{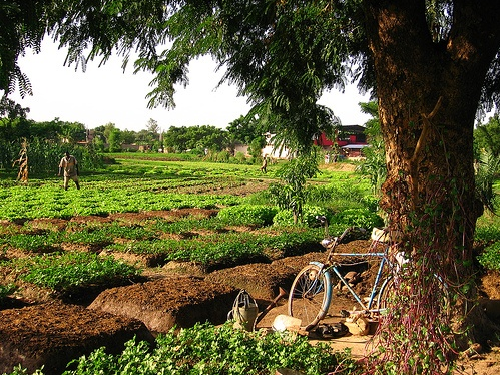}}%
                \hspace{1pt}
                \stackunder[1pt]{UWSOD}{\includegraphics[width=0.15\textwidth, height=0.10\textwidth]{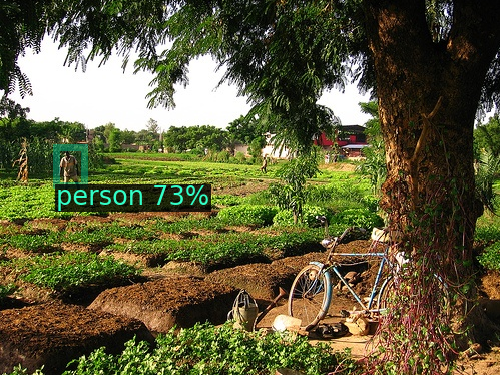}}%
                \hspace{1pt}
                \stackunder[1pt]{\OurMethod}{\includegraphics[width=0.15\textwidth, height=0.10\textwidth]{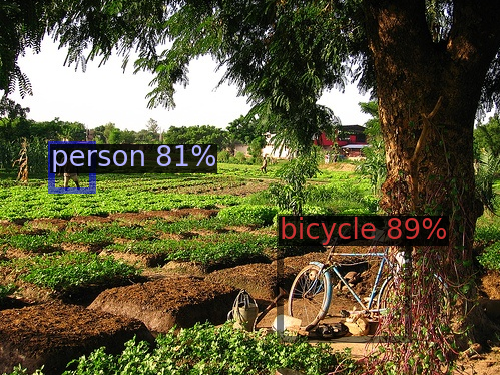}}%
                \hspace{1pt}
                \stackunder[1pt]{\OurMethodx}{\includegraphics[width=0.15\textwidth, height=0.10\textwidth]{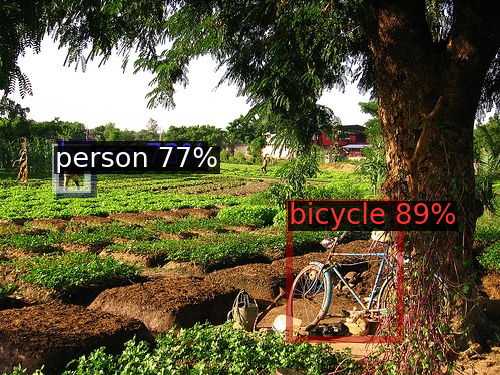}}%
                
                \vspace{2pt}
                
                \includegraphics[width=0.15\textwidth, height=0.10\textwidth]{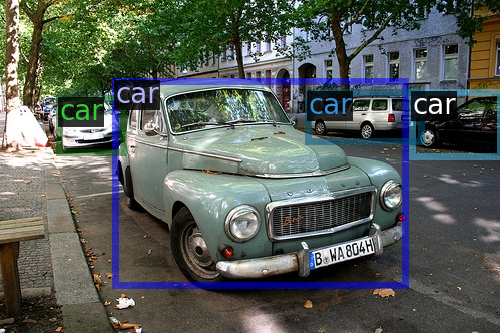}%
                \hspace{1pt}
                \includegraphics[width=0.15\textwidth, height=0.10\textwidth]{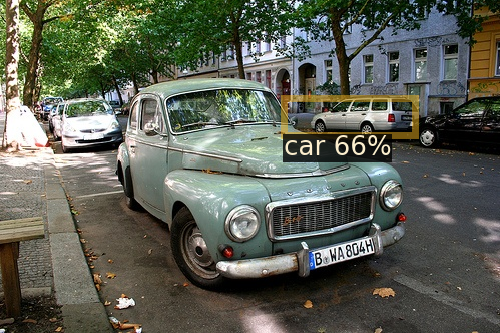}%
                \hspace{1pt}
                \includegraphics[width=0.15\textwidth, height=0.10\textwidth]{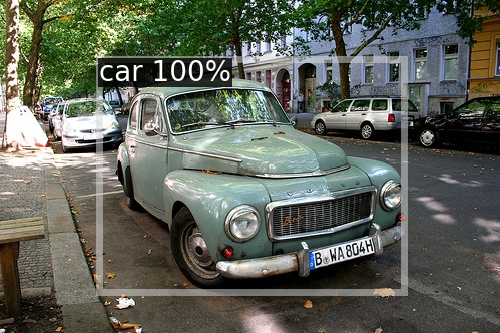}%
                \hspace{1pt}
                \includegraphics[width=0.15\textwidth, height=0.10\textwidth]{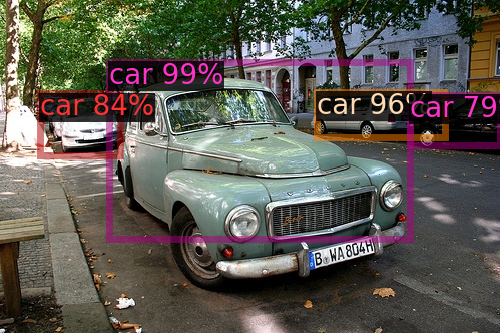}%
                \hspace{1pt}
                \includegraphics[width=0.15\textwidth, height=0.10\textwidth]{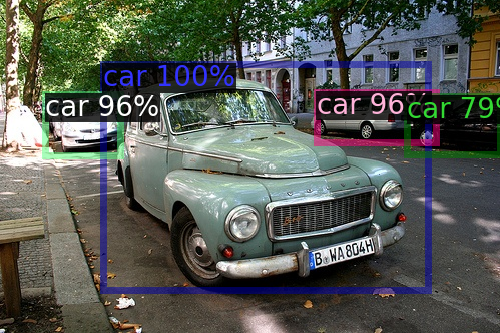}%
                \hspace{1pt}
                \includegraphics[width=0.15\textwidth, height=0.10\textwidth]{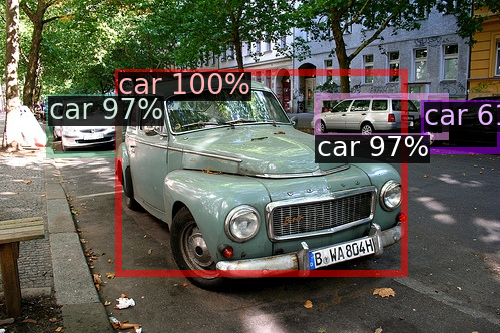}%

                \vspace{2pt}
                
                \includegraphics[width=0.15\textwidth, height=0.10\textwidth]{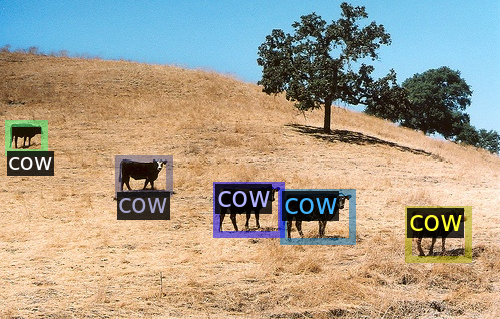}%
                \hspace{1pt}
                \includegraphics[width=0.15\textwidth, height=0.10\textwidth]{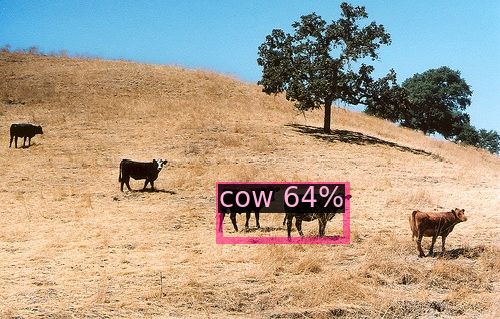}%
                \hspace{1pt}
                \includegraphics[width=0.15\textwidth, height=0.10\textwidth]{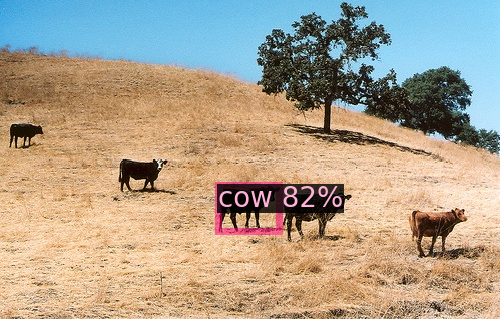}%
                \hspace{1pt}
                \includegraphics[width=0.15\textwidth, height=0.10\textwidth]{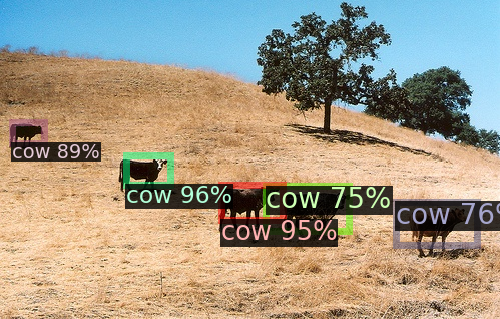}%
                \hspace{1pt}
                \includegraphics[width=0.15\textwidth, height=0.10\textwidth]{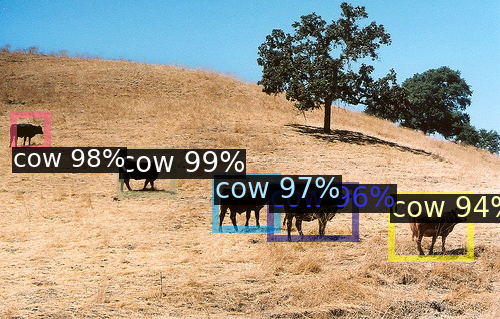}%
                \hspace{1pt}
                \includegraphics[width=0.15\textwidth, height=0.10\textwidth]{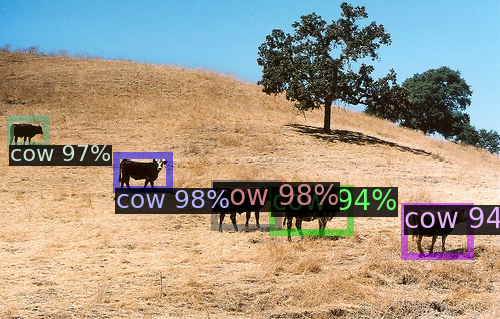}%
                
                \vspace{2pt}
                
                \includegraphics[width=0.15\textwidth, height=0.10\textwidth]{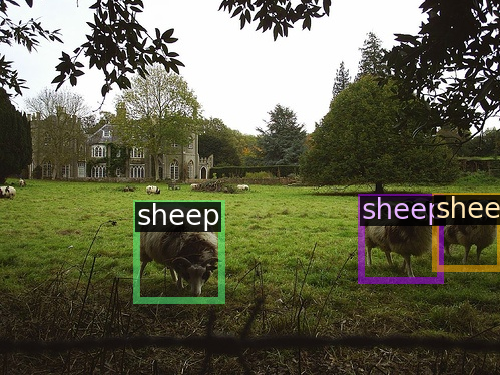}%
                \hspace{1pt}
                \includegraphics[width=0.15\textwidth, height=0.10\textwidth]{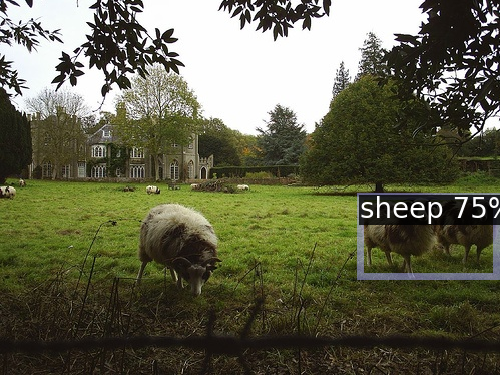}%
                \hspace{1pt}
                \includegraphics[width=0.15\textwidth, height=0.10\textwidth]{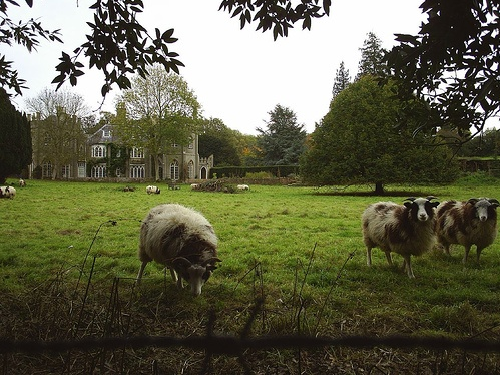}%
                \hspace{1pt}
                \includegraphics[width=0.15\textwidth, height=0.10\textwidth]{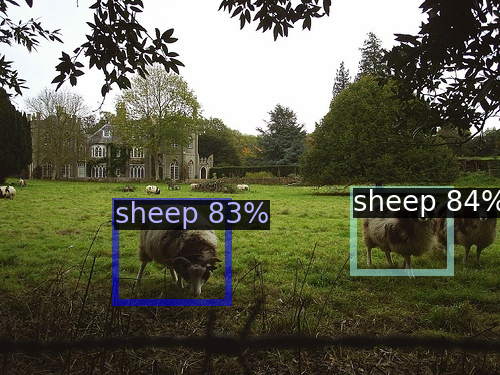}%
                \hspace{1pt}
                \includegraphics[width=0.15\textwidth, height=0.10\textwidth]{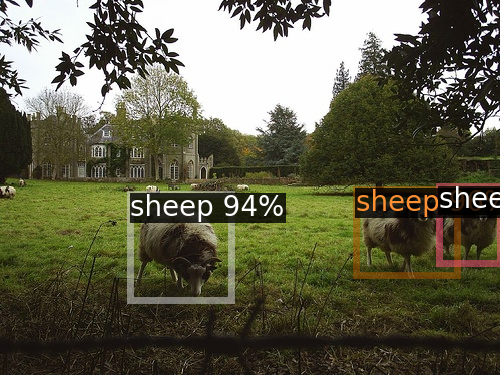}%
                \hspace{1pt}
                \includegraphics[width=0.15\textwidth, height=0.10\textwidth]{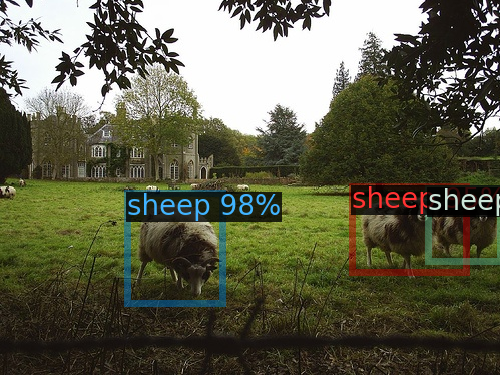}%

                %                \vspace{2pt}
                %                
                %                
                %                
                %                \includegraphics[width=0.15\textwidth, height=0.10\textwidth]{main/figure/result/huwsod_WSR_18_DC5_4x1_160e_20220318_/000059.jpg.gt.png}%
                %                \hspace{1pt}
                %                \includegraphics[width=0.15\textwidth, height=0.10\textwidth]{main/figure/result/wsddn_WSR_18_DC5_20220418_/000059.jpg.pred.png}%
                %                \hspace{1pt}
                %                \includegraphics[width=0.15\textwidth, height=0.10\textwidth]{main/figure/result/oicr_WSR_18_DC5_20220325_/000059.jpg.pred.png}%
                %                \hspace{1pt}
                %                \includegraphics[width=0.15\textwidth, height=0.10\textwidth]{main/figure/result/uwsod_WSR_18_DC5_20220211_105400_/000059.jpg.pred.png}%
                %                \hspace{1pt}
                %                \includegraphics[width=0.15\textwidth, height=0.10\textwidth]{main/figure/result/huwsod_WSR_18_DC5_4x1_160e_20220318_/000059.jpg.pred.png}%
                %                \hspace{1pt}
                %                \includegraphics[width=0.15\textwidth, height=0.10\textwidth]{main/figure/result/huwsod_poi_WSR_18_DC5_4x1_160e_20220318_/000059.jpg.pred.png}%

                \vspace{2pt}
                
                \includegraphics[width=0.15\textwidth, height=0.10\textwidth]{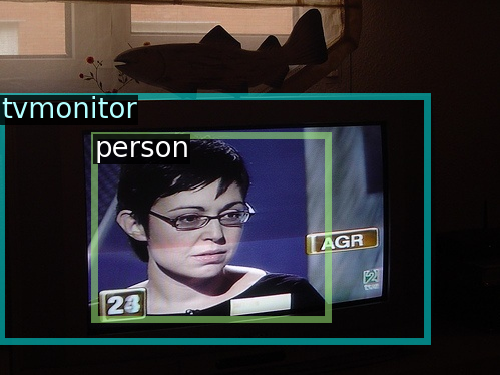}%
                \hspace{1pt}
                \includegraphics[width=0.15\textwidth, height=0.10\textwidth]{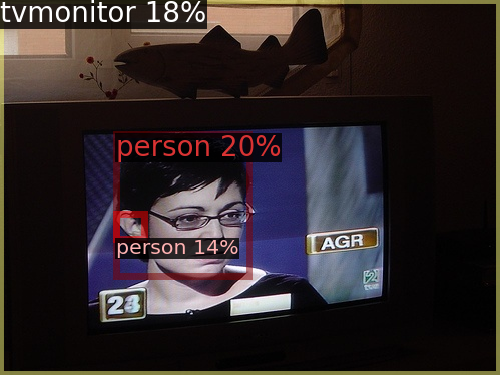}%
                \hspace{1pt}
                \includegraphics[width=0.15\textwidth, height=0.10\textwidth]{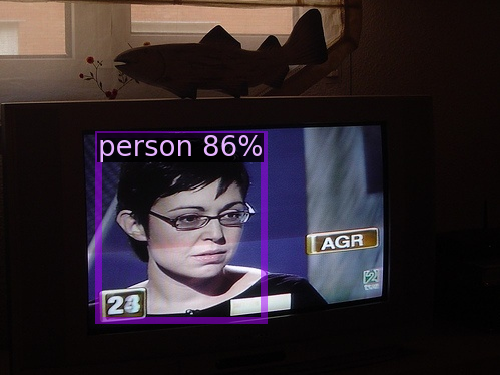}%
                \hspace{1pt}
                \includegraphics[width=0.15\textwidth, height=0.10\textwidth]{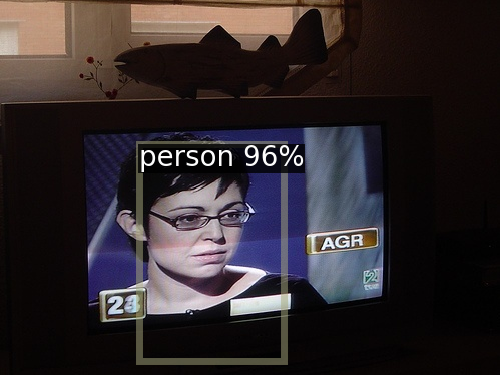}%
                \hspace{1pt}
                \includegraphics[width=0.15\textwidth, height=0.10\textwidth]{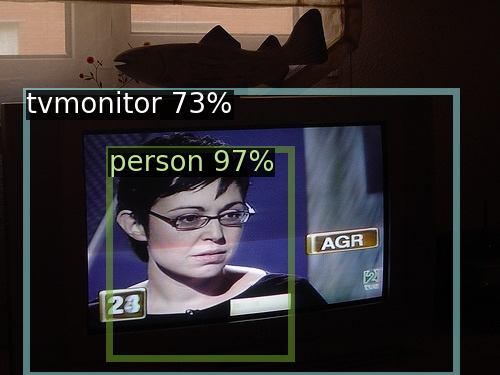}%
                \hspace{1pt}
                \includegraphics[width=0.15\textwidth, height=0.10\textwidth]{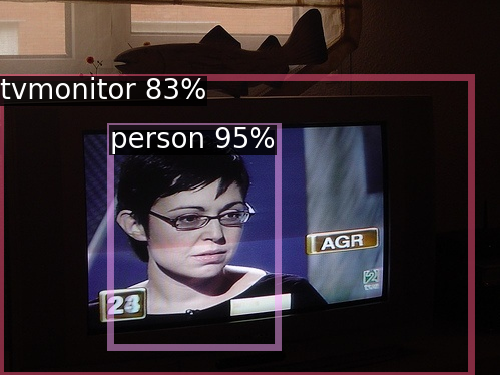}%

                \vspace{2pt}
                
                \includegraphics[width=0.15\textwidth, height=0.10\textwidth]{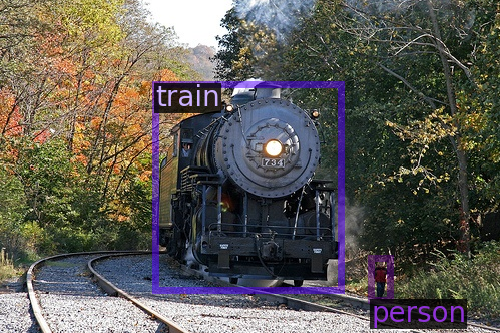}%
                \hspace{1pt}
                \includegraphics[width=0.15\textwidth, height=0.10\textwidth]{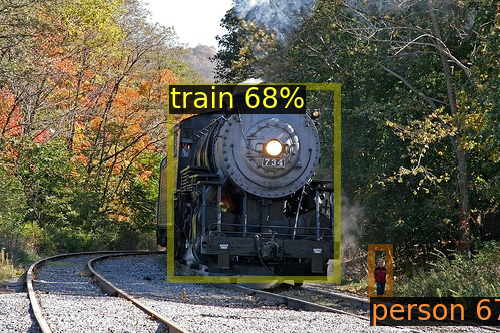}%
                \hspace{1pt}
                \includegraphics[width=0.15\textwidth, height=0.10\textwidth]{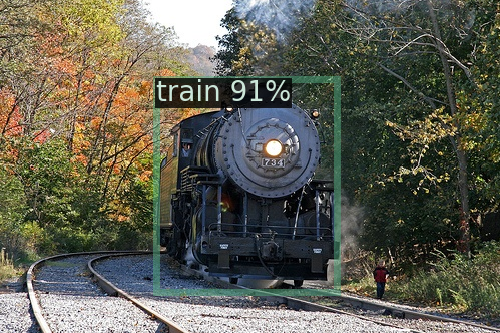}%
                \hspace{1pt}
                \includegraphics[width=0.15\textwidth, height=0.10\textwidth]{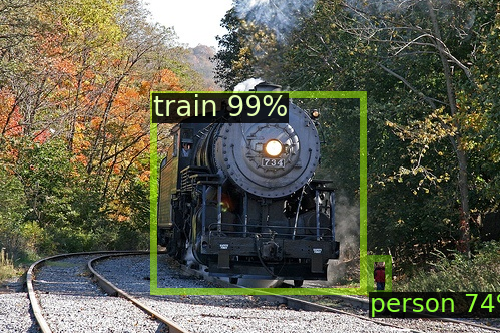}%
                \hspace{1pt}
                \includegraphics[width=0.15\textwidth, height=0.10\textwidth]{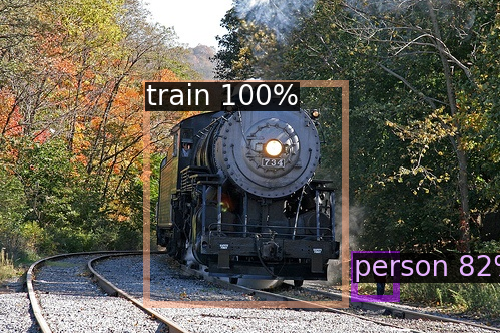}%
                \hspace{1pt}
                \includegraphics[width=0.15\textwidth, height=0.10\textwidth]{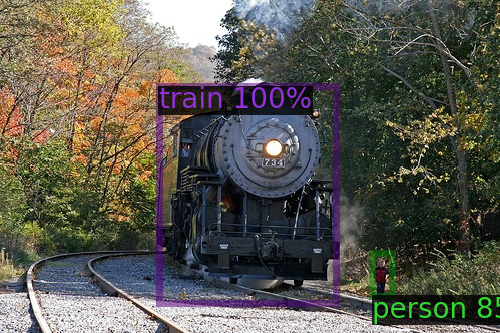}%

                \vspace{2pt}
                
                \includegraphics[width=0.15\textwidth, height=0.10\textwidth]{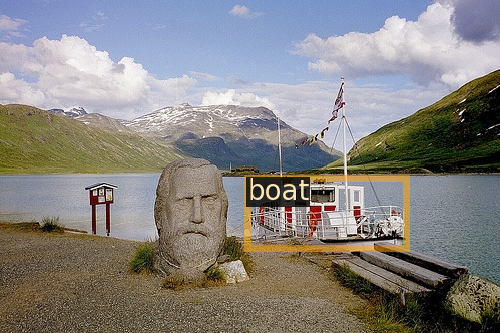}%
                \hspace{1pt}
                \includegraphics[width=0.15\textwidth, height=0.10\textwidth]{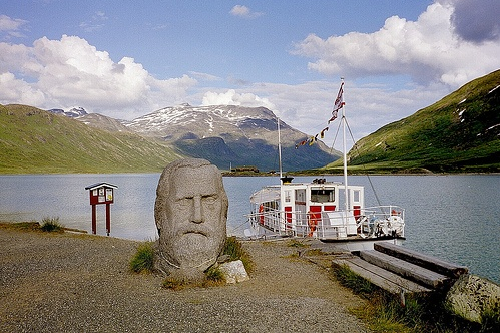}%
                \hspace{1pt}
                \includegraphics[width=0.15\textwidth, height=0.10\textwidth]{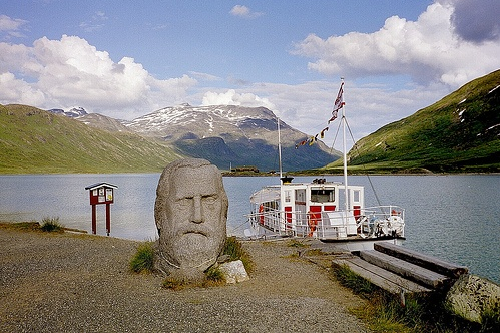}%
                \hspace{1pt}
                \includegraphics[width=0.15\textwidth, height=0.10\textwidth]{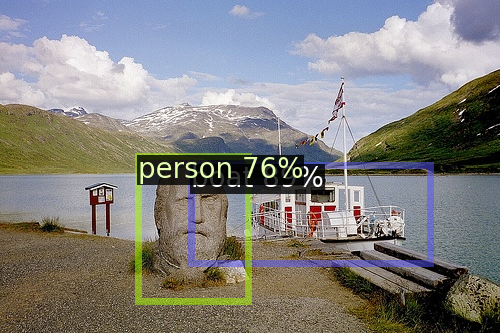}%
                \hspace{1pt}
                \includegraphics[width=0.15\textwidth, height=0.10\textwidth]{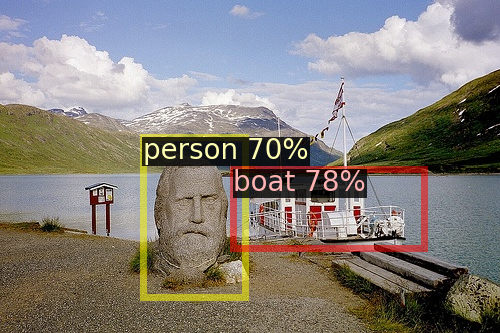}%
                \hspace{1pt}
                \includegraphics[width=0.15\textwidth, height=0.10\textwidth]{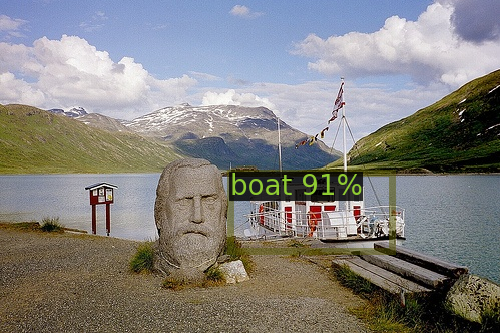}%

                \vspace{2pt}
                
                \includegraphics[width=0.15\textwidth, height=0.10\textwidth]{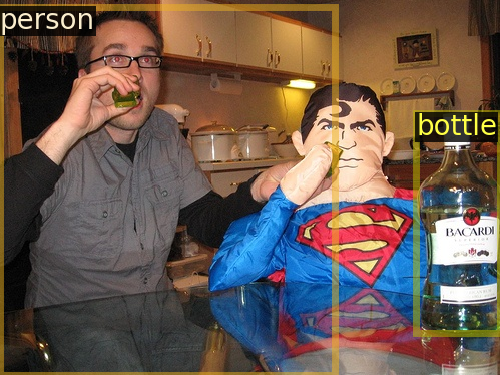}%
                \hspace{1pt}
                \includegraphics[width=0.15\textwidth, height=0.10\textwidth]{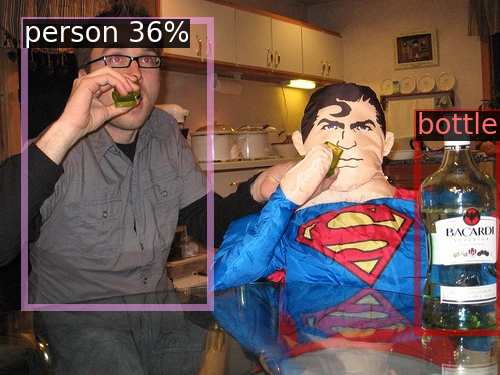}%
                \hspace{1pt}
                \includegraphics[width=0.15\textwidth, height=0.10\textwidth]{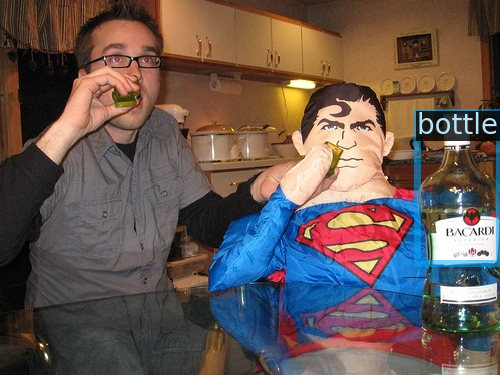}%
                \hspace{1pt}
                \includegraphics[width=0.15\textwidth, height=0.10\textwidth]{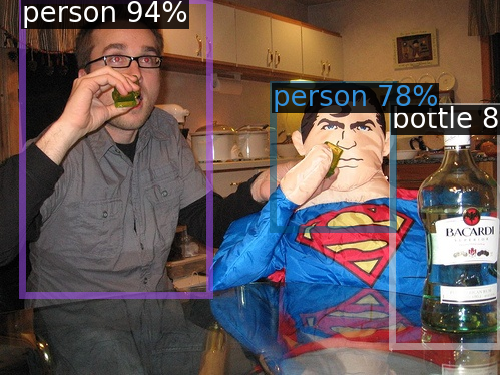}%
                \hspace{1pt}
                \includegraphics[width=0.15\textwidth, height=0.10\textwidth]{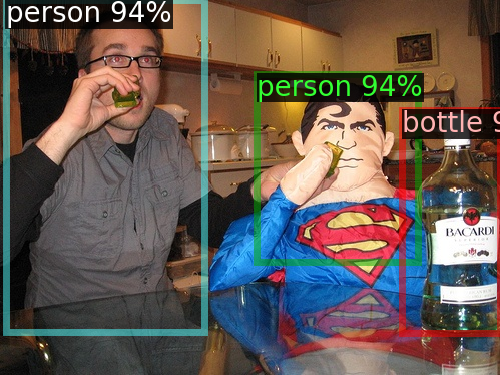}%
                \hspace{1pt}
                \includegraphics[width=0.15\textwidth, height=0.10\textwidth]{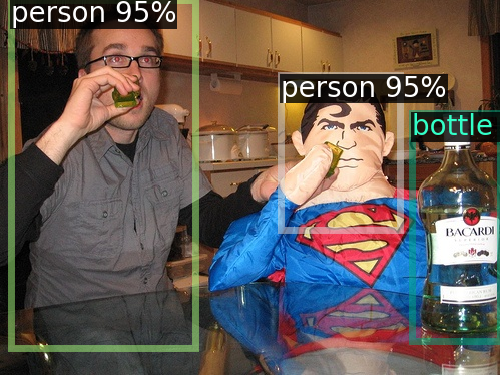}%

                \vspace{2pt}

                \includegraphics[width=0.15\textwidth, height=0.10\textwidth]{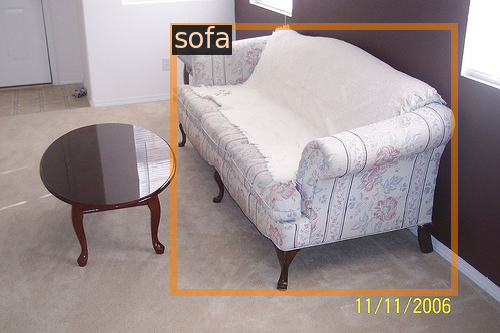}%
                \hspace{1pt}
                \includegraphics[width=0.15\textwidth, height=0.10\textwidth]{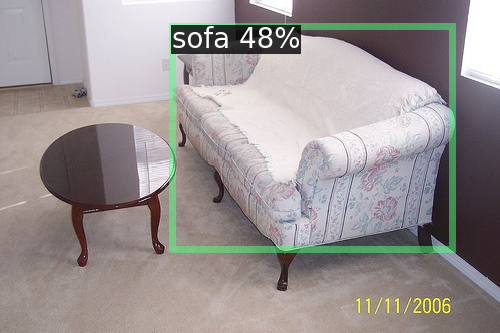}%
                \hspace{1pt}
                \includegraphics[width=0.15\textwidth, height=0.10\textwidth]{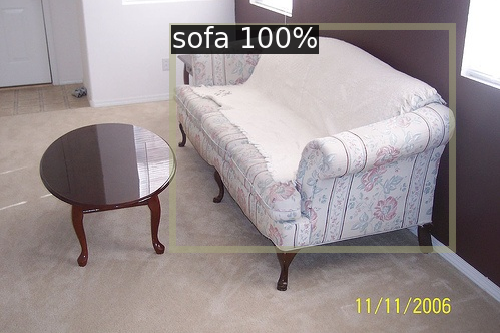}%
                \hspace{1pt}
                \includegraphics[width=0.15\textwidth, height=0.10\textwidth]{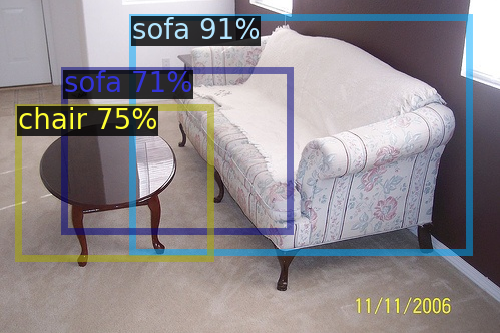}%
                \hspace{1pt}
                \includegraphics[width=0.15\textwidth, height=0.10\textwidth]{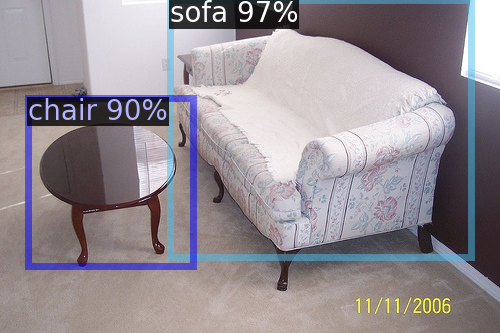}%
                \hspace{1pt}
                \includegraphics[width=0.15\textwidth, height=0.10\textwidth]{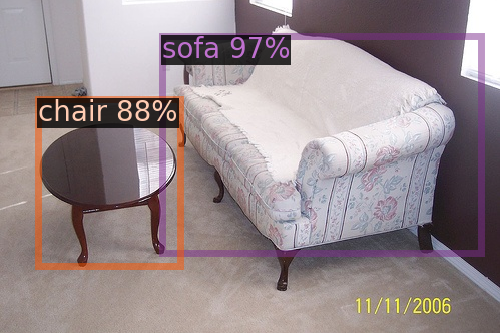}%

            \end{center}
            
        \end{subfigure}
    \end{center}
    %    \vspace{-10pt}
    \caption{
        Visualization of detection results from WSDDN, PCL, UWSOD, the proposed \OurMethod and \OurMethodx.
    }
    \vspace{-15pt}
    \label{fig_vis_result}
\end{figure*}
%%%%%%%%%%%%%%%%%%%%%%%%%%%%%%%%%%%%%
%%
\subsection{Compatibility with other WSOD methods}
\OurMethod is compatible with various orthogonal WSOD methods.
We adapt our method as the WSOD part in multi-stage approaches, \eg, SoS~\cite{Sui} and BiB~\cite{Vo2022}.
SoS proposed a three-stage framework, \ie, WSOD, FSOD, and semi-supervised detection, which pays more attention to relatively high-quality pseudo labels and carries out a dynamic label updating for noisy labels.
BiB introduced an active learning strategy to fine-tune a base pre-trained WSOD model with a few fully-annotated samples, which are automatically selected from the training set according to the failure modes of detectors.
Results are shown in Tab.~\ref{table_voc2007voc2012mscoco_sota_gt_compatibility}.
When integrating with SoS, \OurMethod achieves $64.5\%$ $m$AP and $16.2\%$ AP on VOC 2007 and COCO, respectively.
Under the $10\%$ and $10$ shots setting in active learning, \OurMethod with BiB finally reaches $65.7\%$ $m$AP and $17.9\%$ AP on VOC 2007 and COCO, respectively.
These results demonstrate the flexibility and effectiveness of our \OurMethod.
%%%%%%%%%%%%%%%%%%%%%%%%%%%%%%%%%%%%%%%%%%%%%%%%%%%%%%
\subsection{WSOD in Autonomous Driving}
To explore the value of WSOD in more complex application scenarios, we compare our \OurMethod with common WSOD methods such as WSDDN and OICR on the Cityscapes~\cite{cordts2016cityscapes}.
The Cityscapes dataset contains 2,975 training images and 500 val images  of size 1024 × 2048 taken from a car driving in German cities, labeled with 8 semantic instance categories, i.e. person, rider, car, truck, bus, train, motorcycle and bicycle.
These results shown in Tab.~\ref{table_cityscapes} demonstrate that our \OurMethod comprehensively surpasses the common WSOD method in autonomous driving and achieves a detection performance of 16.1\% $AP_50$.
In the large category, we have significantly improved the OICR by 10.5\%, 14.8\%, 8.7\%, and 4.4\% in the car, truck, bus, and train categories, respectively. This improvement is attributed to our SSOPG, which provides better recall for large objects compared to traditional offline proposal generation methods, and SEM, which better refines the bounding box coordinates.
In the small category, we have improved performance in the person, rider, and motorcycle categories by 0.7\%, 4.0\%, 3.2\%, and 0.2\%, respectively. This is due to our AEOPG, which accurately hypothesize the location of small objects, unlike traditional offline proposal generation methods that often fail to do so. Additionally, our MRRP provides multi-scale information, enabling better detection of small objects.
It is worth noting that person and rider are semantically overlapping and very similar in appearance. Even with full supervision, it is difficult to distinguish between the two. However, our \OurMethod can effectively distinguish between the two under weak supervision.
%

%%%%%%%%%%%%%%%%%%%%%%%%%%%%%%%%%%%%%%%%%%%%%%%%%%%%%%
\subsection{Quantitative Analysis}
We first visualize object proposals from AEOPG, encoded representation $\mathcal{E}(F)$, and low-rank feature maps $F^q$ in Fig.~\ref{fig_vis_aeopg}.
The first three rows show that AEOPG generates high-quality candidates with very few proposals.
AEOPG also distinguishes multiple instances of the same category, as shown in the next two rows.
In the sixth row, AEOPG is able to detect small objects in outdoor scenes.
The seventh row depicts that AEOPG captures salient objects in complex backgrounds.
We also show some incorrect cases in the last row.
Our failure modes mainly come from two parts: (1) object overlap, and (2) cluttered scenes.
Next, we show qualitative comparisons among WSDDN~\cite{Bilen2016}, PCL~\cite{Tang2018b}, UWSOD~\cite{Shen2020UWSOD}, our \OurMethod and \OurMethodx on PASCAL VOC 2007 with ResNet18 backbone in Fig.~\ref{fig_vis_result}.
The first column represents the original images with ground-truth bounding boxes.
In the first two rows, we observe that the proposed \OurMethod provides more accurate detections in complex scenes.
The next two rows show that \OurMethod is able to distinguish multiple instances of the same category.
Our method is robust to the object size, especially for small objects compared to other methods, as shown in the fifth and sixth rows.
We also find that \OurMethodx improves some false positives over \OurMethod in the seventh row of Fig.~\ref{fig_vis_result}.
We finally visualize some failure detection results of our models in the last two rows of Fig.~\ref{fig_vis_result}, which are mainly due to confusion with similar objects.
%%%%%%%%%%%%%%%%%%%%%%%%%%%%%%%%%%%%%%%%%%%%%%%%%%%%%%%%%%%%%%%%%%
\section{Conclusion}
\label{sec_Conclusion}
In this paper, we propose \OurMethod, a novel framework with holistic self-training and unified WSOD network, to develop a high-capacity general detection model with only image-level labels, which is self-contained and does not require external modules or additional supervision.
\OurMethod innovates in two essential prospects, \ie, end-to-end object proposal generation and holistic self-training, both of which are rarely touched in WSOD before.
Moreover, the proposed \OurMethod is upper-bounded in its FSOD counterpart tightly, which demonstrates \OurMethod is capable to achieve fully-supervised accuracy.
Compared with the previous methods that using external proposal modules, \OurMethod has limited performance on challenging MS COCO benchmark, as SSOPG and AEOPG may not have enough supervision to capture large appearance and scale variability.
In a generic large-scale scenario, traditional proposal methods have been a critical auxiliary module for WSOD community to progress.
It enables to see significant improvements before end-to-end proposal generators are available, \ie, SSOPG and AEOPG.
Thus, exploiting well-designed principles in traditional proposal methods is a promising research direction for WSOD proposal learning.
As data augmentation policies plays a key role in CR, further investigation of augmentation variants is also promising, which we leave to our future work.
%%%%%%%%%%%%%%%%%%%%%%%%%%%%%%%%%%%%%%%%%%%%%%%%%%%%%%%%%%

%% IJCV
% \bibliographystyle{spbasic}
\bibliography{../main.bib}

\end{document}